\documentclass{article}

\usepackage[verbose=true,letterpaper]{geometry}
\AtBeginDocument{
  \newgeometry{
    textheight=9in,
    textwidth=5.5in,
    top=1in,
    headheight=12pt,
    headsep=25pt,
    footskip=30pt
  }
}

\usepackage[utf8]{inputenc}
\usepackage[T1]{fontenc}
\usepackage{hyperref}
\usepackage{url}
\usepackage{booktabs}
\usepackage{amsfonts}
\usepackage{nicefrac}
\usepackage{microtype}
\usepackage[svgnames,x11names,table]{xcolor}
\usepackage{afterpage}
\usepackage{lscape}
\usepackage{enumitem}
\usepackage{makecell}
\usepackage{tabularx}
\usepackage{bm}
\usepackage{wrapfig}
\usepackage{multirow}
\usepackage{amsmath}
\usepackage{cleveref}
\usepackage{subcaption}
\usepackage{tikz}
\usepackage{tikzscale}
\usepackage{lipsum}
\usepackage{pgfplots}
\usepackage{placeins}
\usepackage{stmaryrd}
\usepackage{dsfont}
\usepackage{pgfmath}
\usepackage{float}
\usepackage{caption}

\usetikzlibrary{positioning, matrix, shapes.geometric, arrows.meta, fit, backgrounds, calc, intersections, fillbetween}
\usepgfplotslibrary{fillbetween}
\pgfplotsset{compat=1.18}
\usetikzlibrary{external}
\definecolor{darkmagenta}{rgb}{.38,0,0.38}
\newcolumntype{C}{>{\centering\arraybackslash}X}
\def\BibTeX{{\rm B\kern-.05em{\sc i\kern-.025em b}\kern-.08em\TeX}} 

\newcommand{\shortminus}{\mkern-2mu{\scriptscriptstyle-}\mkern-2mu}
\renewcommand{\shortminus}{\mathord{\raisebox{0.2pt}{\scalebox{0.5}[0.8]{\(\mkern-5mu - \mkern-3mu\)}}}}
\newcommand{\ppm}{\mathord{\raisebox{0.1pt}{\scalebox{0.7}{\(\mkern-10mu\pm\mkern-5mu\)}}}}
\newcommand{\sizestd}{3}
\newcommand{\sizestdinter}{4}
\newcommand{\std}[1]{\scalebox{1.}{\color{black!70!white}\fontsize{\sizestd}{\sizestdinter}\selectfont \(\ppm\) #1}}

\newcommand{\half}{\nicefrac{1}{2}}

\usepackage[numbers,sort&compress]{natbib}
\usepackage{authblk}

\DeclareMathOperator{\mean}{mean}

\title{(U)NFV: Supervised and Unsupervised Neural Finite Volume Methods for Solving Hyperbolic PDEs}

\author[1,$\ast$]{Nathan Lichtlé}
\author[1,$\ast$]{Alexi Canesse}
\author[1,$\ast$]{Zhe Fu}
\author[1,$\ast$]{Hossein Nick Zinat Matin}
\author[1]{Maria Laura Delle Monache}
\author[1]{Alexandre M. Bayen}

\affil[1]{University of California, Berkeley}
\affil[$\ast$]{Equal contribution}
\affil[ ]{\texttt{lichtle@berkeley.edu}, \texttt{alexi.canesse@berkeley.edu}, \texttt{zhefu@berkeley.edu}, \texttt{h-matin@berkeley.edu}, \texttt{mldellemonache@berkeley.edu}, \texttt{bayen@berkeley.edu}}

\newcommand{\linkwebsite}{\href{https://nathanlichtle.com/research/nfv/}{\texttt{nathanlichtle.com/research/nfv}}}
\newcommand{\linkcode}{\href{https://github.com/nathanlct/nfv}{\texttt{github.com/nathanlct/nfv}}}
\newcommand{\linkwebsiteshort}{\href{https://nathanlichtle.com/research/nfv/}{our webpage}}

\begin{document}

\date{}
\maketitle

\begin{center}
    \color{RoyalBlue4}
    \textbf{Code, dataset and trained models:} \linkcode \\
    \textbf{Supplementary material and videos:} \linkwebsite
\end{center}

\begin{abstract}
  \noindent\textbf{We introduce (U)NFV, a modular neural network architecture that generalizes classical finite volume (FV) methods for solving hyperbolic conservation laws.}
  Hyperbolic partial differential equations (PDEs) are challenging to solve, particularly conservation laws whose physically relevant solutions contain shocks and discontinuities. FV methods are widely used for their mathematical properties: convergence to entropy solutions, flow conservation, or total variation diminishing, but often lack accuracy and flexibility in complex settings. 
  \emph{Neural Finite Volume} addresses these limitations by learning update rules over extended spatial and temporal stencils while preserving conservation structure. It supports both supervised training on solution data (NFV) and unsupervised training via weak-form residual loss (UNFV). Applied to first-order conservation laws, (U)NFV achieves up to \textbf{10x lower error} than Godunov's method, outperforms ENO/WENO, and rivals discontinuous Galerkin solvers with far less complexity. On traffic modeling problems, both from PDEs and from experimental highway data, (U)NFV captures nonlinear wave dynamics with significantly higher fidelity and scalability than traditional FV approaches.
\end{abstract}

\vspace{-0.3cm}

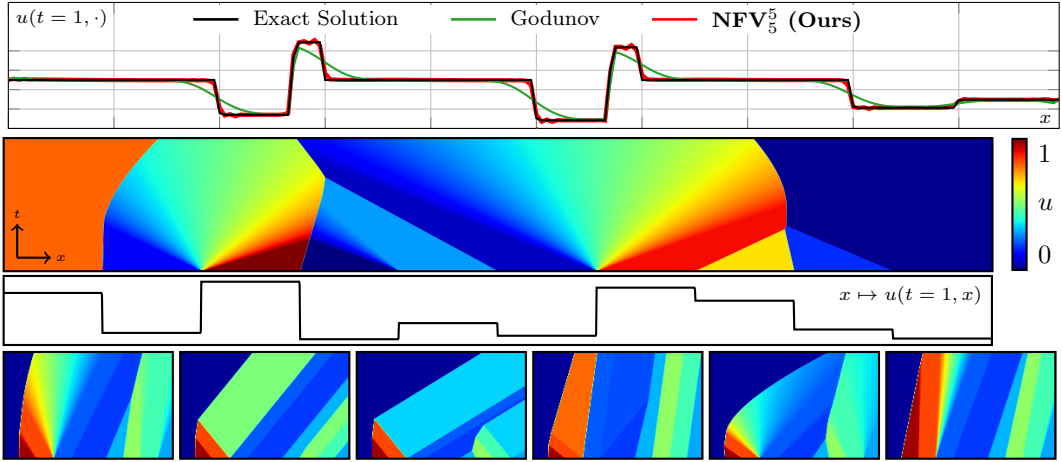
\begin{figure}[h]
  \centering
  \hspace{-10pt}
  \begin{subfigure}[t]{1.0\textwidth}
      \begingroup
      \centering
      \begin{tikzpicture}[baseline]
            \definecolor{darkgray176}{RGB}{176,176,176}
            \definecolor{darkorange25512714}{RGB}{255,0,0}
            \definecolor{forestgreen4416044}{RGB}{44,160,44}
            \definecolor{lightgray204}{RGB}{204,204,204}
            \definecolor{steelblue31119180}{RGB}{100,0,255}
            \definecolor{pltblue}{HTML}{1F77B4}
            \definecolor{pltorange}{HTML}{FF7F0E}
            \definecolor{pltgreen}{HTML}{2CA02C}
          \scriptsize
          \definecolor{darkgray176}{RGB}{176,176,176}
          \definecolor{darkorange25512714}{RGB}{255,0,0}
          \definecolor{forestgreen4416044}{RGB}{44,160,44}
          \definecolor{lightgray204}{RGB}{204,204,204}
          \definecolor{steelblue31119180}{RGB}{100,0,255}
          
          \begin{axis}[
              legend cell align={left},
              legend style={fill opacity=0.8, draw=none, fill=white, fill opacity=1, font=\scriptsize, draw opacity=1, text opacity=1,at={(0.5,0.75)}, anchor=south, draw=none, inner sep=0pt, nodes={inner ysep=1.37pt, inner xsep=5pt}, /tikz/every even column/.append style={column sep=0.5cm}},
              width=1\textwidth,
              height=.12\textwidth,
              xmin=-.1, xmax=.099,
              ymin=-0, ymax=1.3,
              legend columns=3,
              ytick={0,0.2,0.4,0.6,0.8},
              xticklabels=\empty,
              xlabel style={yshift=10pt},
              yticklabels=\empty,
              scale only axis,
              grid=major,
          ]
          \node[anchor=north west] at (axis cs:-0.1,1.31) {$u(t=1,\cdot)$};
          \node[anchor=north east] at (axis cs:0.099,0.22) {$x$};
          
          \addplot[line width=1pt,mark=none,color=black] coordinates {(0,0) (0,0)};
          \addlegendentry{\footnotesize Exact Solution};
          \addplot[line width=1pt,mark=none,color=pltgreen] coordinates {(0,0) (0,0)};
          \addlegendentry{\footnotesize Godunov};
          \addplot[line width=1pt,mark=none,color=red] coordinates {(0,0) (0,0)};
          \addlegendentry{\footnotesize \textbf{NFV\(^5_5\) (Ours)}};
            
          \addplot [ultra thick, red] 
          table[y index=0, x expr=\coordindex*0.001 - .1] {triangular_ours.tex};

          \addplot [thick, forestgreen4416044]  
          table[y index=0, x expr=\coordindex*0.001 - .1] {triangular_godunov.tex};
          
          \addplot [thick, black]
          table[y index=0,  x expr=\coordindex*0.001 - .1] {triangular_lax_hopf.tex};
          \end{axis}
      \end{tikzpicture}
      \endgroup
  \end{subfigure}

  \vspace{-0.12cm}
  \begingroup
  \begin{subfigure}[t]{0.94\textwidth}
    \centering
      \begin{tikzpicture}[scale=0.9, clip=false]
          \tiny
                  \node[anchor=south west, inner sep=0] (img) at (0,0) {\includegraphics[width=\textwidth, height=50pt]{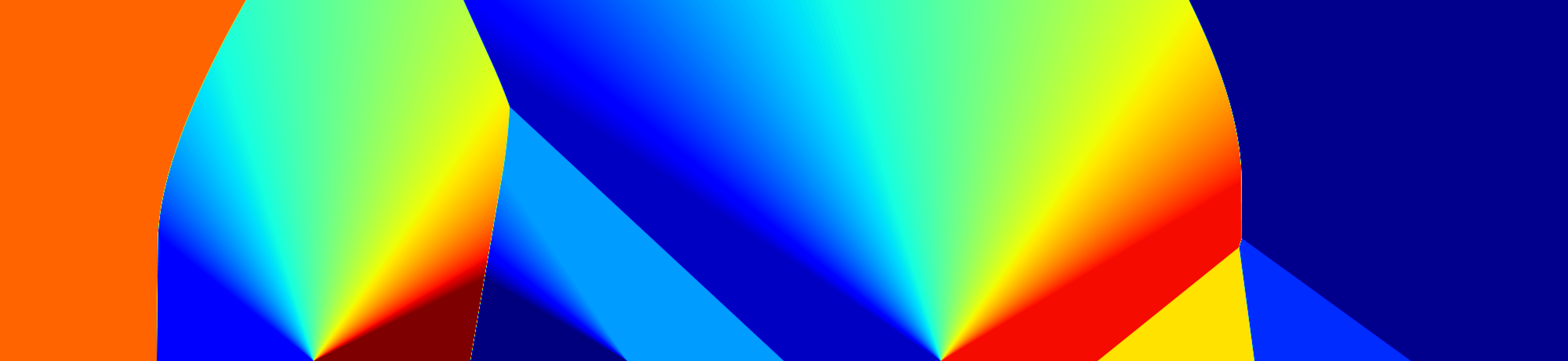}};
                  \draw[black,thick] (img.south west) rectangle (img.north east);
                  
                  \draw[black, thick, -to] (0.2, 0.2) -- (0.2, 0.7);
                  \node[black] at (0.2, 0.85) {\(t\)};
                  
                  \draw[black, thick, -to] (0.2, 0.2) -- (0.7, 0.2);
                  \node[black] at (0.85, 0.2) {\(x\)};
              \end{tikzpicture}
          \end{subfigure}
          \hfill
          \begin{subfigure}[t]{.04\textwidth}
              \centering
              \begin{tikzpicture}[scale=1., clip=false]
                  \node[anchor=south west, inner sep=0] (img) at (0,0) {\includegraphics[height=1.73cm]{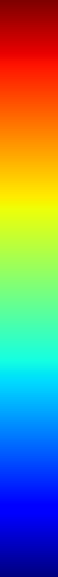}};
                  \draw[black, thick] (img.south west) rectangle (img.north east);
                  \path (img.south east) -- (img.north east) node
                  [midway, right, xshift=+0.03cm] {\(u\)};
                  \node[right, xshift=+0.03cm, yshift=+0.2cm] at (img.south east) {\(0\)};
                  \node[right, xshift=+0.03cm, yshift=-0.2cm] at (img.north east) {\(1\)};
              \end{tikzpicture}
          \end{subfigure}
          \endgroup

\begingroup
          \centering
          \begin{subfigure}[t]{0.945\textwidth}
          \begin{tikzpicture}
            \begin{axis}[
                xlabel={},
                ylabel={},
                xtick=\empty,
                ytick=\empty,
                grid=major,
                thick,
                width=1.115\textwidth,
                height=2.5cm,
                xmin=0,
                xmax=2000,
            ]
                \addplot[
                    black,
                    mark=none
                ] table [
                    y index=0,
                    x expr=\coordindex
                ] {ic_burgers_teaser.tex};
                \node[anchor=north east] at (axis cs:2000,1.11) {\scriptsize $x \mapsto u(t=1,x)$};
            \end{axis}

          \end{tikzpicture}
        \end{subfigure}
        \hfill
        \phantom{a}
\endgroup

          \vspace{0.03cm}
\begin{subfigure}[t]{.16\textwidth}
  \centering
  \begin{tikzpicture}[scale=0.9, clip=false]
    \tiny
    \node[anchor=south west, inner sep=0] (img) at (0,0) {\includegraphics[width=\textwidth, height=40pt]{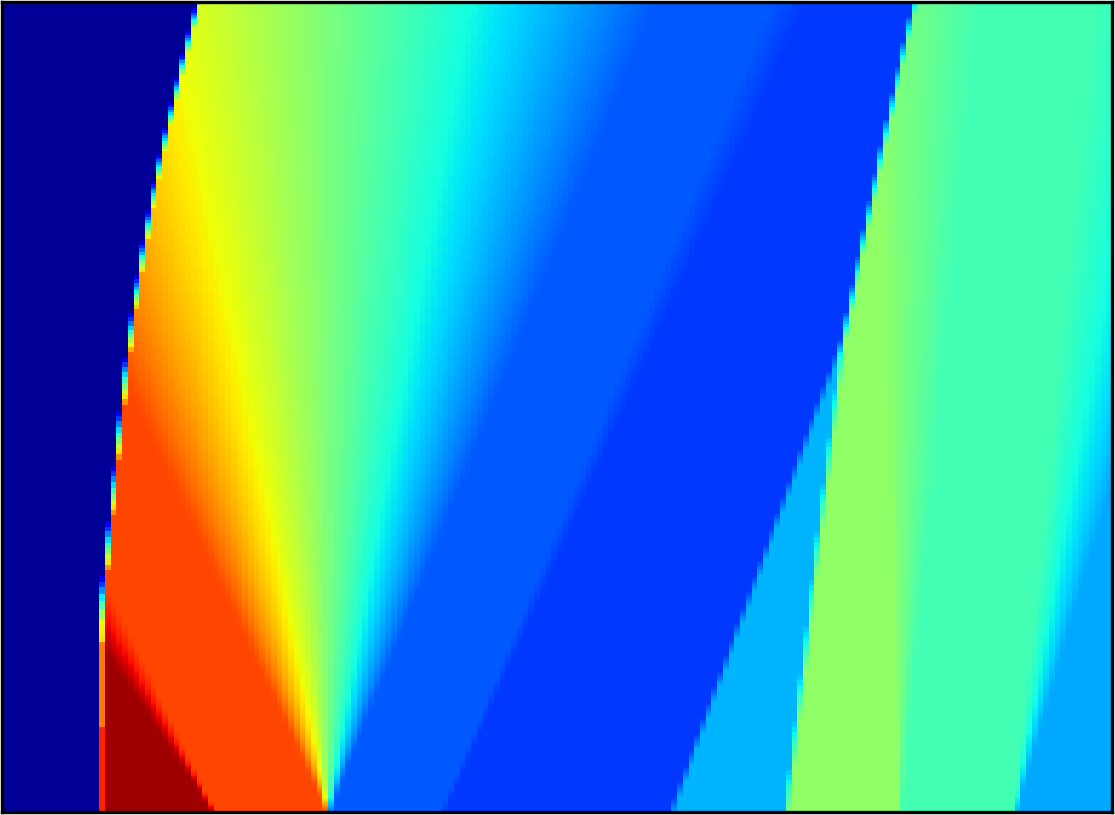}};
    \draw[black,thick] (img.south west) rectangle (img.north east);
  \end{tikzpicture}
\end{subfigure}
\hfill
\begin{subfigure}[t]{.16\textwidth}
  \centering
  \begin{tikzpicture}[scale=0.9, clip=false]
    \tiny
    \node[anchor=south west, inner sep=0] (img) at (0,0) {\includegraphics[width=\textwidth, height=40pt]{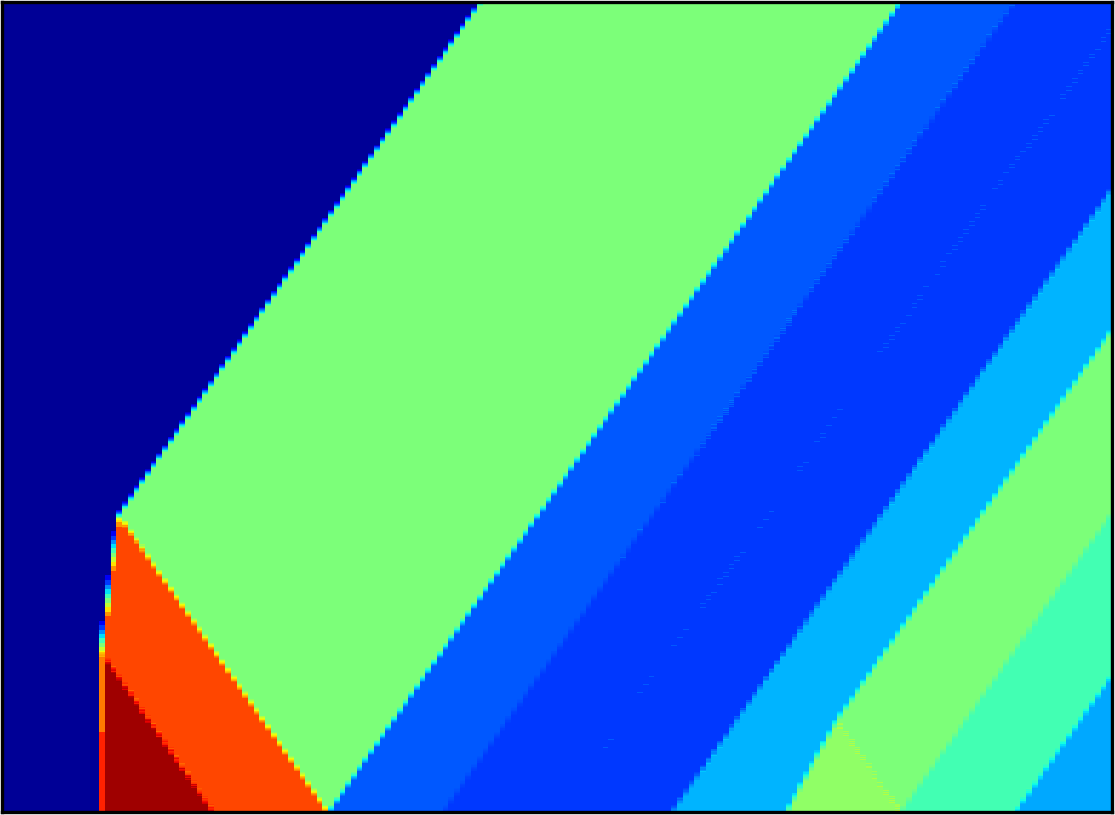}};
    \draw[black,thick] (img.south west) rectangle (img.north east);
  \end{tikzpicture}
\end{subfigure}
\hfill
\begin{subfigure}[t]{.16\textwidth}
  \centering
  \begin{tikzpicture}[scale=0.9, clip=false]
    \tiny
    \node[anchor=south west, inner sep=0] (img) at (0,0) {\includegraphics[width=\textwidth, height=40pt]{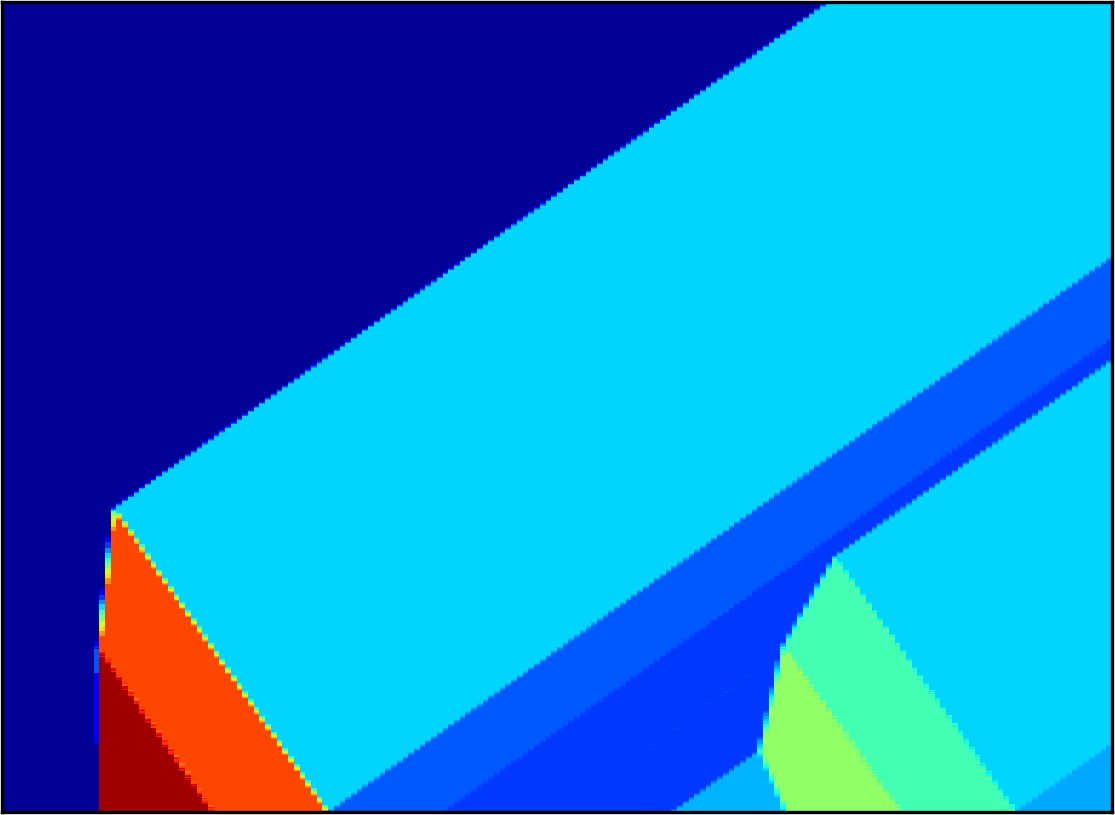}};
    \draw[black,thick] (img.south west) rectangle (img.north east);
  \end{tikzpicture}
\end{subfigure}
\hfill
\begin{subfigure}[t]{.16\textwidth}
  \centering
  \begin{tikzpicture}[scale=0.9, clip=false]
    \tiny
    \node[anchor=south west, inner sep=0] (img) at (0,0) {\includegraphics[width=\textwidth, height=40pt]{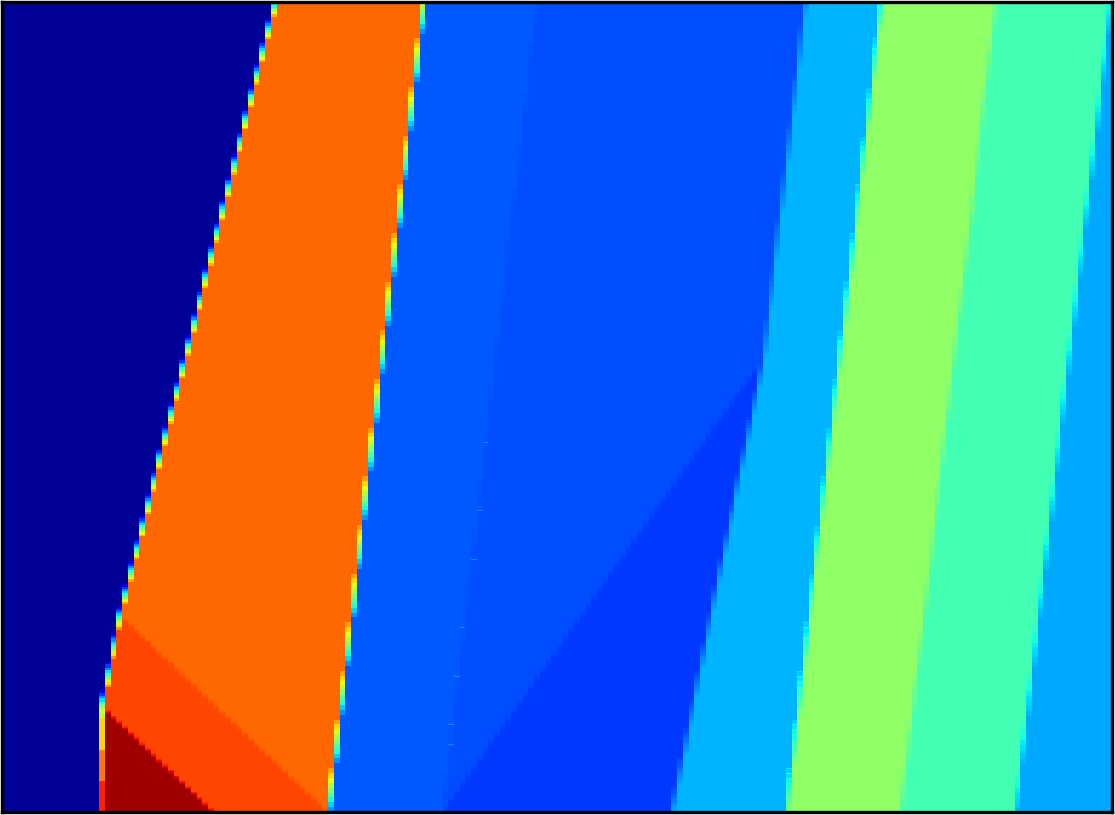}};
    \draw[black,thick] (img.south west) rectangle (img.north east);
  \end{tikzpicture}
\end{subfigure}
\hfill
\begin{subfigure}[t]{.16\textwidth}
  \centering
  \begin{tikzpicture}[scale=0.9, clip=false]
    \tiny
    \node[anchor=south west, inner sep=0] (img) at (0,0) {\includegraphics[width=\textwidth, height=40pt]{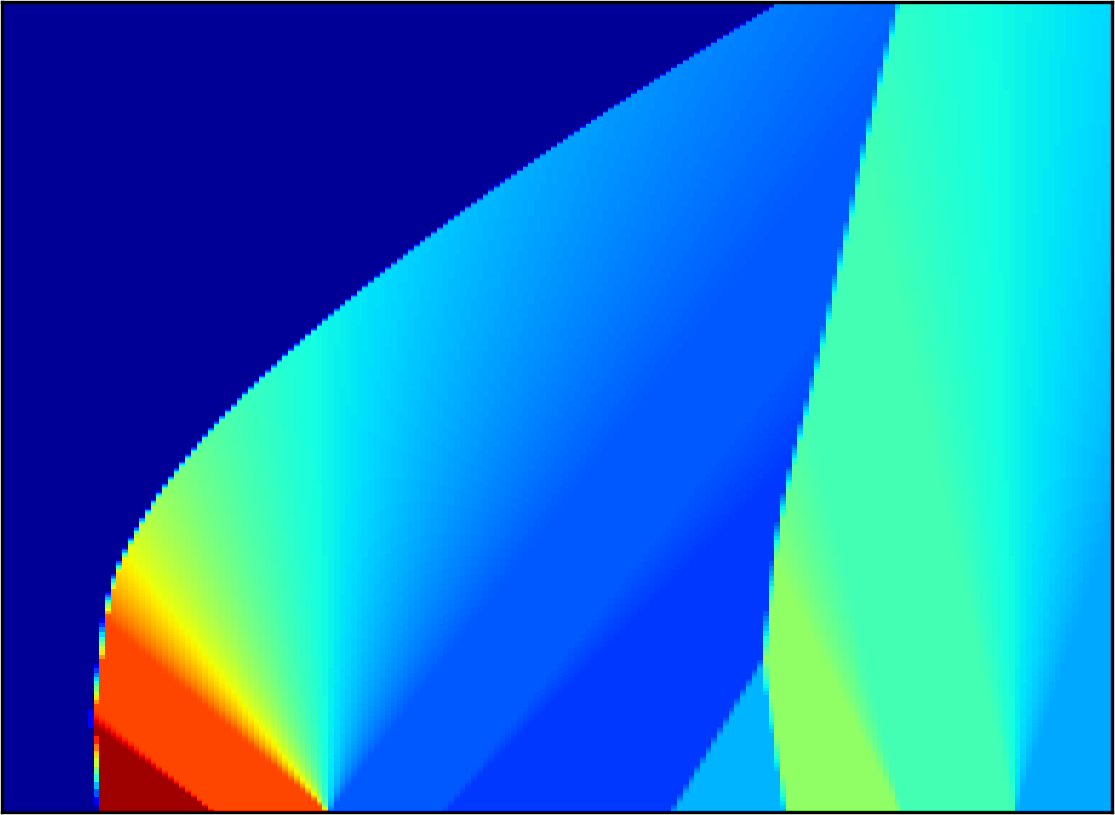}};
    \draw[black,thick] (img.south west) rectangle (img.north east);
  \end{tikzpicture}
\end{subfigure}
\hfill
\begin{subfigure}[t]{.16\textwidth}
  \centering
  \begin{tikzpicture}[scale=0.9, clip=false]
    \tiny
    \node[anchor=south west, inner sep=0] (img) at (0,0) {\includegraphics[width=\textwidth, height=40pt]{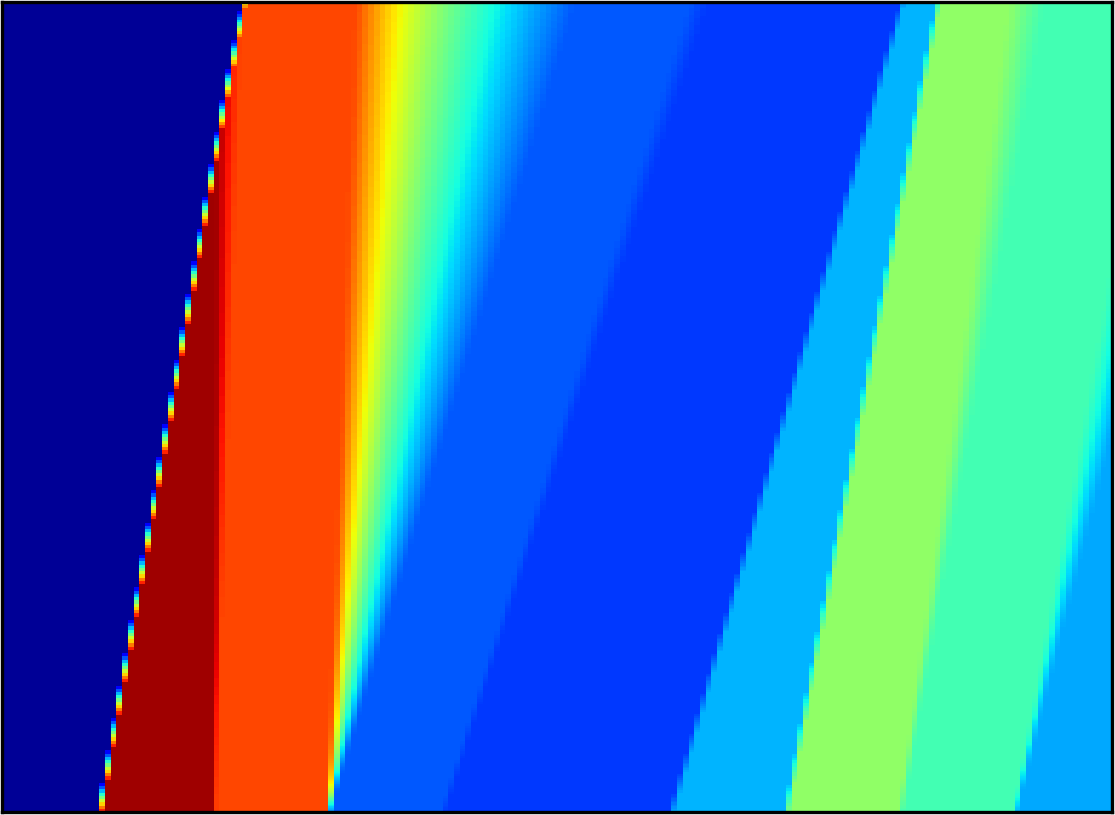}};
    \draw[black,thick] (img.south west) rectangle (img.north east);
  \end{tikzpicture}
\end{subfigure}

\caption{\textbf{Prediction of entropy solutions of hyperbolic PDEs.} \textbf{Top:} NFV\(^5_5\) prediction vs. the Godunov scheme for Burgers' equation at a fixed time. \textbf{Mid:} Entropic solution $u(t,x)$ for Burgers' equation over domain $(t,x) \in [0,1]^2$, and its corresponding initial condition. \textbf{Bottom:} Entropic solution $u(t,x)$ for the LWR equation with different fluxes sharing the same initial condition.}

  \label{fig:teaser}
\end{figure} 
\section{Introduction}
\label{sec:introduction}

\textit{Hyperbolic} partial differential equations (PDEs)~\citep{evans2022partial} are fundamental tools to model propagation and transport phenomena with nonlinear or discontinuous behavior, appearing in areas like fluid dynamics and traffic flow.  
In the present work, we focus on an essential subclass: \emph{conservation laws}~\citep{leveque2002finite}, which encode the principle that certain physical quantities, such as mass, momentum, or energy, must be preserved over time. A general one-dimensional scalar conservation law takes the form:
\begin{equation}
    \partial_t u(x,t) + \partial_x f(u(x,t)) = 0,
    \label{eq:conservation_law}
\end{equation}
where \( u \) is the conserved quantity and \( f \) is the \emph{flux function}.

The solutions of hyperbolic PDEs are difficult to approximate due to discontinuities such as shocks (see~\Cref{fig:teaser}), even when starting from smooth initial conditions. 
Consequently, classical (smooth) solutions typically cease to exist after finite time, and one must instead rely on weak solutions, 
Closed-form solutions exist only in rare cases, such as on simple Riemann problems~\cite{leveque2002finite} or through the Lax-Hopf formula~\citep{lax1957hyperbolic, claudel2010lax, claudel2010lax2}) in specific concave or convex settings. As a result, practical applications almost always rely on numerical methods for approximating the PDE solution, with finite volume (FV) methods~\citep{leveque2002finite} being a popular choice due to their ability to track conserved quantities across discontinuities and capture shock dynamics relatively accurately.

Classical FV methods involve important trade-offs between accuracy near discontinuities, computational cost, stencil size, and implementation complexity. 
In recent years, neural networks have been explored as flexible and powerful alternatives solvers, showing promise in learning complex dynamics from data or residuals. Yet, many such methods are designed for non-specific models, often at the expense of losing physical structure, including conservation laws and entropy behavior.

We introduce the \emph{Neural Finite Volume} (NFV) method, a modular architecture tailored to conservation laws, that blends the structure-preserving benefits of FV schemes with the expressiveness of neural networks. Conservation is built into the NFV model, using extended spatial and temporal stencils. We develop both a supervised version, trained on solution data from simple cases, and an unsupervised variant (UNFV), which learns directly from the PDE via a weak-form residual loss. This flexibility allows (U)NFV to adapt to the availability of data, leveraging accurate synthetic or field data when present, or solving directly from the equation when solutions are inaccurate or expensive to obtain.

\paragraph{Contributions.} Our main contributions are as follows: \vspace{-6pt}
\begin{itemize}[itemsep=0.2em, left=1em]
    \itemsep -2pt{}
    \item We propose (U)NFV, a neural architecture that generalizes the structure of finite volume methods and thus preserves conservation properties by construction.
    \item We introduce two variants: a supervised learning one (NFV) and an unsupervised learning one (UNFV), depending on data availability, using either solution data or a weak-form residual loss.
    \item We demonstrate strong numerical results on several conservation laws, achieving up to 10x lower error than classical FV solvers. Additionally, (U)NFV matches the accuracy of discontinuous Galerkin methods, without their mathematical complexity.
    \item We show that NFV can be trained on field data that does not necessarily strictly satisfy the conservation law, and still predicts accurate solutions with more flexibility and generalizability than classical solvers.
\end{itemize}

The remainder of the article is organized as follows: \Cref{sec:related_work} provides a detailed overview of the related work, \Cref{sec:prerequisites} recalls the finite volume method and introduces necessary notation, \Cref{sec:method} describes the proposed (U)NFV method in detail, \Cref{sec:experiments} presents the experiments and results on hyperbolic PDEs, \Cref{sec:i24} extends the NFV to experimental field highway data, and \Cref{sec:conclusion} concludes the article with discussion on broader impacts. Then, \Cref{sec:fvm} provides details about finite volume schemes, \Cref{app:lwr} illustrates six PDE variants considered in this work, \Cref{app:general_PDE} introduces more general PDE models, \Cref{app:i24} expands on the experimental data handling and results from \Cref{sec:i24}, and \Cref{app:exp_details} details the experimental setup. \section{Related work}
\label{sec:related_work}

\paragraph{Numerical methods.}
 Classical numerical methods for hyperbolic PDEs, such as FV and \textit{discontinuous Galerkin} (DG)~\citep{hu1999discontinuous} methods, are widely used due to their capabilities in capturing shocks and discontinuities. First-order schemes such as the \textit{Lax-Friedrichs}~\citep{lax1954initial} and \textit{Godunov}~\citep{godunov1959finite} methods provide robustness but suffer from excessive numerical diffusion, leading to smeared solutions. To address this, higher-order methods like \textit{Essentially Non-Oscillatory} (ENO)~\citep{shu1999high}, \textit{Weighted ENO} (WENO)~\citep{liu1994weighted}, and DG have been introduced, offering improved accuracy in smooth regions while preserving stability near shocks. DG further improves accuracy through local polynomial approximations but incurs high computational costs~\citep{cockburn1998local}.
Despite their accuracy, these methods often require extensive manual effort and careful stabilization, motivating the development of flexible, data-driven alternatives.

\paragraph{NN approaches for PDEs.}
Deep learning has emerged as a powerful alternative for approximating PDE solutions. In supervised settings, neural operators such as \textit{Fourier Neural Operator} (FNO)~\citep{li2020fourier} and \textit{Deep Operator Networks} (DeepONet)~\citep{lu2021learning} efficiently approximate solution mappings from parametric inputs, without requiring explicit mesh discretization in the case of FNO. While successful for general PDEs, these operators have mainly been validated on elliptic or parabolic PDEs, typically characterized by smooth solutions. Conventional neural architectures, such as CNNs~\citep{lecun1995convolutional} for structured domains and GNNs~\citep{bronstein2017geometric} for irregular geometries, have also been adopted as supervised surrogates. However, supervised models rely heavily on large, high-quality labeled datasets, and often lack intrinsic enforcement of physical constraints, leading to limited generalization and poor accuracy on PDEs involving sharp gradients or shocks~\citep{krishnapriyan2021characterizing}.
To reduce data reliance, unsupervised approaches like \textit{Physics-Informed Neural Networks} (PINNs) incorporate PDE residuals directly into training losses~\citep{raissi2017physics}, proving effective for elliptic and parabolic equations~\citep{raissi2019physics,jagtap2020conservative}. However, PINNs encounter significant difficulties with hyperbolic PDEs, especially in capturing discontinuities and shock dynamics, resulting in unstable optimization, convergence failures, and inaccurate solutions~\citep{wang2021understanding,fuks2020limitations}. Recent variants, such as \textit{Weak PINNs} (wPINNs)~\citep{de2024wpinns}, \textit{Parareal PINNs} (PPINNs)~\citep{meng2020ppinn}, and \textit{Extended PINNs} (XPINNs)~\citep{jagtap2020extended}, aim to overcome these issues through weak formulations or specialized training strategies. Nonetheless, these adaptations often introduce considerable complexity and require extensive hyperparameter tuning, highlighting a persistent need for methods inherently suited to hyperbolic PDE challenges.

\paragraph{NNs for hyperbolic PDEs/conservation laws.}
Neural approaches tailored to hyperbolic PDEs have introduced innovations to handle shocks. Weak PINNs (wPINNs)~\citep{de2024wpinns} integrate weak-form residuals or integral constraints to mitigate issues with discontinuities. Others employ neural networks directly within classical FV schemes to learn improved flux reconstructions~\citep{kossaczka2021enhanced, tong2024roenet}. However, these enhancements typically reintroduce complexity, such as extensive manual parameterization or problem-specific adaptivity, diluting the key advantage of neural flexibility and generality. 

Motivated by these limitations, our proposed NFV approach learns local update rules directly from data or PDE residuals. By preserving the fundamental conservation-law structure of traditional FV methods while flexibly leveraging neural networks, NFV achieves significantly higher accuracy, robustness, and scalability with minimal manual intervention.

 \section{Prerequisites and Notations: Finite Volume Methods}
\label{sec:prerequisites}

Standard FV methods, such as presented in~\citet{leveque2002finite}, solve the integral form of the conservation law~(\ref{eq:conservation_law}) on a mesh of uniform cells $I_i = [x_{i-\half}, x_{i+\half}]$, $i = 1, \cdots, I_{\max}$, with cell length \(\Delta x\).
The average of \(u\) over cell \(I_i\) at time \(t_n = n \Delta t\), $n = 1, \cdots, N$, and the numerical flux through the interface \(x_{i+\half}\) over the time step, are given respectively by
\begin{equation}
    u_i^n = \frac{1}{\Delta x} \int_{I_i} u(t_n, x)\, \mathrm{d} x
    \quad \text{and} \quad
    F_{i+1/2}^n = \int_{t_n}^{t_{n+1}} f(u(t, x_{i+1/2}))\, \mathrm{d} t,
\end{equation}
where $\Delta t$ is the time discretization.
A first-order method $\mathcal{F}$ approximates the numerical flux as \(\hat{F}_{i+\half}^n = \mathcal{F}(u_i^n, u_{i+1}^n)\), while higher-order methods leverage additional cell averages.

Let us generalize this framework by including cell averages from previous time steps in order to construct even better approximations. Let $\text{FV}_a^b$ be the class of methods that use a rectangular stencil of $a$ neighboring spatial cells times $b$ past time steps to estimate numerical fluxes. Specifically, let $\bm{u}_{i\pm\half}^n(a-1,b)$ be the left and right $(a-1) \times b$ sub-stencils, as illustrated in blue and green in \Cref{fig:stencil}. 
 Then, an $\text{FV}_a^b$ method $\mathcal{F}$ estimates numerical fluxes as $\hat{F}_{i\pm\half}^n = \mathcal{F}(\bm{u}_{i\pm\half}^n(a-1,b))$. Classical first-order methods, such as Godunov, fall under class $\text{FV}_3^1$; more details about their computation are provided in \Cref{sec:fvm}. To our knowledge, the vast majority of FV methods in the literature use a single time step (i.e., $b=1$) and a small number of spatial cells space cells. Indeed, designing analytical schemes with larger temporal or spatial stencils becomes exponentially more complex. Finally, the update rule is given by the exact relation
\begin{equation}\label{eq:update_rule}
    u^{n+1}_i = u^n_i - \frac{\Delta t}{\Delta x} \left(F_{i+\half}^n - F_{i-\half}^n \right),
\end{equation}
which in practice is approximated using the numerical fluxes $\hat{F}_{i \pm \half}^n$, leading to an approximation $\hat{u}_i^n$ of $u_i^n$. Note that the influx of one cell is the outflux of another, which ensures conservation.

\begin{figure}
    \centering
    \begin{tikzpicture}[scale=1.0]
        \draw[xstep=1cm,ystep=0.75cm,black,thin] (0,0) grid (5,4*0.75);
        
        \foreach \i in {1,...,5} {
            \foreach \j in {1,...,4} {
                \node at (\i-0.5, \j*0.75-0.5*0.75) {\scriptsize $u_{\i}^{\j}$};
            }
        }
  
        \draw[red, ultra thick] (2,0.75*2) rectangle (3,0.75*3);
        \draw[blue, ultra thick, opacity=0.5] (0,0) rectangle (4,1.5);
        \draw[green!50!black, ultra thick, opacity=0.5] (1,0) rectangle (5,1.5);
        \draw[blue, thick, -to] (1.75, 0.75*1.3) -- (2.25, 0.75*1.3) node[midway, above, xshift=0.09cm] {\scriptsize $F_{2.5}^{2}$};
        \draw[green!50!black, thick, -to] (2.75, 0.75*1.3) -- (3.25, 0.75*1.3) node[midway, above, xshift=0.09cm] {\scriptsize $F_{3.5}^{2}$};
    \end{tikzpicture}
    \caption{Example stencil for $\text{FV}_5^2$, which takes in a rectangular stencil of 2 time steps times 5 space cells to compute the next cell average (in red) using \Cref{eq:update_rule} (specifically, both the in-flow and out-flow are computed using the illustrated 2x4 sub-stencils).}
    \vspace{-1em}
    \label{fig:stencil}
  \end{figure}

 \section{Our method: Neural Finite Volume (NFV)}
\label{sec:method}
Our method builds upon the FV framework by using neural networks to approximate the numerical flux. Specifically, we define $\text{NFV}_a^b$ as a generalization of $\text{FV}_a^b$, where the numerical flux $\hat{F}_{i \pm \half}^n$ is predicted by a neural network $\mathcal{N}$ based on a local $a \times b$ spatiotemporal stencil:
\[ \hat{F}_{i \pm \half}^n = \mathcal{N}(\bm{u}_{i \pm \half}^n(a-1,b))\]
The prediction of the solution is then updated using the classical FV update rule~\eqref{eq:update_rule}, ensuring mass conservation. We explore NFV models ranging from $\text{NFV}_3^1$ (matching Godunov's stencil) to $\text{NFV}_{11}^{11}$, using 11 spatial cells and 11 past time steps -- configurations that would be exceedingly complex to design manually due to the high-dimensional stencil involved. This extension enables accurate learning even from noisy field data. In practice, we implement NFV as a CNN~\citep{lecun1995convolutional}, which allows efficient computation across stencils due to the vectorized nature of CNNs. We refer to \Cref{app:exp_details} for implementation details.

We propose two variants: the supervised $\text{NFV}_a^b$, trained on solution data, and the unsupervised $\text{UNFV}_a^b$, trained directly from the PDE via a weak-form residual loss. Both share the same neural network architecture and only differ in their training objective. This framework is thus designed to be adapted to different kinds of problems.

\subsection{Learning Objectives: Supervised and Unsupervised}
Solving hyperbolic PDEs presents unique challenges, particularly due to the absence of closed-form solutions, even under simple initial conditions. To address this, we propose two distinct learning objectives: supervised and unsupervised. The supervised objective is applicable when reference solutions are available. Conversely, the unsupervised objective is employed when such solutions are unavailable, allowing the model to infer solutions by adhering to the underlying physical laws. Notably, supervised learning can also be utilized in scenarios where the governing equations are unknown, provided that observational data is accessible. This approach facilitates the application of numerical solvers to empirical data, imposing only fundamental physical constraints, such as mass conservation, without requiring extensive hyperparameter tuning (see Section~\ref{sec:i24}).

\subsubsection{Supervised Learning}
Supervised learning offers a straightforward framework for training models when reference solutions are available. In this study, we employ supervised learning not only to approximate the solution of known equations but also to predict field data with unknown governing equations. Although solutions to hyperbolic PDEs are typically defined in the \(L_1\) space, we consider their restrictions to compact subsets where the functions are bounded, thereby allowing treatment within the \(L_2\) space. Accordingly, the loss function is defined as the standard mean square error:
\begin{equation*}
    \mathcal L_s = \underset{\substack{u_0 \sim \mathcal R}}{\mathbb E}||u - \hat u||_2^2
\end{equation*} where \(u\) is the true solution, \(\hat u\) is the predicted solution, and \(\mathcal R\) is a distribution over initial conditions.

\subsubsection{Unsupervised Learning}
 Unsupervised learning for hyperbolic PDEs is inherently more challenging due to the nature of their solutions. Closed-form solutions rarely exist, even for simple cases, and classical (strong) solutions often do not exist altogether. As a result, these equations are typically solved in a weak form. However, weak solutions are not unique, as multiple solutions can satisfy the PDE, but only one adheres to the entropy condition, making it physically meaningful.

The unsupervised loss function is defined to minimize the residuals of the weak formulation, in order to approximate the entropic solution. While imposing this loss does not guarantee convergence to the entropic solution, empirical results indicate that our method successfully converges to the entropy solution across various equations and numerous trials. To enhance learning efficiency, we optimize the weak formulation independently at each time step by minimizing the squared residuals. The collection of test functions \(\Phi\) consists of 250 randomly sampled, compactly supported polynomials of degree 50 over the spatial domain. The unsupervised loss reads:
\begin{equation*}
    \mathcal L_w = \underset{\substack{\varphi \in \Phi\\ u_0 \sim \mathcal R}}{\mathbb E} \left[ \sum_{n=1}^N\left(\sum_{i=1}^{I_{\max}}\left((\Delta t)^{-1}(\hat u_i^n - \hat u_i^{n-1}) \int_{I_i}\varphi + f(\hat u_i^n)[\varphi]_{x_{i - 1/2}}^{x_{i + 1/2}} \right)\right)^2 \right]
\end{equation*} where $\hat u_i^n$ denotes the predicted solution at spatial index $i$ and time step $n$, and \(\mathcal R\) is a distribution over initial conditions.
 \section{Experiments}
\label{sec:experiments}

Experiments have been designed to answer four main questions: \vspace{-6pt}
\begin{itemize}[itemsep=0.2em, left=1em]
    \itemsep -2pt{}
    \item Is (U)NFV a compelling alternative to classical finite volume methods?
    \item Does UNFV converge to an entropic solution despite being trained on the weak formulation?
    \item How does (U)NFV compare to much more complicated finite element methods?
    \item Can NFV perform well on field data that contains noise and may not be conservative?
\end{itemize}

\subsection{Baselines}

Selecting appropriate baselines for PDE solvers poses challenges due to the diversity in computational frameworks: methods vary by mesh dependency (mesh-free versus mesh-based), solution generation (autoregressive versus single-pass), and generalizability (operator-based versus retrained per initial condition). Therefore, we adopt classical numerical schemes, the foundation of our NFV method, as baselines, ensuring a fair comparison. Given the fact that NFV is developed based on traditional first-order FV methods, the present work provides a compelling case for replacing standard FV solvers with the simpler yet effective NFV method whenever FV methods are typically employed. We consider all the numerical schemes introduced in \Cref{sec:related_work} as baselines: first-order FV methods (Godunov, Lax-Friedrichs, and Engquist-Osher), higher-order ones (ENO, WENO), and DG, a finite-element method that is well-known for superior accuracy but suffers from computational burden. More details can be found in \Cref{sec:fvm}.

\begin{table}
    \centering
    \small
    \caption{Performance comparison between neural network models and classical numerical schemes.
    Results are computed over the evaluation set of 1000 piecewise constant initial conditions. For each method, we report mean and standard deviation in \(L_2\) norm (\(\mean((u-\hat u)^2)\)).}
    \begin{tabularx}{\linewidth}{CCCCCCCCC} \hline \Xhline{1.pt} 
         & \multicolumn{5}{c}{\textbf{1\textsuperscript{st} order FV}} & \multicolumn{2}{c}{\textbf{Higher order FV}} & \multicolumn{1}{c}{\textbf{FEM}}\\
         \Xhline{1.pt} 
         & \textbf{NFV}$_3^1$ & \textbf{UNFV}$_3^1$ & GD & LF & EO & ENO & WENO & DG\\
         \Xhline{1.pt} 
          G.shields & \(\bm{1.3\mathrm{e}^{\shortminus 4}}\)\std{\(4\mathrm{e}^{\shortminus 5}\)} & \(2.0\mathrm{e}^{\shortminus 4}\)\std{\(6\mathrm{e}^{\shortminus 5}\)} & \(4.5\mathrm{e}^{\shortminus 4}\)\std{\(2\mathrm{e}^{\shortminus 4}\)} & \(1.3\mathrm{e}^{\shortminus 2}\)\std{\(4\mathrm{e}^{\shortminus 3}\)} & \(4.5\mathrm{e}^{\shortminus 4}\)\std{\(2\mathrm{e}^{\shortminus 4}\)} & \(6.4\mathrm{e}^{\shortminus 4}\)\std{\(4\mathrm{e}^{\shortminus 4}\)} & \(6.4\mathrm{e}^{\shortminus 4}\)\std{\(4\mathrm{e}^{\shortminus 4}\)} & \(3.1\mathrm{e}^{\shortminus 5}\)\std{\(1\mathrm{e}^{\shortminus 5}\)} \\
  Tri. 1 & \(\bm{1.4\mathrm{e}^{\shortminus 3}}\)\std{\(6\mathrm{e}^{\shortminus 4}\)} & \(1.9\mathrm{e}^{\shortminus 3}\)\std{\(9\mathrm{e}^{\shortminus 4}\)} & \(2.3\mathrm{e}^{\shortminus 3}\)\std{\(1\mathrm{e}^{\shortminus 3}\)} & \(9.6\mathrm{e}^{\shortminus 3}\)\std{\(4\mathrm{e}^{\shortminus 3}\)} & \(2.3\mathrm{e}^{\shortminus 3}\)\std{\(1\mathrm{e}^{\shortminus 3}\)} & \(2.0\mathrm{e}^{\shortminus 3}\)\std{\(2\mathrm{e}^{\shortminus 3}\)} & \(1.9\mathrm{e}^{\shortminus 3}\)\std{\(2\mathrm{e}^{\shortminus 3}\)} & \(2.6\mathrm{e}^{\shortminus 4}\)\std{\(1\mathrm{e}^{\shortminus 4}\)} \\
  Tri. 2 & \(\bm{2.4\mathrm{e}^{\shortminus 3}}\)\std{\(1\mathrm{e}^{\shortminus 3}\)} & \(3.1\mathrm{e}^{\shortminus 3}\)\std{\(2\mathrm{e}^{\shortminus 3}\)} & \(3.8\mathrm{e}^{\shortminus 3}\)\std{\(2\mathrm{e}^{\shortminus 3}\)} & \(1.4\mathrm{e}^{\shortminus 2}\)\std{\(8\mathrm{e}^{\shortminus 3}\)} & \(3.8\mathrm{e}^{\shortminus 3}\)\std{\(2\mathrm{e}^{\shortminus 3}\)} & \(5.8\mathrm{e}^{\shortminus 3}\)\std{\(4\mathrm{e}^{\shortminus 3}\)} & \(5.8\mathrm{e}^{\shortminus 3}\)\std{\(4\mathrm{e}^{\shortminus 3}\)} & \(4.1\mathrm{e}^{\shortminus 4}\)\std{\(2\mathrm{e}^{\shortminus 4}\)} \\
  Trapez. & \(\bm{1.1\mathrm{e}^{\shortminus 3}}\)\std{\(4\mathrm{e}^{\shortminus 4}\)} & \(1.6\mathrm{e}^{\shortminus 3}\)\std{\(7\mathrm{e}^{\shortminus 4}\)} & \(2.1\mathrm{e}^{\shortminus 3}\)\std{\(8\mathrm{e}^{\shortminus 4}\)} & \(2.5\mathrm{e}^{\shortminus 2}\)\std{\(1\mathrm{e}^{\shortminus 2}\)} & \(2.1\mathrm{e}^{\shortminus 3}\)\std{\(8\mathrm{e}^{\shortminus 4}\)} & \(6.2\mathrm{e}^{\shortminus 4}\)\std{\(2\mathrm{e}^{\shortminus 4}\)} & \(5.3\mathrm{e}^{\shortminus 4}\)\std{\(2\mathrm{e}^{\shortminus 4}\)} & \(2.9\mathrm{e}^{\shortminus 4}\)\std{\(1\mathrm{e}^{\shortminus 4}\)} \\
  G.berg & \(\bm{1.4\mathrm{e}^{\shortminus 4}}\)\std{\(9\mathrm{e}^{\shortminus 5}\)} & \(3.8\mathrm{e}^{\shortminus 4}\)\std{\(2\mathrm{e}^{\shortminus 4}\)} & \(4.9\mathrm{e}^{\shortminus 4}\)\std{\(2\mathrm{e}^{\shortminus 4}\)} & \(5.3\mathrm{e}^{\shortminus 3}\)\std{\(2\mathrm{e}^{\shortminus 3}\)} & \(4.9\mathrm{e}^{\shortminus 4}\)\std{\(2\mathrm{e}^{\shortminus 4}\)} & \(1.1\mathrm{e}^{\shortminus 3}\)\std{\(6\mathrm{e}^{\shortminus 4}\)} & \(1.2\mathrm{e}^{\shortminus 3}\)\std{\(9\mathrm{e}^{\shortminus 4}\)} & \(3.4\mathrm{e}^{\shortminus 4}\)\std{\(2\mathrm{e}^{\shortminus 3}\)}\\
  U.wood & \(\bm{3.8\mathrm{e}^{\shortminus 4}}\)\std{\(1\mathrm{e}^{\shortminus 4}\)} & \(6.9\mathrm{e}^{\shortminus 4}\)\std{\(2\mathrm{e}^{\shortminus 4}\)} & \(9.2\mathrm{e}^{\shortminus 4}\)\std{\(3\mathrm{e}^{\shortminus 4}\)} & \(2.7\mathrm{e}^{\shortminus 2}\)\std{\(1\mathrm{e}^{\shortminus 2}\)} & \(9.2\mathrm{e}^{\shortminus 4}\)\std{\(3\mathrm{e}^{\shortminus 4}\)} & \(1.1\mathrm{e}^{\shortminus 4}\)\std{\(3\mathrm{e}^{\shortminus 5}\)} & \(9.8\mathrm{e}^{\shortminus 5}\)\std{\(2\mathrm{e}^{\shortminus 5}\)} & \(5.9\mathrm{e}^{\shortminus 5}\)\std{\(2\mathrm{e}^{\shortminus 5}\)}\\
  Burgers & \(\bm{8.5\mathrm{e}^{\shortminus4}}\)\std{\(3\mathrm{e}^{\shortminus4}\)} & \(1.3\mathrm{e}^{\shortminus3}\)\std{\(6\mathrm{e}^{\shortminus4}\)} & \(1.9\mathrm{e}^{\shortminus3}\)\std{\(7\mathrm{e}^{\shortminus4}\)} & & \(2.6\mathrm{e}^{\shortminus3}\)\std{\(1\mathrm{e}^{\shortminus3}\)} & \(2.7\mathrm{e}^{\shortminus3}\)\std{\(1\mathrm{e}^{\shortminus3}\)} & \(2.8\mathrm{e}^{\shortminus3}\)\std{\(1\mathrm{e}^{\shortminus3}\)} & \(1.0\mathrm{e}^{\shortminus4}\)\std{\(4\mathrm{e}^{\shortminus5}\)} \\\Xhline{1.pt}
    \end{tabularx}
    \label{tab:lwr_metrics}
\end{table}

\newcommand{\Fig}[5]{%
    \input{#2}
    \begin{subfigure}[t]{.3\textwidth}
        \centering
        \tiny
        \begin{tikzpicture}[scale=.45]
              \foreach \y [count=\n] in \mydata {
                \foreach \x [count=\m] in \y {
                  \ifnum\n=\m
                    \node[fill=black, draw=none, minimum size=4.5mm, text=black] at (\m,-\n) {};
                  \else
                    \node[fill=orange!40!red!\x!cyan!50!white, minimum size=4.5mm, text=black, inner sep=0] at (\m,-\n) {\x};
                  \fi
            }
                }

          \draw[thick] (0.5,-2.5) -- (8.5,-2.5);
          \draw[thick, dashed] (5.5,-.5) -- (5.5,-8.5);
          \draw[thick, dashed] (7.5,-.5) -- (7.5,-8.5);

        \ifthenelse{\equal{#4}{true}}{%
            \foreach \a [count=\i] in {NFV$_3^1$,UNFV$_3^1$,GD,LF,EO,ENO,WENO,DG}  {
                \node[inner sep=0, anchor=south] at (\i, -.3) {\rotatebox{90}{\a}};
              }%
        }{}
              \foreach \a [count=\i] in {NFV$_3^1$,UNFV$_3^1$,GD,LF,EO,ENO,WENO,DG} {
                \node[anchor=east] at (.3,-\i) {\rotatebox{0}{\a}};
              }%
        \end{tikzpicture}
        \subcaption{#1}
    \end{subfigure}%
}

\begin{figure}
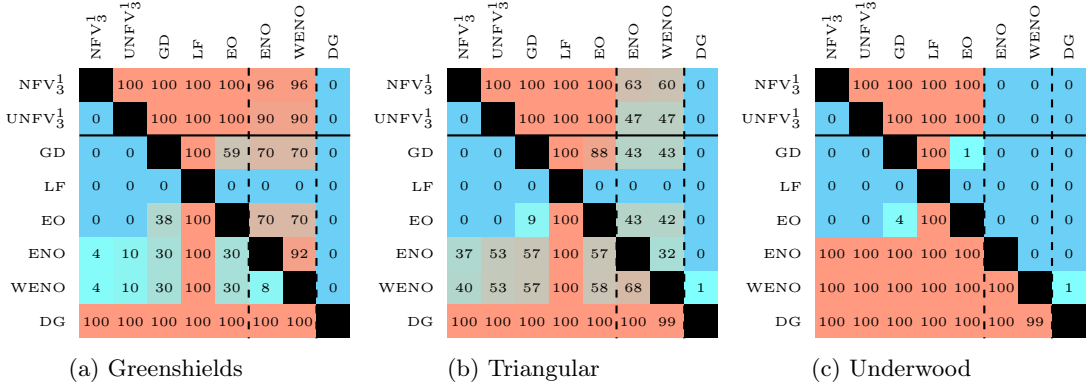

    \centering
    \Fig{Greenshields}{greenshield.tex}{true}{true}{.35} \hfill
    \Fig{Triangular}{triangular.tex}{false}{true}{.3} \hfill
    \Fig{Underwood}{underwood}{false}{true}{.3}
    \caption{\textbf{Comparison of numerical schemes across flow functions.} Each cell shows the proportion of the evaluation set on which the row scheme outperforms the column scheme. DG, the only FEM tested, is rarely beaten. NFV$_3^1$ and UNFV$_3^1$ outperform other first-order schemes and rival higher-order ones, making them strong choices depending on the equation.}
    \label{fig:lwr_winrates}
\end{figure} 
\subsection{Equations} \label{sec:equations}

\textbf{The Lighthill-Whitham-Richards} model~\citep{lighthill1955kinematic, R56}, known as LWR, is a first-order hyperbolic conservation law used to model traffic flow. It is expressed as
\begin{equation}
    \partial_t \rho + \partial_x (\rho v(\rho)) = 0
\end{equation}
ewhere \(\rho\) is the density of the traffic, \(f : \rho \mapsto \rho v(\rho)\) is the flux function and \(v\) is the velocity. The flux function is typically modeled as a concave function of the density. Variations in the underlying velocity function give rise to different traffic flow models. In this work, six different models have been considered: Greenshields'~\citep{greenshields1935study}, Triangular~\citep{geroliminis2008existence}, Triangular skewed~\citep{geroliminis2008existence}, Trapezoidal~\citep{geroliminis2011properties}, Greenberg~\citep{greenberg1959analysis} and Underwood~\citep{Underwood1961}. These models behave \textit{very} differently and should be considered as different equations, as shown in~\Cref{fig:teaser}. Formulations and illustrations of those six models are given in~\Cref{app:lwr}.

\textbf{The inviscid Burgers' equation} is a well-known hyperbolic conservation law used in various domains such as fluid mechanics~\citep{burgers1939math}, non-linear acoustics~\citep{lombard2013fractional}, gas dynamics~\citep{panayotounakos1995closed}, and traffic flow~\citep{musha1978traffic}. We refer the reader to~\citet{cameron2011notes} for a thorough introduction. It is expressed as
\begin{equation}
    \partial_t u + \dfrac{1}{2} \partial_x u^2 = 0.
\end{equation} This equation can be written in the classical form of a conservation law using the flux function \(f: u \mapsto \half\cdot u^2\). Exact solutions to Riemann initial conditions are also known for this problem. Visualization of some solutions, including videos, are available on \linkwebsiteshort{} and in \Cref{fig:teaser}.

While we focus on one-dimension equations in this work, our methods are, at least theoretically, generalizable to the multi-dimensional case of conservation laws~\citep{bressan}. In addition, we expect that our proposed methods can be extended to second-order hyperbolic equations such as ARZ-type models~\citep{aw2000resurrection, zhang2000structural} and shallow water equations~\citep{leveque2002finite}; see \Cref{app:general_PDE} for details.

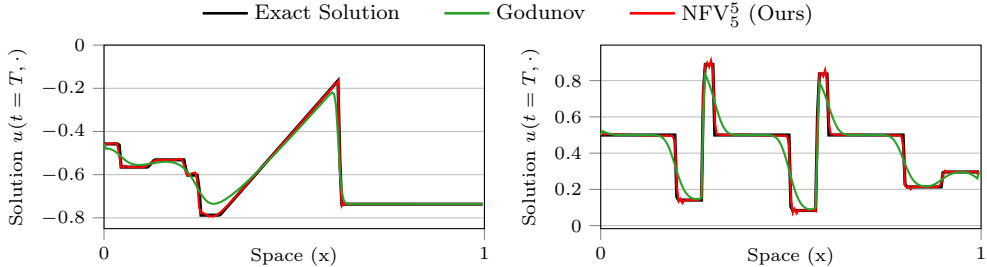
\begin{figure}
    \centering
    \begin{subfigure}{\textwidth}
      \centering
      \begin{tikzpicture}
        \definecolor{darkgray176}{RGB}{176,176,176}
        \definecolor{darkorange25512714}{RGB}{255,0,0}
        \definecolor{forestgreen4416044}{RGB}{44,160,44}
        \definecolor{lightgray204}{RGB}{204,204,204}
        \definecolor{steelblue31119180}{RGB}{100,0,255}
        \definecolor{pltblue}{HTML}{1F77B4}
        \definecolor{pltorange}{HTML}{FF7F0E}
        \definecolor{pltgreen}{HTML}{2CA02C}
        \begin{axis}[
          hide axis,
          legend columns=5,
          legend style={
            draw=none,
            /tikz/every even column/.append style={column sep=0.5cm}
          },
          width=0.8\textwidth,
        ]
          \addplot[line width=1pt,mark=none,color=black] coordinates {(0,0) (0,0)};
          \addlegendentry{\footnotesize Exact Solution};
          \addplot[line width=1pt,mark=none,color=pltgreen] coordinates {(0,0) (0,0)};
          \addlegendentry{\footnotesize Godunov};
          \addplot[line width=1pt,mark=none,color=red] coordinates {(0,0) (0,0)};
          \addlegendentry{\footnotesize NFV\(^5_5\) (Ours)};
        \end{axis}
      \end{tikzpicture}
    \end{subfigure}
    \vspace{-3.8cm}

    \begin{subfigure}[t]{.48\textwidth}
        \begingroup
        \centering
        \begin{tikzpicture}[baseline]
            \scriptsize
            \definecolor{darkgray176}{RGB}{176,176,176}
            \definecolor{darkorange25512714}{RGB}{255,0,0}
            \definecolor{forestgreen4416044}{RGB}{44,160,44}
            \definecolor{lightgray204}{RGB}{204,204,204}
            \definecolor{steelblue31119180}{RGB}{100,0,255}
            
            \begin{axis}[
                legend pos=north west,
                legend cell align={left},
                legend style={fill opacity=0.8, draw opacity=1, text opacity=1, draw=lightgray204},
                tick align=outside,
                tick pos=left,
                width=0.99\textwidth,
                height=.6\textwidth,
                xlabel={Space (x)},
                xmajorgrids,
                ymin=-.85, ymax=0,
                xmin=0-.1, xmax=.1,
                ylabel={Solution \(\displaystyle u(t=T, \cdot)\)},
                ymajorgrids,
                legend columns=1,
                xtick={-.1, .1},
                xticklabels={0,1},
                xlabel style={yshift=10pt},
            ]
            \addplot [very thick, black]
            table[y index=0,  x expr=\coordindex*0.001 - .1] {burgers_lax_hopf.tex};
            
            \addplot [thick,red] 
            table[y index=0, x expr=\coordindex*0.001 - .1] {burgers_ours.tex};

            \addplot [thick,forestgreen4416044]  
            table[y index=0, x expr=\coordindex*0.001 - .1] {burgers_godunov.tex};

            \end{axis}
        \end{tikzpicture}
        \endgroup
    \end{subfigure}
    \begin{subfigure}[t]{.48\textwidth}
        \begingroup
        \centering
        \begin{tikzpicture}[baseline]
            \scriptsize
            \definecolor{darkgray176}{RGB}{176,176,176}
            \definecolor{darkorange25512714}{RGB}{255,0,0}
            \definecolor{forestgreen4416044}{RGB}{44,160,44}
            \definecolor{lightgray204}{RGB}{204,204,204}
            \definecolor{steelblue31119180}{RGB}{100,0,255}
            
            \begin{axis}[
                legend pos=north east,
                legend cell align={left},
                legend style={fill opacity=0.8, draw opacity=1, text opacity=1, draw=lightgray204},
                tick align=outside,
                tick pos=left,
                width=0.99\textwidth,
                height=.6\textwidth,
                xlabel={Space (x)},
                xmajorgrids,
                xmin=-.1, xmax=.1,
                ylabel={Solution \(\displaystyle u(t=T, \cdot)\)},
                ymajorgrids,
                legend columns=1,
                xtick={-.1, .1},
                xticklabels={0,1},
                xlabel style={yshift=10pt},
            ]
            \addplot [very thick, black]
            table[y index=0,  x expr=\coordindex*0.001 - .1] {triangular_lax_hopf.tex};
            
            \addplot [thick, red] 
            table[y index=0, x expr=\coordindex*0.001 - .1] {triangular_ours.tex};

            \addplot [thick, forestgreen4416044]  
            table[y index=0, x expr=\coordindex*0.001 - .1] {triangular_godunov.tex};

            \end{axis}
        \end{tikzpicture}
        \endgroup
    \end{subfigure}
    \caption{Comparison of the final density of the Burgers' equation (left) and LWR triangular equation (right) for NFV\(_5^5\) and the Godunov Scheme. The proposed method displays an excellent approximation of the exact solution, capturing sharp features such as discontinuities and points of non-differentiability. It contains some minor oscillations in the solution, which are not present in the Godunov scheme. The latter, however, fails to capture the discontinuities and points of non-differentiability, offering a very smoothed solution.}
    \label{fig:2d_final_density}
\end{figure} 

\subsection{Datasets}\label{sec:datasets}

Training is performed using solutions derived from Riemann problems, which are initial value problems characterized by piecewise constant initial conditions with a single discontinuity (see~\Cref{fig:lwr_exact_sols} for examples). These problems are fundamental in the study of hyperbolic PDEs and serve as essential test cases for numerical methods. For the scenarios considered in this work, analytical solutions to Riemann problems are available, making supervised learning possible.
Evaluation is performed on a more complicated set of a several hundred initial conditions to assess the model's generalization capabilities. These conditions consist of piecewise constant functions with ten discontinuities, introducing increased complexity. Exact solutions for these test cases are computed using the Lax-Hopf algorithm~\citep{lax1957hyperbolic, claudel2010lax, claudel2010lax2} on a finer grid (see \Cref{app:exp_details}).

\subsection{Results and discussion}

\Cref{tab:lwr_metrics} reports \(L_2\) error for NFV$_3^1$, UNFV$_3^1$, and  baseline methods across the seven benchmark equations. Our models consistently outperform all first-order FV methods, and surpass ENO/WENO schemes on about half of the equations. As expected, the higher-order DG method achieves significantly lower errors. \Cref{tab:2d_metrics} shows that NFV\(_5^5\), while as simple to implement as standard NFV$_3^1$, achieves up to 10x better accuracy, approaching the performance of DG.

\Cref{fig:lwr_winrates} shows the fraction of test cases each method wins. NFV$_3^1$ and UNFV$_3^1$ consistently surpass first-order FV methods. Against ENO/WENO, performance varies: our models outperform on some equations, match on others, and underperform in a few, highlighting the complexity of benchmarking across diverse problem settings. Still, the fact that NFV$_3^1$ and UNFV$_3^1$ consistently do better than first-order methods is seen as a sign that the approach appears to converge well. Specifically, NFV$_3^1$ and UNFV$_3^1$ consistently produce errors bounded by those of Godunov, emphasizing their robustness.

Since all methods use autoregressive prediction, evaluating performance at the final time step provides a good proxy for cumulative error. \Cref{fig:2d_final_density} shows that the prediction of NFV\(_5^5\) closely aligns with the exact solution, with only minor oscillations observed. Notably, NFV\(_5^5\) effectively captures sharp discontinuities with high accuracy without relying on smoothing techniques, which are commonly employed in traditional FV methods to mitigate numerical artifacts.
 
\begin{figure}
    \centering
    \begin{tikzpicture}
    \begin{axis}[
        xlabel={\(\Delta t\)},
        ylabel={\(L_2\) error},
        xlabel style={yshift=6pt},
        grid=both,   
        xmode=log,
        ymode=log,
        x label style={at={(axis description cs:.5,-.1)}},
        legend style={at={(.27,0.84)}, anchor=center},
        legend cell align=left,
        width=.8\textwidth,
        height=.3\textwidth,
        legend style={font=\footnotesize},
        legend columns=3,
        legend pos = north west,
        ymin=0.0013,
        ymax=0.04
    ]
    
    \addplot[red, 
    mark=x,
    mark options={very thick, scale=1.5},
    ] coordinates {
    (0.001, 0.012398871031558099) +- (0.002606770378204459, 0.002606770378204459)
    (0.0007742636826811271, 0.009808386748395601) +- (0.0020948545447002634, 0.0020948545447002634)
    (0.000599484250318941, 0.008000624403867375) +- (0.0018837330042691983, 0.0018837330042691983)
    (0.0004641588833612779, 0.006292124619445425) +- (0.0014867072580761918, 0.0014867072580761918)
    (0.00035938136638046273, 0.004953528186474813) +- (0.0011505203252874792, 0.0011505203252874792)
    (0.0002782559402207124, 0.0040670904715640055) +- (0.0009853362593965048, 0.0009853362593965048)
    (0.0002154434690031884, 0.0035447599931533852) +- (0.0008459611690404069, 0.0008459611690404069)
    (0.0001668100537200059, 0.002947867099663713) +- (0.0007179783456420399, 0.0007179783456420399)
    (0.00012915496650148844, 0.0025506941116867107) +- (0.0005716555331476384, 0.0005716555331476384)
    (9.999999999999998e-05, 0.0022822297822621355) +- (0.0005064092785838558, 0.0005064092785838558)
    };
    \addlegendentry{NFV$_3^1$};
    
    \addplot[blue, 
    mark=x,
    mark options={very thick, scale=1.5},
    ] coordinates {
    (0.001, 0.01751107666113228) +- (0.004354443993719672, 0.004354443993719672)
    (0.0007742636826811271, 0.013397033639360269) +- (0.0027598736739124496, 0.0027598736739124496)
    (0.000599484250318941, 0.010798895899196886) +- (0.0023861777480975507, 0.0023861777480975507)
    (0.0004641588833612779, 0.008338250248229498) +- (0.0019417621218139338, 0.0019417621218139338)
    (0.00035938136638046273, 0.0065911226747773655) +- (0.0016938655926175478, 0.0016938655926175478)
    (0.0002782559402207124, 0.005256896162823991) +- (0.001359921357957374, 0.001359921357957374)
    (0.0002154434690031884, 0.004480195132407675) +- (0.0013006906911855578, 0.0013006906911855578)
    (0.0001668100537200059, 0.003647601879135498) +- (0.0010632675179153542, 0.0010632675179153542)
    (0.00012915496650148844, 0.0030884805502485115) +- (0.0010175122636898468, 0.0010175122636898468)
    (9.999999999999998e-05, 0.0026790538326641154) +- (0.0009695418685616329, 0.0009695418685616329)
    };
    \addlegendentry{UNFV$_3^1$};
    
    \addplot[green!70!black, 
    mark=x,
     mark options={very thick, scale=1.5},
    ] coordinates { 
    (0.001, 0.02184995597091381) +- (0.0043518786623826635, 0.0043518786623826635)
    (0.0007742636826811271, 0.01838306431386583) +- (0.0036746066675479644, 0.0036746066675479644)
    (0.000599484250318941, 0.015653029710674806) +- (0.003349639050582098, 0.003349639050582098)
    (0.0004641588833612779, 0.012793922524939815) +- (0.0027536709411917237, 0.0027536709411917237)
    (0.00035938136638046273, 0.010466471035864865) +- (0.002237362854026388, 0.002237362854026388)
    (0.0002782559402207124, 0.00859585272119339) +- (0.0019355612437412593, 0.0019355612437412593)
    (0.0002154434690031884, 0.00731524856816584) +- (0.0016680487247411687, 0.0016680487247411687)
    (0.0001668100537200059, 0.005937887644113963) +- (0.0013556247170399092, 0.0013556247170399092)
    (0.00012915496650148844, 0.0048693872096709015) +- (0.0010829769972929286, 0.0010829769972929286)
    (9.999999999999998e-05, 0.004004110215715265) +- (0.0008950858175905402, 0.0008950858175905402)
    };
    \addlegendentry{Godunov};
    
    \addplot[name path=upper, draw=none] coordinates {
    (0.001, {0.02184995597091381 + 0.0043518786623826635})
    (0.0007742636826811271, {0.01838306431386583 + 0.0036746066675479644})
    (0.000599484250318941, {0.015653029710674806 + 0.003349639050582098})
    (0.0004641588833612779, {0.012793922524939815 + 0.0027536709411917237})
    (0.00035938136638046273, {0.010466471035864865 + 0.002237362854026388})
    (0.0002782559402207124, {0.00859585272119339 + 0.0019355612437412593})
    (0.0002154434690031884, {0.00731524856816584 + 0.0016680487247411687})
    (0.0001668100537200059, {0.005937887644113963 + 0.0013556247170399092})
    (0.00012915496650148844, {0.0048693872096709015 + 0.0010829769972929286})
    (9.999999999999998e-05, {0.004004110215715265 + 0.0008950858175905402})
    };
    \addplot[name path=lower, draw=none] coordinates {
    (0.001, {0.02184995597091381 - 0.0043518786623826635})
    (0.0007742636826811271, {0.01838306431386583 - 0.0036746066675479644})
    (0.000599484250318941, {0.015653029710674806 - 0.003349639050582098})
    (0.0004641588833612779, {0.012793922524939815 - 0.0027536709411917237})
    (0.00035938136638046273, {0.010466471035864865 - 0.002237362854026388})
    (0.0002782559402207124, {0.00859585272119339 - 0.0019355612437412593})
    (0.0002154434690031884, {0.00731524856816584 - 0.0016680487247411687})
    (0.0001668100537200059, {0.005937887644113963 - 0.0013556247170399092})
    (0.00012915496650148844, {0.0048693872096709015 - 0.0010829769972929286})
    (9.999999999999998e-05, {0.004004110215715265 - 0.0008950858175905402})
    };
    \addplot[green!70!black, forget plot, fill opacity=0.15] fill between[of=upper and lower];
    
    \addplot[name path=upper, draw=none] coordinates {
    (0.001, {0.012398871031558099 + 0.002606770378204459})
    (0.0007742636826811271, {0.009808386748395601 + 0.0020948545447002634})
    (0.000599484250318941, {0.008000624403867375 + 0.0018837330042691983})
    (0.0004641588833612779, {0.006292124619445425 + 0.0014867072580761918})
    (0.00035938136638046273, {0.004953528186474813 + 0.0011505203252874792})
    (0.0002782559402207124, {0.0040670904715640055 + 0.0009853362593965048})
    (0.0002154434690031884, {0.0035447599931533852 + 0.0008459611690404069})
    (0.0001668100537200059, {0.002947867099663713 + 0.0007179783456420399})
    (0.00012915496650148844, {0.0025506941116867107 + 0.0005716555331476384})
    (9.999999999999998e-05, {0.0022822297822621355 + 0.0005064092785838558})
    };
    \addplot[name path=lower, draw=none] coordinates {
    (0.001, {0.012398871031558099 - 0.002606770378204459})
    (0.0007742636826811271, {0.009808386748395601 - 0.0020948545447002634})
    (0.000599484250318941, {0.008000624403867375 - 0.0018837330042691983})
    (0.0004641588833612779, {0.006292124619445425 - 0.0014867072580761918})
    (0.00035938136638046273, {0.004953528186474813 - 0.0011505203252874792})
    (0.0002782559402207124, {0.0040670904715640055 - 0.0009853362593965048})
    (0.0002154434690031884, {0.0035447599931533852 - 0.0008459611690404069})
    (0.0001668100537200059, {0.002947867099663713 - 0.0007179783456420399})
    (0.00012915496650148844, {0.0025506941116867107 - 0.0005716555331476384})
    (9.999999999999998e-05, {0.0022822297822621355 - 0.0005064092785838558})
    };
    \addplot[red, forget plot, fill opacity=0.15] fill between[of=upper and lower];
    
    \addplot[name path=upper, draw=none] coordinates {
    (0.001, {0.01751107666113228 + 0.004354443993719672})
    (0.0007742636826811271, {0.013397033639360269 + 0.0027598736739124496})
    (0.000599484250318941, {0.010798895899196886 + 0.0023861777480975507})
    (0.0004641588833612779, {0.008338250248229498 + 0.0019417621218139338})
    (0.00035938136638046273, {0.0065911226747773655 + 0.0016938655926175478})
    (0.0002782559402207124, {0.005256896162823991 + 0.001359921357957374})
    (0.0002154434690031884, {0.004480195132407675 + 0.0013006906911855578})
    (0.0001668100537200059, {0.003647601879135498 + 0.0010632675179153542})
    (0.00012915496650148844, {0.0030884805502485115 + 0.0010175122636898468})
    (9.999999999999998e-05, {0.0026790538326641154 + 0.0009695418685616329})
    };
    \addplot[name path=lower, draw=none] coordinates {
    (0.001, {0.01751107666113228 - 0.004354443993719672})
    (0.0007742636826811271, {0.013397033639360269 - 0.0027598736739124496})
    (0.000599484250318941, {0.010798895899196886 - 0.0023861777480975507})
    (0.0004641588833612779, {0.008338250248229498 - 0.0019417621218139338})
    (0.00035938136638046273, {0.0065911226747773655 - 0.0016938655926175478})
    (0.0002782559402207124, {0.005256896162823991 - 0.001359921357957374})
    (0.0002154434690031884, {0.004480195132407675 - 0.0013006906911855578})
    (0.0001668100537200059, {0.003647601879135498 - 0.0010632675179153542})
    (0.00012915496650148844, {0.0030884805502485115 - 0.0010175122636898468})
    (9.999999999999998e-05, {0.0026790538326641154 - 0.0009695418685616329})
    };
    \addplot[blue, forget plot, fill opacity=0.15] fill between[of=upper and lower];

    \addplot[black, dashed, thick] coordinates {(1e-3, 1e-5) (1e-3, 1)};
    
    \end{axis}
    \end{tikzpicture}
    
    \caption{\textbf{Convergence plots on Greenshields' flux.} The \(L_2\) error is computed against the exact solution on the evaluation set for different mesh discretizations. We report both error average and standard deviation, on a log-log scale. The dashed vertical line illustrates the discretization on which NFV$_3^1$ and UNFV$_3^1$ were trained; the models generalize to smaller discretizations. The ratio \(\Delta t / \Delta x = 0.1\) remains constant as the mesh is refined.}
    \label{fig:lwr_convergence_plot}
    \end{figure}
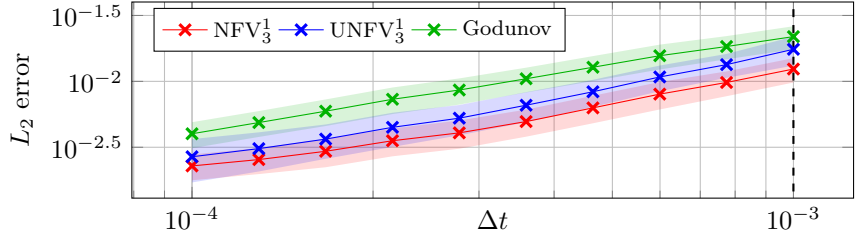

A strong strength of classical numerical schemes is their proven convergence when the discretization is refined. While this is not proven for our method,~\Cref{fig:lwr_convergence_plot} shows that both NFV$_3^1$ and UNFV$_3^1$ can consistently perform better than Godunov, a scheme proven to converge, even when the grid is refined. Our schemes are thus expected to converge. The log-log plot exhibits a linear trends, which suggests that the rate of convergence could be polynomial.

\begin{table}
    \definecolor{deepcarmine}{rgb}{0.66, 0.13, 0.24}
    \caption{Evaluation of NFV\(_5^5\) using piecewise constant initial conditions. Error is reported in \(L_2\) norm. NFV\(\bm{^5_5}\) achieve outstanding performance, gaining up to an order of magnitude improvement compared to Godunov and WENO. Its performance is close to DG, while keeping the implementation simplicity of a finite volume method and the computational complexity of NVF.}
    \centering
    \small
        \begin{tabularx}{\textwidth}{cCCCCC}
        \Xhline{1.pt} 
        & Godunov & WENO & \textbf{NFV}$\bm{_3^1}$ & \textbf{NFV}\(\bm{^5_5}\) & DG\\
        \Xhline{1.pt} 
        Burgers' & \(1.8\color{deepcarmine}\mathrm{e}^{-3}\)\std{\(6\mathrm{e}^{-4}\)} & \(2.6\color{deepcarmine}\mathrm{e}^{-3}\)\std{\(1\mathrm{e}^{-3}\)} & \(8.3\color{deepcarmine}\mathrm{e}^{-4}\)\std{\(3\mathrm{e}^{-4}\)} & \(2.2\color{deepcarmine}\mathrm{e}^{-4}\)\std{\(1\mathrm{e}^{-4}\)} & \(\bm{1.0}\color{deepcarmine}\bm{\mathrm{e}^{-4}}\)\std{\(4\mathrm{e}^{-5}\)}\\
        Greenshields & \(4.1\color{deepcarmine}\mathrm{e}^{-4}\)\std{\(1\mathrm{e}^{-4}\)} & \(6.9\color{deepcarmine}\mathrm{e}^{-4}\)\std{\(4\mathrm{e}^{-4}\)}& \(1.2\color{deepcarmine}\mathrm{e}^{-4}\)\std{\(4\mathrm{e}^{-5}\)} & \(4.6\color{deepcarmine}\mathrm{e}^{-5}\)\std{\(3\mathrm{e}^{-5}\)} & \(\bm{4.2}\color{deepcarmine}\bm{\mathrm{e}^{-5}}\)\std{\(2\mathrm{e}^{-5}\)}\\
        Triangular & \(2.2\color{deepcarmine}\mathrm{e}^{-3}\)\std{\(1\mathrm{e}^{-3}\)} & \(2.0\color{deepcarmine}\mathrm{e}^{-3}\)\std{\(2\mathrm{e}^{-3}\)} & \(1.3\color{deepcarmine}\mathrm{e}^{-3}\)\std{\(6\mathrm{e}^{-4}\)} & \(2.9\color{deepcarmine}\mathrm{e}^{-4}\)\std{\(2\mathrm{e}^{-4}\)} & \(\bm{2.7}\color{deepcarmine}\bm{\mathrm{e}^{-4}}\)\std{\(1\mathrm{e}^{-4}\)}\\
        \Xhline{1.pt} 
    \end{tabularx}
	\label{tab:2d_metrics}
\end{table}

\section{Application: Modeling Large-Scale Experimental Field Data using NFV}\label{sec:i24}

We apply the proposed NFV method to large-scale traffic field data collected on Interstate 24 (I-24) in Tennessee, USA, using the I-24 MOTION infrastructure~\citep{gloudemans202324,gloudemans2023so}, which enables high-resolution vehicle trajectory collection and constitutes the most extensive publicly available traffic dataset to date. Rather than predicting traffic speed, we focus on modeling traffic density, which is more directly tied to conservation laws and often exhibits sharp transitions that are challenging to capture. Although conservation of mass is not strictly satisfied in highway traffic data due to merges, exits, and incidents, it serves as a strong inductive bias. We show that NFV achieves superior predictive accuracy compared to classical numerical schemes. Moreover, incorporating the PDE structure leads to substantially more stable training, particularly in data-scarce regimes. These findings suggest that our approach can enhance the accuracy and efficiency of traffic simulations, thereby contributing to better-informed decision-making in urban planning and traffic management.

\subsection{Dataset} 
\label{sec:i24_dataset}

We evaluate our method on the I-24 MOTION dataset~\citep{gloudemans202324}, which provides high-resolution vehicle trajectories collected on a four-mile stretch of Interstate 24 (mile markers 58.7 to 62.7) near Nashville, Tennessee. The data is captured by a network of high-definition cameras mounted along the highway as part of the I-24 MOTION infrastructure, leading to intricate trajectory data as illustrated in \Cref{fig:i24_teaser}. Vehicle trajectories are reconstructed using a computer vision and data association pipeline~\citep{wang2022automatic}, resulting in high-fidelity, though inherently noisy, field data. 

The dataset consists of 10 days of vehicle trajectory data, collected during the morning rush hour (6:00 AM to 10:00 AM) over the 4-mile segment. From the raw trajectory data, we construct spatiotemporal vehicle density fields by aggregating vehicle counts over fixed spatial cells. Details of the data cleaning, processing, and preparation are provided in \Cref{app:i24_data}. Visualization of the resulting density fields is shown in \Cref{fig:i24_data}.

\subsection{Results and Discussion}
\label{sec:i24_part1}

We compare NFV to numerical schemes using the flux functions from \Cref{app:lwr}. These functions, each defined by a few parameters, were calibrated via optimization to minimize the Godunov scheme’s prediction error on the training set. The search ranges were intentionally broad, prioritizing predictive performance over physical plausibility to ensure a fair comparison. We evaluate three NFV variants of increasing capacity: $\text{NFV}_3^1$, $\text{NFV}_5^5$, and $\text{NFV}_{11}^{11}$ (training details can be found in \Cref{app:exp_details}), to assess how well they generalize and capture complex field dynamics.

\begingroup
\setlength{\tabcolsep}{1pt}
\begin{table}
    \centering
    \caption{\textbf{Improvements of NFV at different scales against numerical methods with fitted flow functions on field data.} The reported metrics include L1 error (\(\mean(|u-\hat u|)\)), L2 error (\(\mean((u-\hat u)^2)\)), and relative error (\(\mean(|u-\hat u|/|\max\{\varepsilon, u\}|)\)). The larger the input size of NFV, the better the performance. $\text{NFV}_3^1$ outperforms all calibrated Godunov fits, despite having the same input size and underlying structure.}
    \resizebox{0.95\textwidth}{!}{
    \begin{tabularx}{\linewidth}{cCCCCCCCC}
        \hline
         \Xhline{1.pt} 
         & \multicolumn{5}{c}{\textbf{Calibrated numerical schemes (Godunov)}} & \multicolumn{3}{c}{\textbf{NFV (Ours)}} \\
         \Xhline{1.pt} 
          & \footnotesize Greenshields & \footnotesize Triangular & \footnotesize Trapezoidal & \footnotesize Greenberg & \footnotesize Underwood & $\text{NFV}_{3}^{1}$ & $\text{NFV}_{5}^{5}$ & $\text{NFV}_{11}^{11}$ \\
         \Xhline{1.pt}
         {L1} & \( 6.05 \mathrm{e}^{-2} \) & \( 2.77 \mathrm{e}^{-2} \) & \( 2.73 \mathrm{e}^{-2} \) & \( 2.79 \mathrm{e}^{-2} \) & \( 4.98 \mathrm{e}^{-2} \) & \( \bm{2.37 \mathrm{e}^{-2}} \) & \( 2.31 \mathrm{e}^{-2} \) & \( \bm{2.02 \mathrm{e}^{-2}} \) \\
         {L2} & \( 1.93 \mathrm{e}^{-1} \) & \( 1.31 \mathrm{e}^{-1} \) & \( 1.30 \mathrm{e}^{-1} \) & \( 1.33 \mathrm{e}^{-1} \) & \( 1.81 \mathrm{e}^{-1} \) & \( \bm{1.23 \mathrm{e}^{-1}} \) & \( 1.21 \mathrm{e}^{-1} \) & \( \bm{1.09 \mathrm{e}^{-1}} \) \\
         {Rel.} & \( 5.04 \mathrm{e}^{-1} \) & \( 3.83 \mathrm{e}^{-1} \) & \( 3.74 \mathrm{e}^{-1} \) & \( 3.75 \mathrm{e}^{-1} \) & \( 5.45 \mathrm{e}^{-1} \) & \( \bm{3.57 \mathrm{e}^{-1}} \) & \( 3.51 \mathrm{e}^{-1} \) & \( \bm{2.83 \mathrm{e}^{-1}} \) \\
         \Xhline{1.pt} 
    \end{tabularx}
    }
    \label{tab:i24_part1_metrics_table}
\end{table}
\endgroup 

\begin{figure}
	\centering
    \begin{subfigure}[t]{.49\textwidth}
        \begin{tikzpicture}
            \node[anchor=south west, inner sep=0, draw=black, line width=1.5pt] (img) at (0,0) {\includegraphics[width=0.99\textwidth]{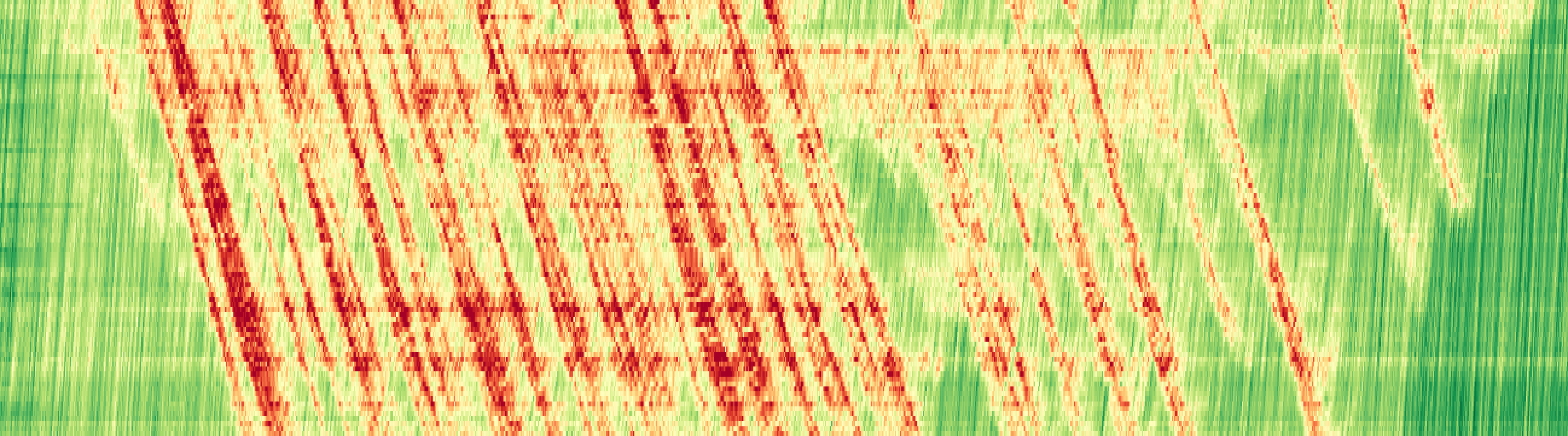}};
        \end{tikzpicture}
    \end{subfigure}
    \hfill
    \begin{subfigure}[t]{.49\textwidth}
        \begin{tikzpicture}
            \node[anchor=south west, inner sep=0, draw=black, line width=1.5pt] (img) at (0,0) {\includegraphics[width=0.99\textwidth]{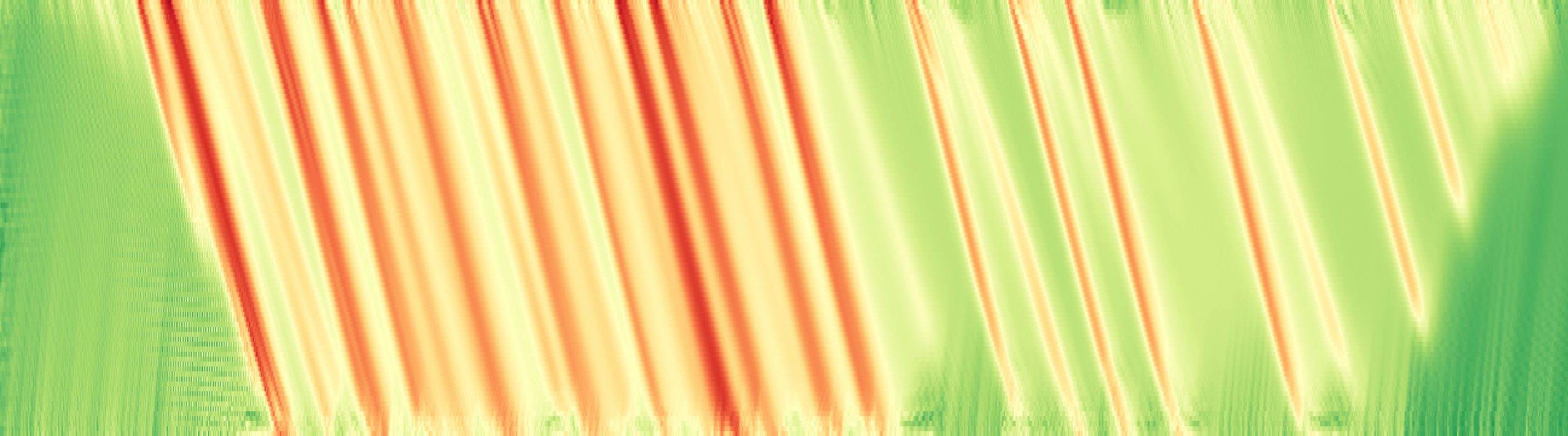}};
        \end{tikzpicture}    
    \end{subfigure}

    \caption{\textbf{Autoregressive prediction of $\text{NFV}_{11}^{11}$ (right) compared to the ground truth (left).} Full results are shown in \Cref{fig:i24_part1_all_predictions}. See \Cref{sec:how_to_read} for how to read the heatmaps.}
    \label{fig:i24_part1_best_predictions}
\end{figure}

\Cref{tab:i24_part1_metrics_table} shows that all NFV models outperform the five tuned Godunov schemes, with performance improving as input size increases. This trend matches what was seen on synthetic data (\Cref{sec:experiments}). Despite training on just one hour of data, NFV predicts nearly four hours of traffic evolution autoregressively (\Cref{fig:i24_part1_best_predictions}). While performance degrades in out-of-distribution zones (e.g., dark green regions unseen during training), the models still capture key wave patterns with high fidelity. Larger stencils help smooth out noise and improve accuracy, as seen in \Cref{fig:i24_part1_all_predictions}.

We further evaluate generalization on 7 other days. As shown in \Cref{fig:i24_part1_generalization_all_predictions} and \Cref{tab:i24_part1_metrics_generalization_table}, $\text{NFV}_{11}^{11}$ consistently outperforms the best Godunov scheme on the evaluation set, even though both perform similarly on the training day. Indeed, although far from perfect, it is able to capture the evolution of free-flow traffic (dark green) with much greater accuracy, allowing it to successfully capture the end of congestion waves (red).
NFV scales naturally with capacity: $\text{NFV}_{11}^{11}$ adds only 1728 parameters over $\text{NFV}_{3}^{1}$ but achieves significantly better accuracy with similar runtime and memory usage, unlike hand-crafted schemes, which significantly grow in complexity (see for example \Cref{sec:fvm}). 

\begingroup
\begin{table}[h]
    \centering
    \caption{\textbf{Generalization of NFV against Godunov on 7 days of I-24 data never seen during training.} As in \Cref{tab:i24_part1_metrics_table}, we report mean and standard deviation of L1, L2 and relative errors.}
    \begin{tabular}{cccc}
      \Xhline{1.pt}
      & L1 error & L2 error & Relative error \\
      \Xhline{1.pt}
      Godunov & \( 1.56 \mathrm{e}^{-1} \pm 2.02 \mathrm{e}^{-2} \) 
              & \( 3.74 \mathrm{e}^{-2} \pm 8.25 \mathrm{e}^{-3} \) 
              & \( 6.26 \mathrm{e}^{-1} \pm 2.58 \mathrm{e}^{-1} \) \\
      $\text{NFV}_{11}^{11}$ & \( \bm{1.12 \mathrm{e}^{-1} \pm 7.39 \mathrm{e}^{-3}} \) 
                             & \( \bm{2.20 \mathrm{e}^{-2} \pm 2.59 \mathrm{e}^{-3}} \) 
                             & \( \bm{3.59 \mathrm{e}^{-1} \pm 7.58 \mathrm{e}^{-2}} \) \\
      \Xhline{1.pt}
  \end{tabular}
    \label{tab:i24_part1_metrics_generalization_table}
\end{table}
\endgroup 

 \section{Conclusion}
\label{sec:conclusion}

We introduced a neural network-based framework for solving conservation laws that significantly improves accuracy over classical FV methods, while retaining the simplicity and structure of first-order schemes. Despite these empirical gains, a fundamental limitation remains: NFV and UNFV currently lack the rigorous theoretical guarantees, such as stability and convergence, established for traditional numerical methods. Nonetheless, the proposed approach demonstrates strong practical performance across diverse settings.
Looking ahead, several important research directions warrant further exploration. First, establishing theoretical foundations for neural network-based solvers, particularly with respect to convergence and the preservation of physical constraints, is essential. Second, extending the methodology to more complex PDEs and high-dimensional domains will be critical for assessing its scalability and generalizability. Finally, hybrid strategies that integrate data-driven learning with classical numerical techniques may offer enhanced robustness and accuracy, bridging the gap between theoretical rigor and practical flexibility.

\textbf{Broader impacts:} The proposed NFV framework has the potential to enhance simulations in domains such as traffic flow, environmental modeling, and fluid dynamics, thereby supporting more informed decision-making in urban planning and infrastructure development. However, reliance on data-driven models without rigorous validation may lead to inaccuracies, potentially resulting in misguided policies or designs. To mitigate such risks, we advocate for thorough validation against empirical data and collaboration with domain experts to ensure responsible application of the technology.

\FloatBarrier

\bibliographystyle{unsrtnat}
\bibliography{references}

\newpage
\appendix

\section{Finite Volume Methods}\label{sec:fvm}

Several finite volume-based numerical schemes are studied in this work. They include the following common classical first-order schemes:

\paragraph{Godunov~\citep{G59}.}
\begin{equation*}
    \forall i,n \quad \hat F_{i-\half}^n = \left\{
        \begin{array}{ll}
            \underset{[u^n_{i-1}, u^n_{i}]}{\min}f & \text{if } u^n_{i-1} \leq u^n_{i}\\
            \underset{[u^n_{i}, u^n_{i-1}]}{\max}f & \text{if } u^n_{i-1} > u^n_{i}
        \end{array}
        \right.
\end{equation*}

\paragraph{Lax-Friedrichs~\citep{lax1954initial}.}
\begin{equation*}
    \forall i,n \quad \hat F_{i-\half}^n = \frac{1}{2}\left(f(u^n_i) + f(u^n_{i-1})\right) - \frac{1}{2}\dfrac{\Delta x}{\Delta t} \times |u^n_i - u^n_{i-1}|.
\end{equation*}

\paragraph{Engquist-Osher~\citep{engquistOsher1981}.}
\begin{equation*}
    \forall i,n \quad \hat F_{i-\half}^n(u^n_{i-1}, u^n_{i}) = \frac{1}{2}\left(f(u^n_i) + f(u^n_{i-1})\right) - \frac{1}{2} \int_{u^n_{i-1}}^{u^n_{i}}\left|f'\right|.
\end{equation*}

Additionally, higher-order schemes such as the \textbf{Essentially Non-Oscillatory (ENO) method~\citep{shu1999high}} and the \textbf{Weighted Essentially Non-Oscillatory (WENO) method~\citep{liu1994weighted}} are considered. The main idea in these methods is that by considering more stencils, one can expect to increase the accuracy of approximation of the solution. 

For the ENO scheme, we consider the semi-discrete form of 
\begin{equation}
    \partial_t u_i = -\frac{1}{\Delta x} \Big(\hat F_{i + 1/2} - \hat F_{i - 1/2} \Big).
\end{equation}
Using the Lax-Friedrichs Splitting technique, we have 
\begin{equation}
    f(u) = f^+(u) + f^-(u) , \quad f^{\pm}(u) = \frac 12 (f(u) \pm \alpha u),
\end{equation}
where $\alpha = \max \lvert{f'(u)}\rvert$ is the maximum wave speed. 
The key point in the ENO scheme is the high-order upwind interpolation of $f^+$
 and $f^-$ based on the smoothest stencils. For instance, for the 2-stencil ENO scheme, the procedure is as follows:
 \begin{enumerate}
  \item Evaluate the smoothness indicators:
  \[
    \delta_{-} = |f_{i}^{+} - f_{i-1}^{+}|, \quad \delta_{+} = |f_{i+1}^{+} - f_{i}^{+}|
  \]
  
  \item Select the stencil that minimizes the smoothness indicator:
  \begin{itemize}
    \item If $\delta_{+} < \delta_{-}$, choose the stencil $\{f_{i}^{+}, f_{i+1}^{+}\}$.
    \item Otherwise, choose the stencil $\{f_{i-1}^{+}, f_{i}^{+}\}$.
  \end{itemize}
  
  \item Perform linear interpolation to compute the numerical flux:
  \[
    \hat{f}_{i+\frac{1}{2}}^{+} = f_{i}^{+} + \frac{1}{2} \delta^{+}
  \]
  where $\delta^{+}$ is the difference between the selected stencil elements.
\end{enumerate}

A similar approach is applied to compute $\hat{f}_{i+\frac{1}{2}}^{-}$ using the right-biased stencil.

The final numerical flux at the interface is obtained by combining the positive and negative parts:
\[
  \hat{f}_{i+\frac{1}{2}} = \hat{f}_{i+\frac{1}{2}}^{+} + \hat{f}_{i+\frac{1}{2}}^{-}
\]

In this work, we have used a 3-stencil scheme for ENO. 

The WENO scheme follows the same idea as ENO by using specific weights in defining $\hat{f}^+_{i + 1/2}$, rather than explicit conditions. In this work, we use the 5-stencil WENO scheme.  
 \section{Variants of LWR}
\label{app:lwr}

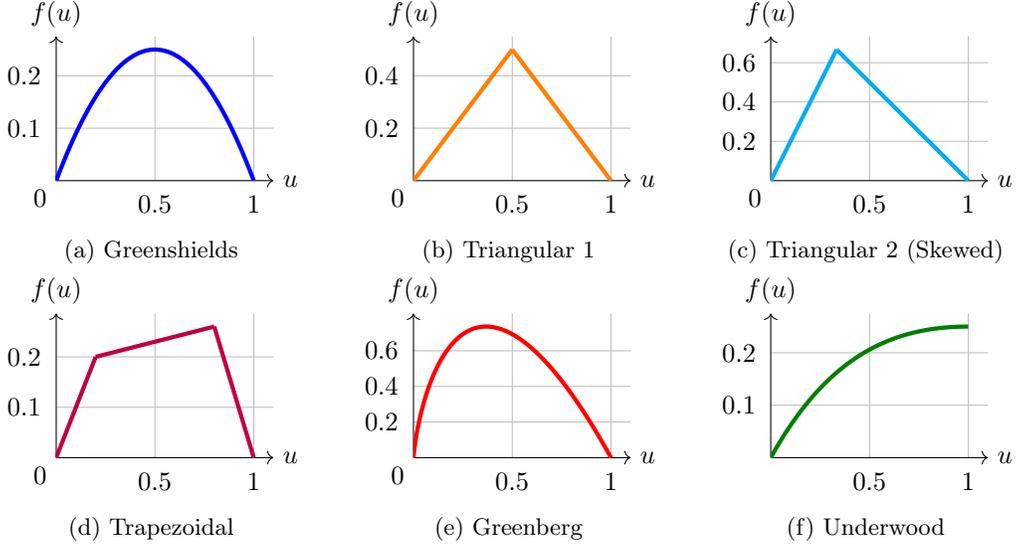
\begin{figure}
    \centering
    \begin{subfigure}[t]{.32\textwidth}
        \centering
        \begin{tikzpicture}
            \begin{axis}[
                axis lines=middle,
                width=\textwidth, height=3.5cm,
                xlabel={\(u\)},
                ylabel={\(f(u)\)},
                xmax=1,
                xtick={0,0.5,1},
                xticklabels={0,0.5,1},
                ylabel style={anchor=south}, 
                xlabel style={anchor=west},
                grid=major,
                samples=200,
                ymin=0,
                enlargelimits={upper},
                axis line style={->},
                tick style={draw=none}
            ]
              \addplot[ultra thick, smooth, color=blue, domain=0:1] {x*(1-x)};
            \end{axis}
            \node at (0,0) [anchor=north east] {0}; 
        \end{tikzpicture}
        \caption{Greenshields}
    \end{subfigure}
    \hfill
    \begin{subfigure}[t]{.32\textwidth}
        \centering
        \begin{tikzpicture}
            \begin{axis}[
                axis lines=middle,
                width=\textwidth, height=3.5cm,
                xlabel={\(u\)},
                ylabel={\(f(u)\)},
                xmax=1,
                xtick={0,0.5,1},
                xticklabels={0,0.5,1},
                ylabel style={anchor=south}, 
                xlabel style={anchor=west},
                grid=major,
                samples=200,
                ymin=0,
                enlargelimits={upper},
                axis line style={->},
                tick style={draw=none}
            ]
                \addplot[ultra thick, smooth, color=orange, domain=0.:.5] {x};
                \addplot[ultra thick, smooth, color=orange, domain=.5:1., forget plot] {1-x};
            \end{axis}
            \node at (0,0) [anchor=north east] {0}; 
        \end{tikzpicture}
        \caption{Triangular 1}
    \end{subfigure}
    \hfill
    \begin{subfigure}[t]{.32\textwidth}
        \centering
        \begin{tikzpicture}
            \begin{axis}[
                axis lines=middle,
                width=\textwidth, height=3.5cm,
                xlabel={\(u\)},
                ylabel={\(f(u)\)},
                xmax=1,
                xtick={0,0.5,1},
                xticklabels={0,0.5,1},
                ylabel style={anchor=south}, 
                xlabel style={anchor=west},
                grid=major,
                samples=200,
                ymin=0,
                enlargelimits={upper},
                axis line style={->},
                tick style={draw=none}
            ]
                \addplot[ultra thick, smooth, color=cyan, domain=0.:1/3] {2*x};
                \addplot[ultra thick, smooth, color=cyan, domain=1/3:1.] {1-x};
            \end{axis}
            \node at (0,0) [anchor=north east] {0}; 
        \end{tikzpicture}
        \caption{Triangular 2 (Skewed)}
    \end{subfigure}
    \vspace{0.2cm}
    \begin{subfigure}[t]{.32\textwidth}
        \centering
        \begin{tikzpicture}
            \begin{axis}[
                axis lines=middle,
                width=\textwidth, height=3.5cm,
                xlabel={\(u\)},
                ylabel={\(f(u)\)},
                xmax=1,
                xtick={0,0.5,1},
                xticklabels={0,0.5,1},
                ylabel style={anchor=south}, 
                xlabel style={anchor=west},
                grid=major,
                samples=200,
                ymin=0,
                enlargelimits={upper},
                axis line style={->},
                tick style={draw=none}
            ]
                \addplot[ultra thick, smooth, color=purple, domain=0:.2] {x};
                \addplot[ultra thick, smooth, color=purple, domain=.2:.8, forget plot] {.1*(x-.2) + .2};
                \addplot[ultra thick, smooth, color=purple, domain=.8:1, forget plot] {1.3 - 1.3*x};
            \end{axis}
            \node at (0,0) [anchor=north east] {0}; 
        \end{tikzpicture}
        \caption{Trapezoidal}
    \end{subfigure}
    \hfill
    \begin{subfigure}[t]{.32\textwidth}
        \centering
        \begin{tikzpicture}
            \begin{axis}[
                axis lines=middle,
                width=\textwidth, height=3.5cm,
                xlabel={\(u\)},
                ylabel={\(f(u)\)},
                xmax=1,
                xtick={0,0.5,1},
                xticklabels={0,0.5,1},
                ytick={0,0.2,0.4,0.6},
                yticklabels={0,0.2,0.4,0.6},
                ylabel style={anchor=south}, 
                xlabel style={anchor=west},
                grid=major,
                samples=200,
                ymin=0,
                enlargelimits={upper},
                axis line style={->},
                tick style={draw=none}
            ]
                \addplot[ultra thick, smooth, color=red, domain=0:1] {-2*x*ln(x)};
            \end{axis}
            \node at (0,0) [anchor=north east] {0}; 
        \end{tikzpicture}
        \caption{Greenberg}
    \end{subfigure}
    \hfill
    \begin{subfigure}[t]{.32\textwidth}
        \centering
        \begin{tikzpicture}
            \begin{axis}[
                axis lines=middle,
                width=\textwidth, height=3.5cm,
                xlabel={\(u\)},
                ylabel={\(f(u)\)},
                xmax=1,
                xtick={0,0.5,1},
                xticklabels={0,0.5,1},
                ylabel style={anchor=south}, 
                xlabel style={anchor=west},
                grid=major,
                samples=200,
                ymin=0,
                enlargelimits={upper},
                axis line style={->},
                tick style={draw=none}
            ]
                \addplot[ultra thick, smooth, color=green!50!black, domain=0.:1.] {.25*exp(1 - x)*x};
            \end{axis}
        \end{tikzpicture}
        \caption{Underwood}
    \end{subfigure}
    \caption{\textbf{Flow models for LWR.} We consider six different variants of the LWR PDE with the flows illustrated here, each mapping road density (veh/m) to traffic flow (veh/s).}
  \label{fig:flows}
  \end{figure} 

We consider six different LWR PDEs variants, each consisting of a different fundamental diagram, illustrated in \Cref{fig:flows}. All of the considered flows are concave continuous mappings from $[0, \rho_{\max}]$ to $\mathbb{R}_+$, where $\rho_{\max}$ is the maximum density, with the exception of the Greenberg flow whose domain is $(0, \rho_{\max}]$. The critical density $\rho_c$ denotes the density at which the flow is maximized, i.e. $\rho_c = \arg\max_{\rho \in [0, \rho_{\max}]} f(\rho)$. The following introduces the six flow models we consider in this work, each time detailing the flow's parameters, the parameter values we use in \Cref{sec:experiments} (in parentheses), and the flow's definition. Note that we consider normalized parameter values lying between 0 and 1 for the most part.

\paragraph{Greenshields.} Parameters: free-flow speed $v_{\max}$ ($1$ m/s), maximum density $\rho_{\max}$ ($1$ veh/m).
\[ f(\rho) = v_{\max} \rho \left(1 - \frac{\rho}{\rho_{\max}}\right) \]

\paragraph{Triangular 1 (symmetrical).} Parameters: free-flow speed $v_{\max}$ ($1$ m/s), critical density $\rho_c$ ($0.5$ veh/m), maximum density $\rho_{\max}$ ($1$ veh/m), wave propagation speed ($-1$ m/s).
\[  f(\rho) = \begin{cases}
    v_{\max} \rho & \text{if } \rho < \rho_c \\
    w (\rho - \rho_{\max}) & \text{if } \rho \geq \rho_c
\end{cases} \]

\paragraph{Triangular 2 (skewed).} A non-symmetric variant of the Triangular flow, with parameters $v_{\max} = 2$ m/s, $\rho_c = \nicefrac{1}{3}$ veh/m, and $w = -1$ m/s.

\paragraph{Trapezoidal.} Parameters: free-flow speed $v_{\max}$ ($1$ m/s), first density cusp $\rho_1$ ($0.2$ veh/m), second density cusp $\rho_2$ ($0.8$ veh/m), maximum density $\rho_{\max}$ ($1$ veh/m), wave propagation speed ($-1.5$ m/s).
\[  f(\rho) = \begin{cases}
    v_{\max} \rho & \text{if } \rho < \rho_1 \\
    (w (\rho_2 - \rho_{\max}) - v_{\max} \rho_1) \displaystyle \frac{\rho - \rho_1}{\rho_2 - \rho_1} + v_{\max} \rho_1 & \text{if } \rho_1 \leq \rho \leq \rho_2 \\
    w (\rho - \rho_{\max}) & \text{if } \rho > \rho_2
\end{cases} \]

\paragraph{Greenberg.} Parameters: maximum density $\rho_{\max}$ ($1$ veh/m), coefficient $c_0$ ($2$). 
\[ f(\rho) = c_0 \rho \log \left( \rho_{\max} / \rho \right) \]

\paragraph{Underwood.} Parameters: maximum density $\rho_{\max}$ ($1$ veh/m), coefficients $c_1$ ($0.25$) and $c_2$ ($1$). 
\[ f(\rho) = c_1 \rho \exp \left( 1 - c_2 \rho \right) \]

\begin{figure}
    \centering
    \begin{subfigure}[t]{.30\textwidth}
        \centering
        \begin{tikzpicture}[scale=.9, clip=false]
            \node[anchor=south west, inner sep=0] (img) at (0,0) {\includegraphics[height=1.9cm]{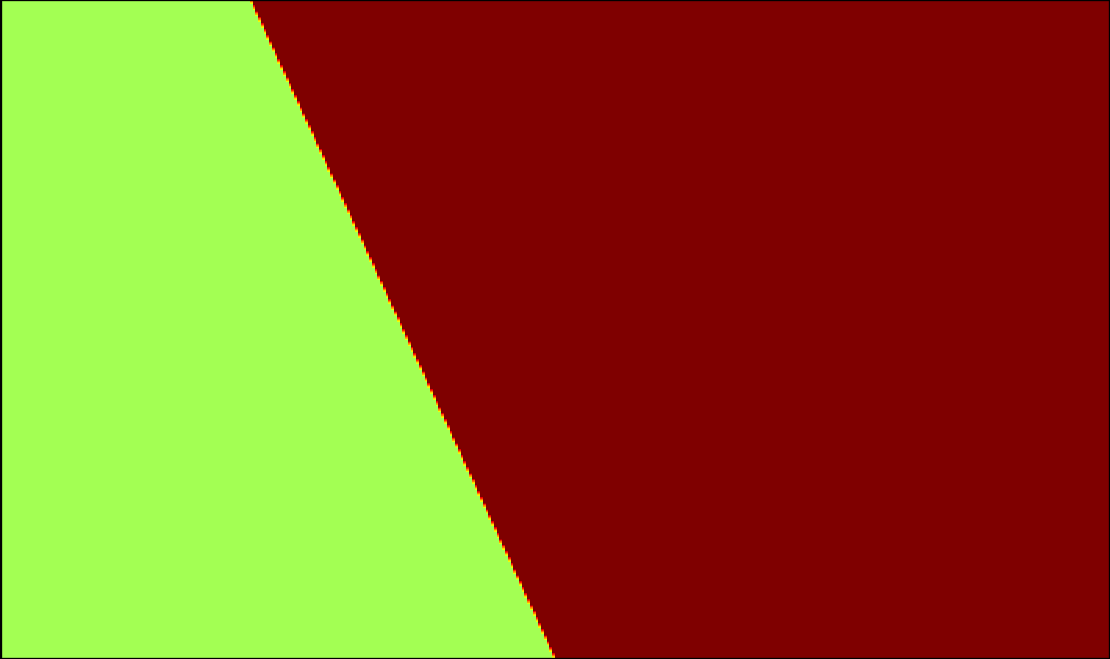}};
            \draw[black, thick] (img.south west) rectangle (img.north east);
            
            \path (img.south west) -- (img.south east) node[midway, below, yshift=-0.08cm] {\(x\)};
            \path (img.south west) -- (img.north west) node[midway, left, xshift=-0.08cm] {\(t\)};
            
            \node[below, yshift=-0.08cm] at (img.south west) {\(0\)};
            \node[below, yshift=-0.08cm] at (img.south east) {\(1\)};
            \node[left, xshift=-0.08cm] at (img.south west) {\(0\)};
            \node[left, xshift=-0.08cm] at (img.north west) {\(1\)};
        \end{tikzpicture}
        \begin{tikzpicture}
        \begin{axis}[
            legend pos=north west,
            legend cell align={left},
            legend style={fill opacity=0.8, draw opacity=1, text opacity=1, draw=lightgray204},
            tick align=outside,
            scale only axis=true,
            tick pos=left,
            width=.82\textwidth,
            height=.3\textwidth,
            xmajorgrids,
            ytick={0, 0.5, 1},
            yticklabels={0,,1},
            ylabel={\small\(\displaystyle x \mapsto u(t=0, x)\)},
            ylabel style={xshift=25pt, yshift=-70pt, rotate=-90},
            ymajorgrids,
            ymin=0.,
            legend columns=2,
        ]
        \addplot [thick, darkmagenta]  table[y index=0, x expr=\coordindex*0.0025, col sep=comma] {ic_data_0.csv};
        \end{axis}
        \end{tikzpicture}
        \caption{A Shock wave.}
        \label{fig:lwr_heatmap_exact_shock}
    \end{subfigure}
    \begin{subfigure}[t]{.30\textwidth}
        \centering
        \begin{tikzpicture}[scale=.9, clip=false]
            \node[anchor=south west, inner sep=0] (img) at (0,0) {\includegraphics[height=1.9cm]{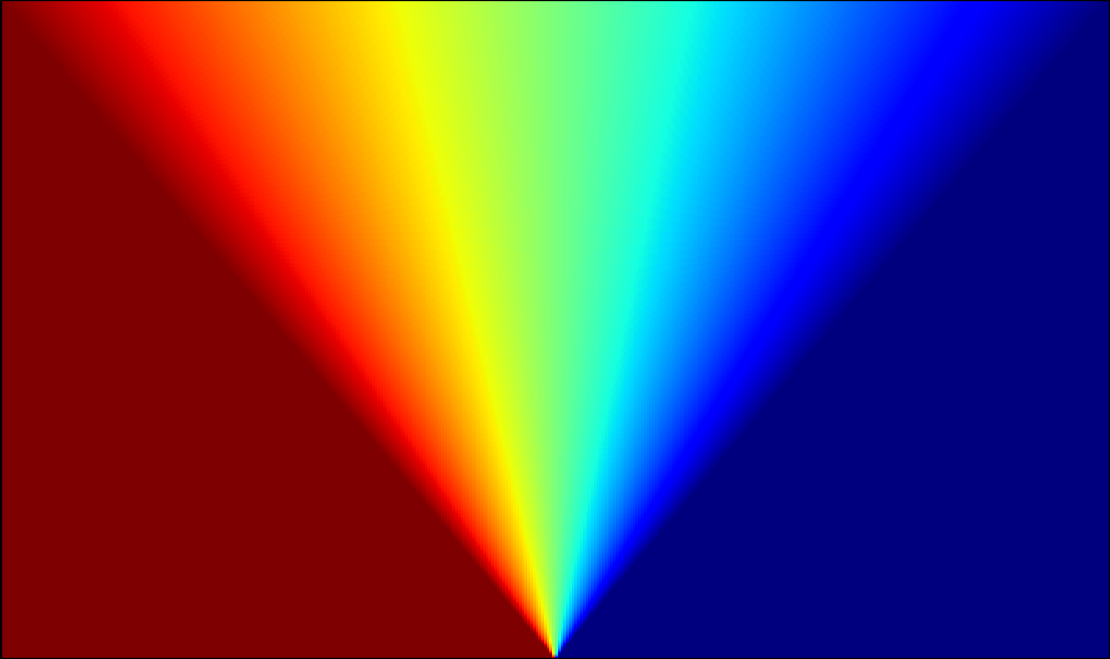}};
            \draw[black, thick] (img.south west) rectangle (img.north east);
            
            \path (img.south west) -- (img.south east) node[midway, below, yshift=-0.08cm] {\(x\)};
            \path (img.south west) -- (img.north west) node[midway, left, xshift=-0.08cm] {\(t\)};
            
            \node[below, yshift=-0.08cm] at (img.south west) {\(0\)};
            \node[below, yshift=-0.08cm] at (img.south east) {\(1\)};
            \node[left, xshift=-0.08cm] at (img.south west) {\(0\)};
            \node[left, xshift=-0.08cm] at (img.north west) {\(1\)};
        \end{tikzpicture}
        \begin{tikzpicture}
        \vspace{-20pt}
        \begin{axis}[
            legend pos=north west,
            legend cell align={left},
            legend style={fill opacity=0.8, draw opacity=1, text opacity=1, draw=lightgray204},
            tick align=outside,
            scale only axis=true,
            tick pos=left,
            width=.82\textwidth,
            height=.3\textwidth,
            xmajorgrids,
            ytick={0, 0.5, 1},
            yticklabels={0, ,1},
            ylabel={\small\(\displaystyle x \mapsto u(t=0, x)\)},
            ylabel style={xshift=25pt, yshift=-70pt, rotate=-90},
            ymajorgrids,
            legend columns=2,
        ]
        \addplot [thick, darkmagenta]  table[y index=0, x expr=\coordindex*0.0025, col sep=comma] {ic_data_1.csv};
        \end{axis}
        \end{tikzpicture}
        \caption{Rarefaction wave.}
        \label{fig:lwr_heatmap_exact_rarefaction}
    \end{subfigure}
    \begin{subfigure}[t]{.30\textwidth}
        \centering
        \begin{tikzpicture}[scale=.9, clip=false]
            \node[anchor=south west, inner sep=0] (img) at (0,0) {\includegraphics[height=1.9cm]{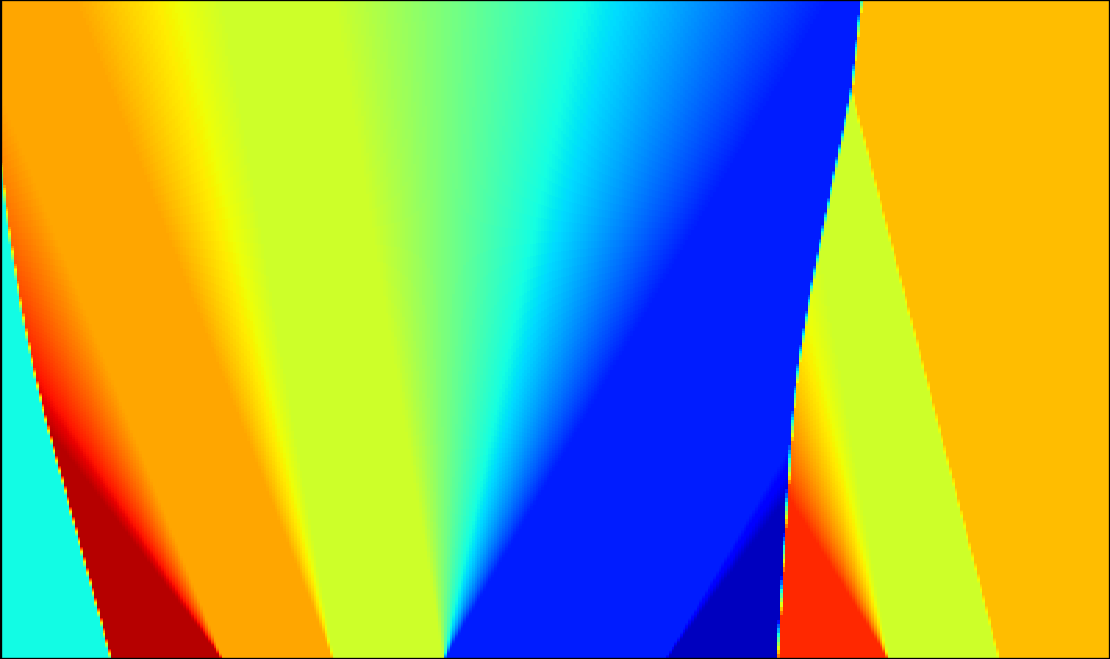}};
            \draw[black, thick] (img.south west) rectangle (img.north east);
            
            \path (img.south west) -- (img.south east) node[midway, below, yshift=-0.08cm] {\(x\)};
            \path (img.south west) -- (img.north west) node[midway, left, xshift=-0.08cm] {\(t\)};
            
            \node[below, yshift=-0.08cm] at (img.south west) {\(0\)};
            \node[below, yshift=-0.08cm] at (img.south east) {\(1\)};
            \node[left, xshift=-0.08cm] at (img.south west) {\(0\)};
            \node[left, xshift=-0.08cm] at (img.north west) {\(1\)};
        \end{tikzpicture}
        \begin{tikzpicture}
        \vspace{-20pt}
        \begin{axis}[
            legend pos=north west,
            legend cell align={left},
            legend style={fill opacity=0.8, draw opacity=1, text opacity=1, draw=lightgray204},
            tick align=outside,
            scale only axis=true,
            tick pos=left,
            width=.82\textwidth,
            height=.3\textwidth,
            xmajorgrids,
            ytick={0, 0.5, 1},
            yticklabels={0, ,1},
            ylabel={\small\(\displaystyle x \mapsto u(t=0, x)\)},
            ylabel style={xshift=25pt, yshift=-70pt, rotate=-90},
            ymajorgrids,
            legend columns=2,
        ]
        \addplot [thick, darkmagenta]  table[y index=0, x expr=\coordindex*0.0025, col sep=comma] {ic_data_2.csv};
        \end{axis}
        \end{tikzpicture}
        \caption{Eval IC.}
        \label{fig:heatmap_complex_laxhopf}
    \end{subfigure}
    \hfill
    \begin{subfigure}[t]{.07\textwidth}
        \centering
        \begin{tikzpicture}[scale=1., clip=false]
            \node[anchor=south west, inner sep=0] (img) at (0,0) {\includegraphics[height=1.9cm]{colorbar_flipped.png}};
            \draw[black, thick] (img.south west) rectangle (img.north east);
            \path (img.south east) -- (img.north east) node
            [midway, right, xshift=+0.03cm] {\(u\)};
            \node[right, xshift=+0.03cm] at (img.south east) {\(0\)};
            \node[right, xshift=+0.03cm] at (img.north east) {\(1\)};
            \node[below, yshift=-0.08cm] at (img.south east) {\phantom{0}};
        \end{tikzpicture}
    \end{subfigure}
    \caption{Exact solution for two Riemann problems (left, middle) and one piecewise-constant initial condition (right) from the evaluation set.}
    \label{fig:lwr_exact_sols}
\end{figure}

Example solutions of the Greenshields LWR are shown on various initial conditions in \Cref{fig:lwr_exact_sols}.

 \section{More General PDE Models}
\label{app:general_PDE}
\subsection*{2D Conservation Laws}
A general form of conservation law equations in higher dimension can be written as 
\begin{equation}
\begin{cases}
    \partial_t u + \nabla \cdot f(u) = 0 \\
    u(0, x) = u_\circ(x)
\end{cases} \end{equation}
for $(t, x) \in \mathbb{R_+} \times \mathbb R^n$, initial condition $u_\circ$, and $f: \mathbb R \to \mathbb R^n$. The Riemann solution for such model in higher dimension does no longer exist. Therefore, the unsupervised approach can be very helpful in this case. 

\subsection*{Conservation Laws with Discontinuity in the Flux}
A more general case of equations of conservation laws will include flux with dependence on the space variable. More precisely, 
\begin{equation}
    \begin{cases}
        \partial_t u(t,x) + \partial_x f(\gamma(x), u(t,x)) = 0\\
        u(0, x) = u_\circ(x).
    \end{cases}
\end{equation}
Such equations are more complex in nature due to the discontinuity of the flux function and consequently lack some analytical properties such as bounded variations. 
However, in the one-dimensional model, the Riemann solution can still be defined. 

\subsection*{ARZ-Type Equations}
The conservative form of the Aw-Rascle-Zhang (ARZ) equations~\citep{awr,zhang2002non} can be written in the form of
\begin{equation}
    \begin{cases}
        \partial_t \rho + \partial_x(\rho u) = 0 \\
        \partial_t(\rho(u + p(\rho))) + \partial_x(\rho u (u + p(\rho))) = 0 
    \end{cases}
\end{equation}
where $\rho$ is density, $u$ is the velocity function and $\rho\mapsto p(\rho)$ is know as pressure term. In particular, the first equation represents the mass conservation, while the second equation models the conservation of momentum.  \section{I-24 Experimental Dataset: Density Extraction, Training, and Evaluation}\label{app:i24}

\subsection{Density Fields Extraction From I-24 MOTION Dataset}\label{app:i24_data}

\begin{figure}
	\includegraphics[width=1.0\linewidth]{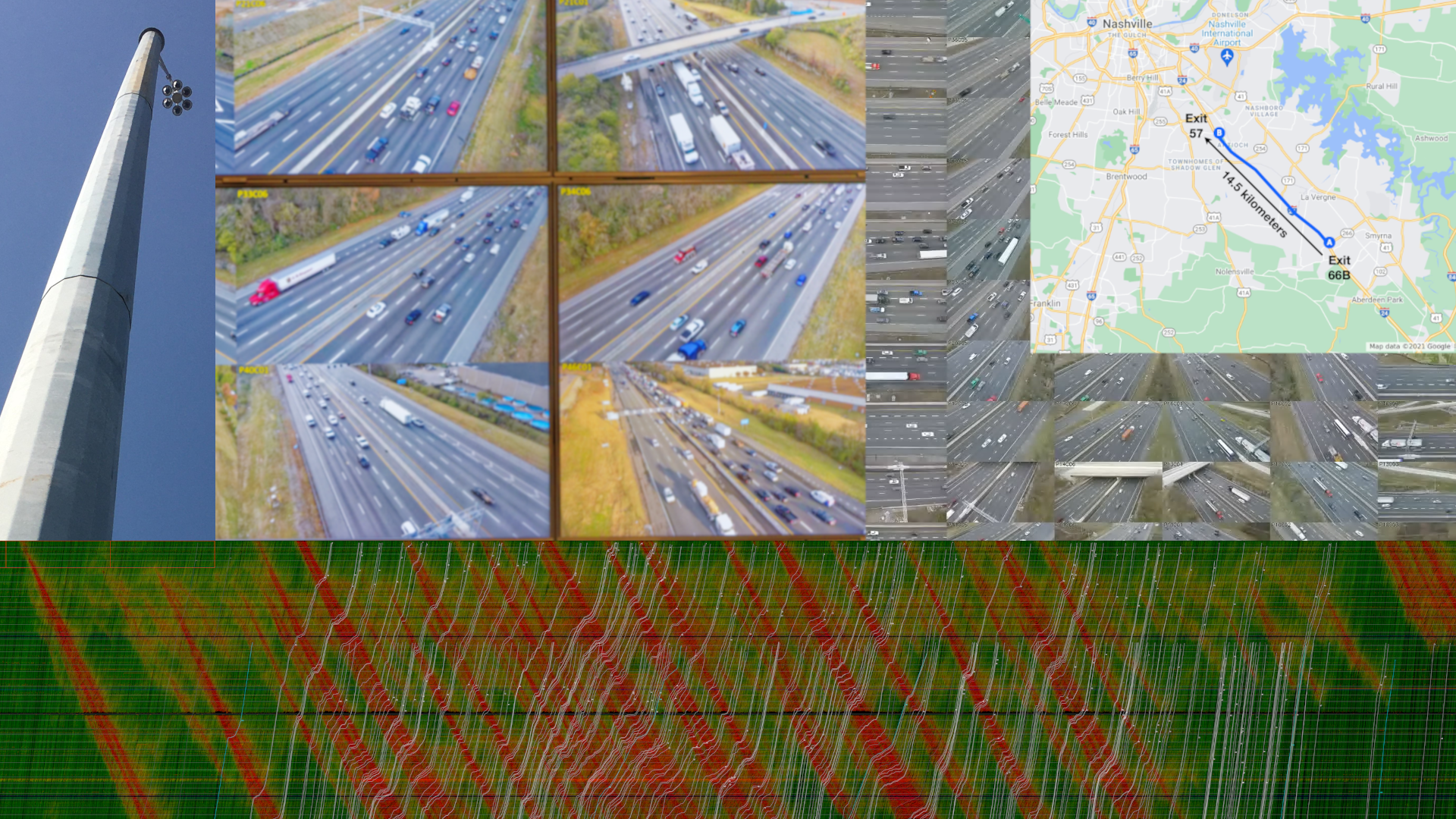}
	\caption{\textbf{I-24 MOTION illustration.} High-definition camera poles are mounted along a portion of I-24 at regular intervals. This generates massive amounts of video data, which is processed through a software stack. The resulting data for a single day is shown in the time-space diagram, displaying thousands of individual vehicle trajectories color-coded by speed (red for low speeds, green for high speeds), illustrating the complexity of the dataset.}
	\label{fig:i24_teaser}
\end{figure}

I-24 MOTION is a large-scale traffic monitoring system installed along a section of Interstate 24 near Nashville, Tennessee. It uses a dense network of high-resolution cameras and computer vision algorithms to capture detailed, real-time vehicle trajectories across multiple lanes and miles of highway. The data collection network and resulting trajectory data are illustrated in \Cref{fig:i24_teaser}.

For our experiments, we use the INCEPTION dataset\footnote{Available at \href{https://i24motion.org}{\texttt{i24motion.org}} as part of the INCEPTION data release.}~\citep{gloudemans202324} from I-24 MOTION, consisting of ten days of data, each covering the morning rush hour (6:00 AM to 10:00 AM). The dataset for each day comprises 15-20 GB stored as a single JSON file. We first split each file into manageable 1 GB chunks and parse them efficiently using \texttt{simdjson}~\citep{langdale2019parsing}, which enables extraction of density fields in approximately 3-5 minutes per 20 GB file.

To construct the density fields, we discretize the spatiotemporal domain into cells of size 0.02 miles ($\approx$32 meters) in space and 0.1 seconds in time, aggregating data across all four lanes. Vehicle counts in each cell are normalized to obtain densities in vehicles per kilometer per lane. To reduce noise, we average over 100 consecutive time steps (i.e., 10 seconds) and over 2 adjacent spatial cells (i.e., 0.04 miles or $\approx$64 meters). This results in a grid of 100 spatial cells (4 miles / 0.04 miles) and approximately 1440 time steps (4 hours / 10 seconds). We clip the first and last segments of each day to exclude low-density, free-flow regimes with incomplete data, retaining 1300 time steps per day depending on data quality. To avoid extreme outliers, we cap densities at 140 vehicles/km/lane. For all training and evaluation purposes, we then normalize densities so that the maximum density is $1$.

Due to occasional sensor failures, such as malfunctioning camera poles or occlusions by bridges, there are seven spatial locations with missing data. We fill these gaps by linear interpolation between the adjacent upstream and downstream cells.

\begin{figure}
	\centering

    \begin{subfigure}[t]{.49\textwidth}
        \begin{tikzpicture}
            \node[anchor=south west, inner sep=0, draw=black, line width=1.5pt] (img) at (0,0) {\includegraphics[width=0.99\textwidth]{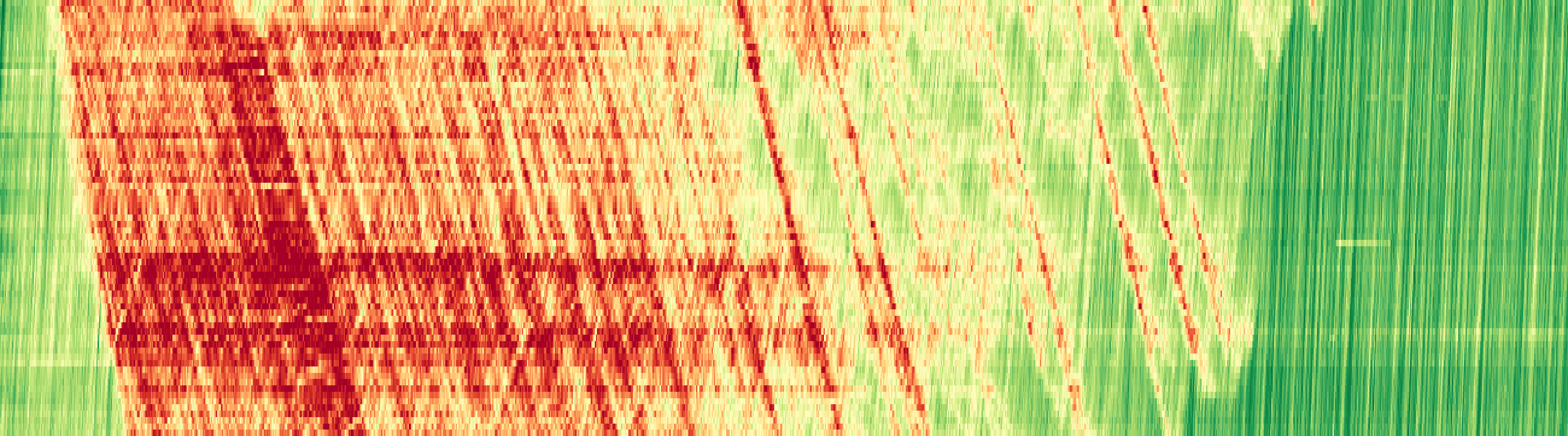}};
        \end{tikzpicture}
        \caption{Nov 21, 2022}
    \end{subfigure}
    \hfill
    \begin{subfigure}[t]{.49\textwidth}
        \begin{tikzpicture}
            \node[anchor=south west, inner sep=0, draw=black, line width=1.5pt] (img) at (0,0) {\includegraphics[width=0.99\textwidth]{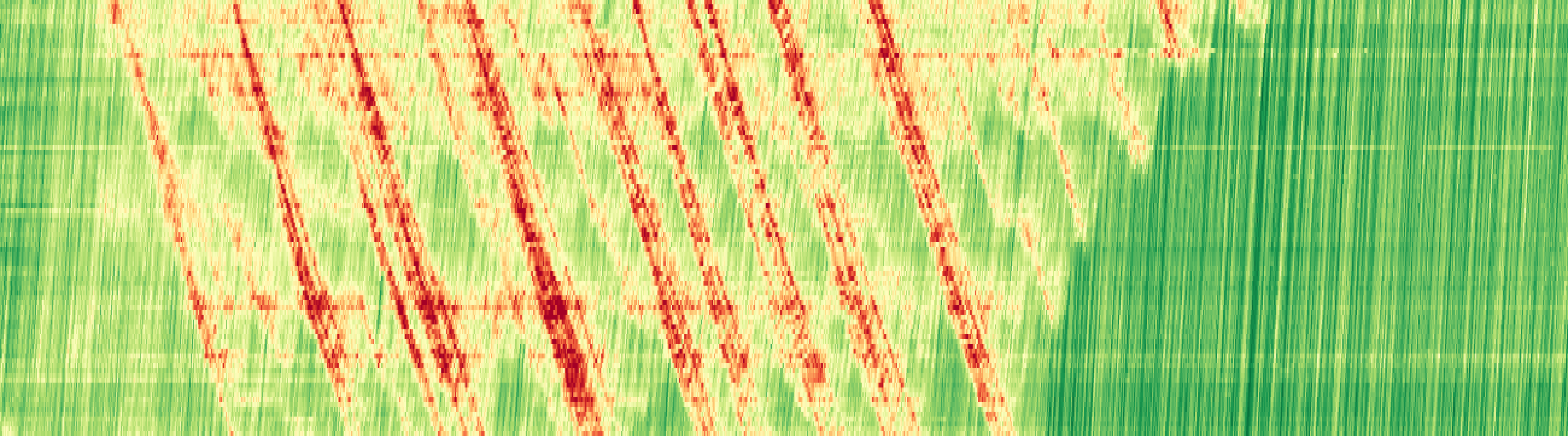}};
        \end{tikzpicture}    
        \caption{Nov 22, 2022}
    \end{subfigure}

    \vspace{0.2cm}
    \begin{subfigure}[t]{.49\textwidth}
        \begin{tikzpicture}
            \node[anchor=south west, inner sep=0, draw=black, line width=1.5pt] (img) at (0,0) {\includegraphics[width=0.99\textwidth]{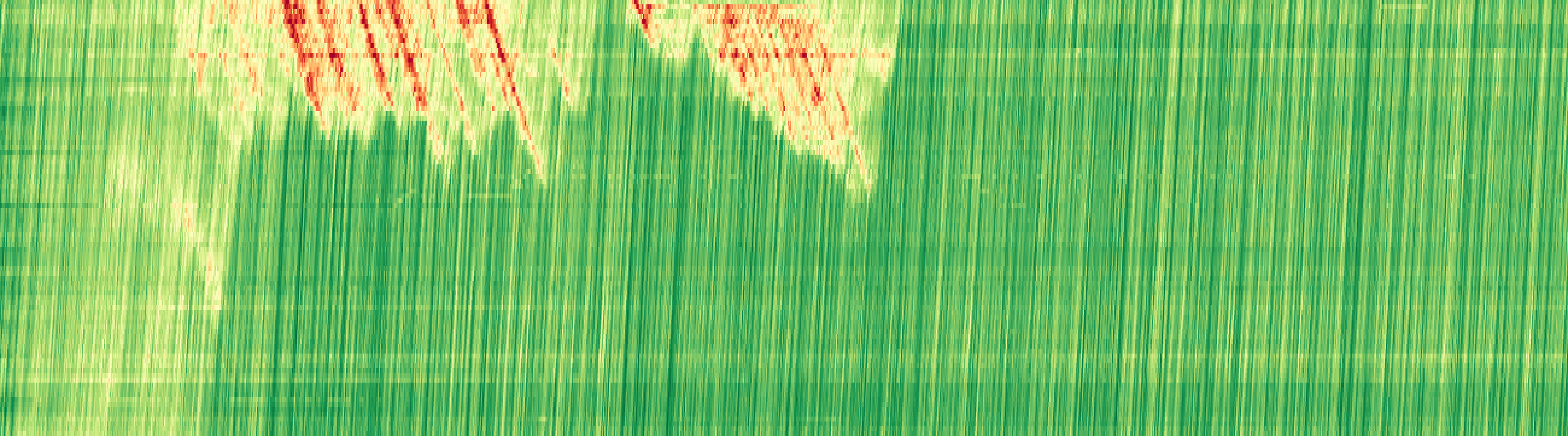}};
        \end{tikzpicture}
        \caption{Nov 23, 2022}
    \end{subfigure}
    \hfill
    \begin{subfigure}[t]{.49\textwidth}
        \begin{tikzpicture}
            \node[anchor=south west, inner sep=0, draw=black, line width=1.5pt] (img) at (0,0) {\includegraphics[width=0.99\textwidth]{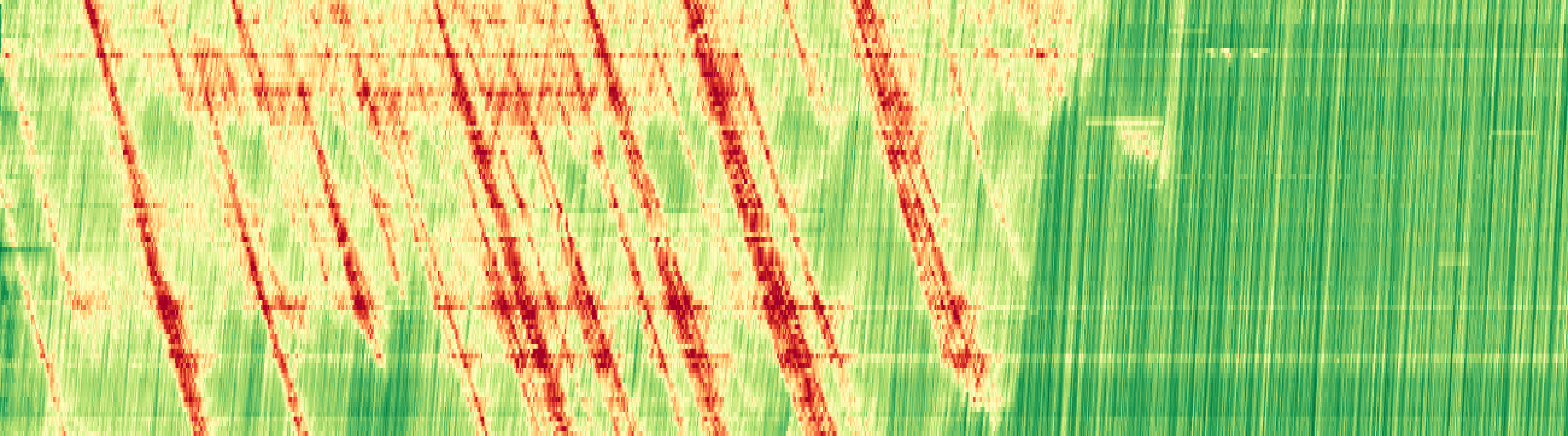}};
        \end{tikzpicture}    
        \caption{Nov 28, 2022}
    \end{subfigure}
    
    \vspace{0.2cm}
    \begin{subfigure}[t]{.49\textwidth}
        \begin{tikzpicture}
            \node[anchor=south west, inner sep=0, draw=black, line width=1.5pt] (img) at (0,0) {\includegraphics[width=0.99\textwidth]{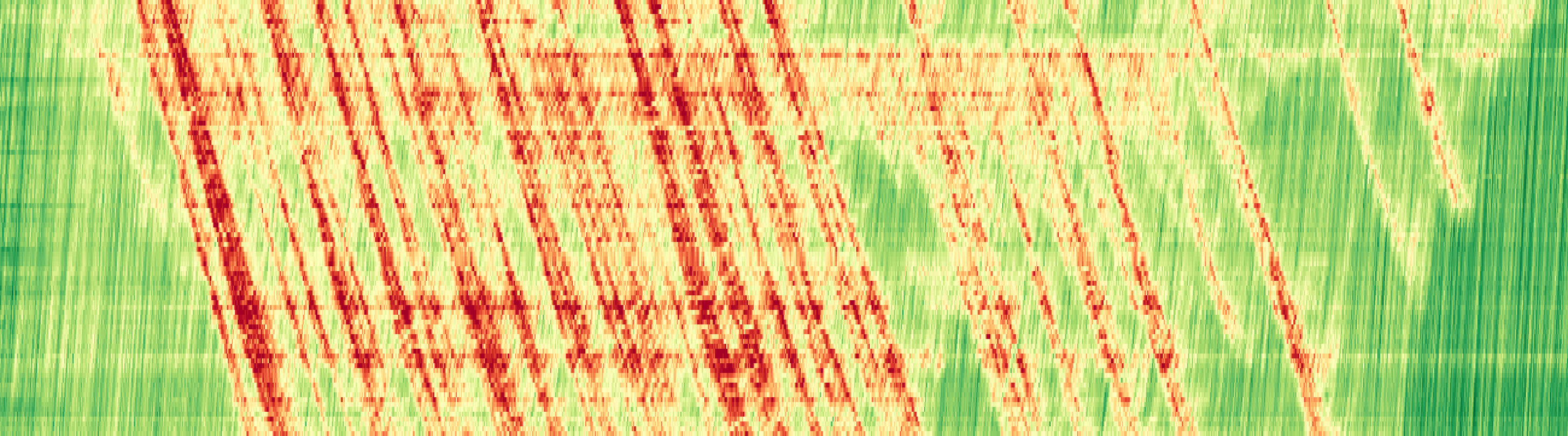}};
        \end{tikzpicture}
        \caption{Nov 29, 2022}
    \end{subfigure}
    \hfill
    \begin{subfigure}[t]{.49\textwidth}
        \begin{tikzpicture}
            \node[anchor=south west, inner sep=0, draw=black, line width=1.5pt] (img) at (0,0) {\includegraphics[width=0.99\textwidth]{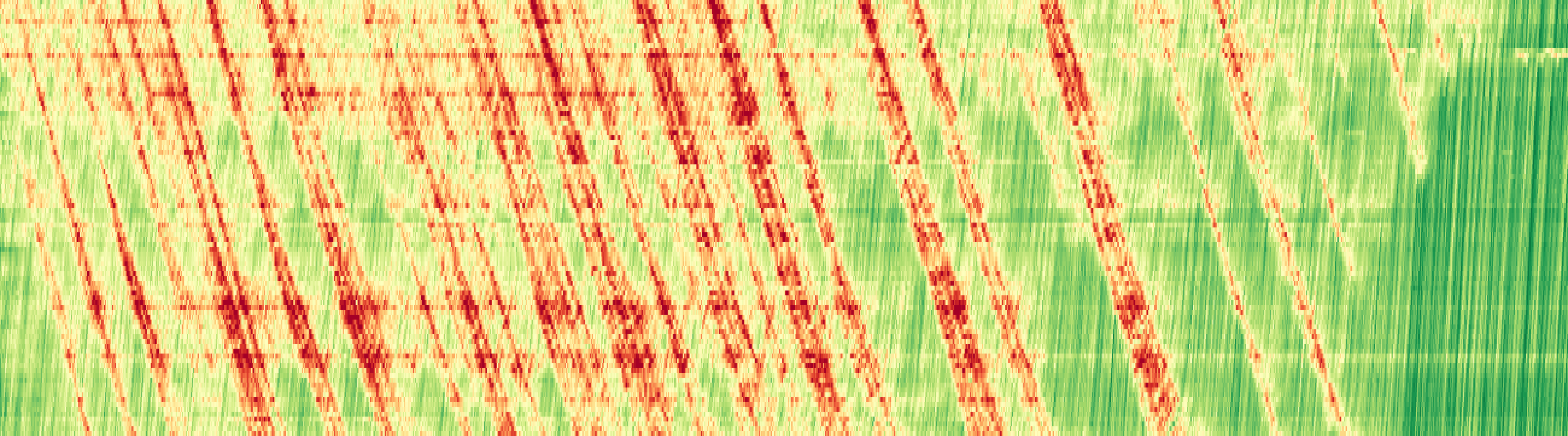}};
        \end{tikzpicture}    
        \caption{Nov 30, 2022}
    \end{subfigure}

    \vspace{0.2cm}
    \begin{subfigure}[t]{.49\textwidth}
        \begin{tikzpicture}
            \node[anchor=south west, inner sep=0, draw=black, line width=1.5pt] (img) at (0,0) {\includegraphics[width=0.99\textwidth]{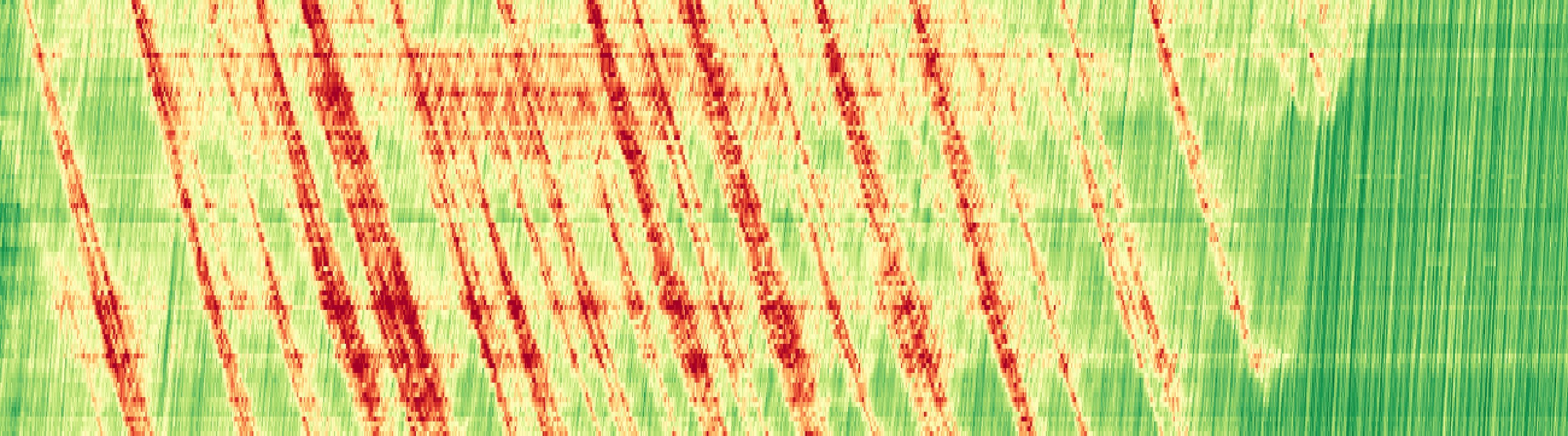}};
        \end{tikzpicture}
        \caption{Dec 01, 2022}
    \end{subfigure}
    \hfill
    \begin{subfigure}[t]{.49\textwidth}
        \begin{tikzpicture}
            \node[anchor=south west, inner sep=0, draw=black, line width=1.5pt] (img) at (0,0) {\includegraphics[width=0.99\textwidth]{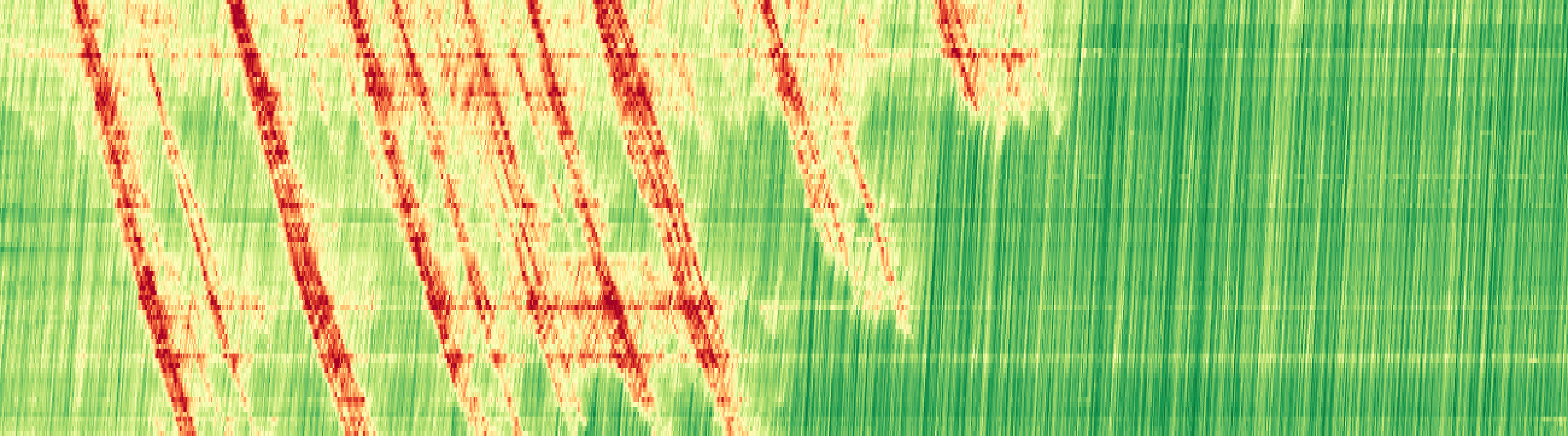}};
        \end{tikzpicture}    
        \caption{Dec 02, 2022}
    \end{subfigure}
	\caption{Time-space diagrams of car trajectories extracted from the video, colour-coded by density, for different dates. See \Cref{sec:how_to_read} for how to read the heatmaps.}
    \label{fig:i24_data}
\end{figure}

\Cref{fig:i24_data} shows the density fields we extracted from I-24 MOTION data. Higher densities (in red) correspond to stop-and-go waves and congestion, while lower densities (green) correspond to free-flow traffic. The processing code and resulting data are available in our codebase.

We exclude data from November 24 and 25, 2022, from our analysis, as both days correspond to holiday periods with purely free-flow, low-traffic conditions and no observable stop-and-go waves. These days are therefore not relevant to our study, which focuses on modeling traffic dynamics in the presence of congestion. The remaining days still include sufficient free-flow segments to evaluate model robustness in those regimes. Nevertheless, we include the excluded days in the released dataset for completeness.

\subsection{Boundary Conditions} \label{app:i24_boundary_conditions}

For both training and evaluation of NFV on the I-24 dataset, we initialize the model using a single time step of real data and provide one boundary cell at each end of the road, using the corresponding real values. While it is possible to use additional ground truth data to improve accuracy, we deliberately restrict ourselves for two main reasons: (1) to allow fair comparison with the Godunov scheme, which uses a single boundary cell per side, and (2) to reflect realistic deployment scenarios, where boundary densities might only be measured at a few fixed points (e.g., at the road extremities), or predicted using a separate model.

\begin{figure}
	\centering

    \begin{subfigure}[t]{.49\textwidth}
        \begin{tikzpicture}
            \node[anchor=south west, inner sep=0, line width=1.5pt] (img) at (0,0) {\includegraphics[width=0.99\textwidth]{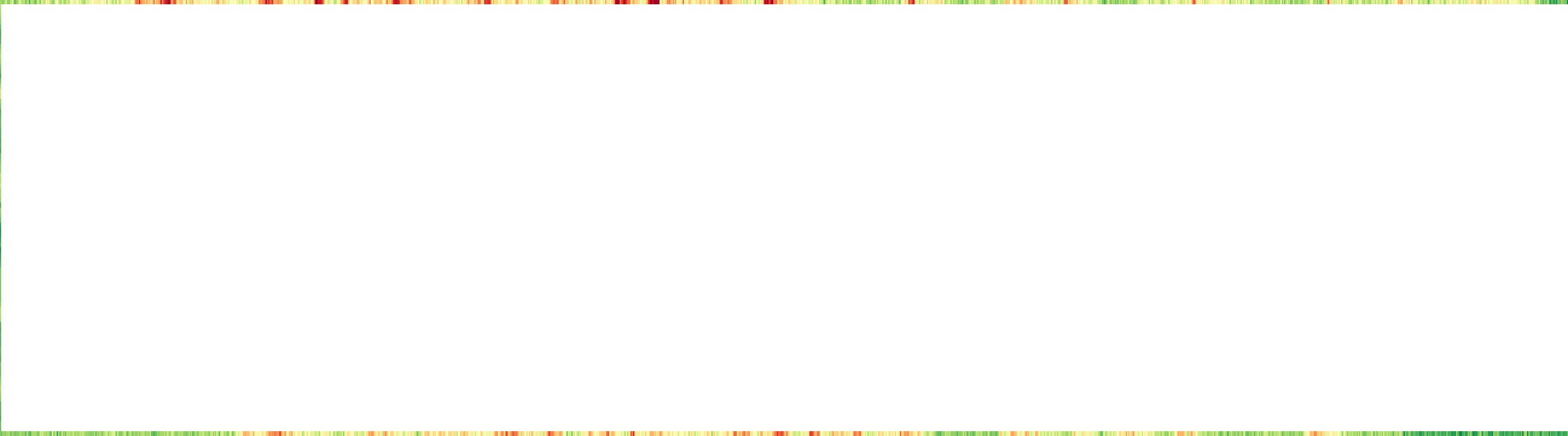}};
        \end{tikzpicture}
        \caption{Nov 29, 2022 boundary conditions for $\text{NFV}_3^1$}
    \end{subfigure}
    \hfill
    \begin{subfigure}[t]{.49\textwidth}
        \begin{tikzpicture}
            \node[anchor=south west, inner sep=0, line width=1.5pt] (img) at (0,0) {\includegraphics[width=0.99\textwidth]{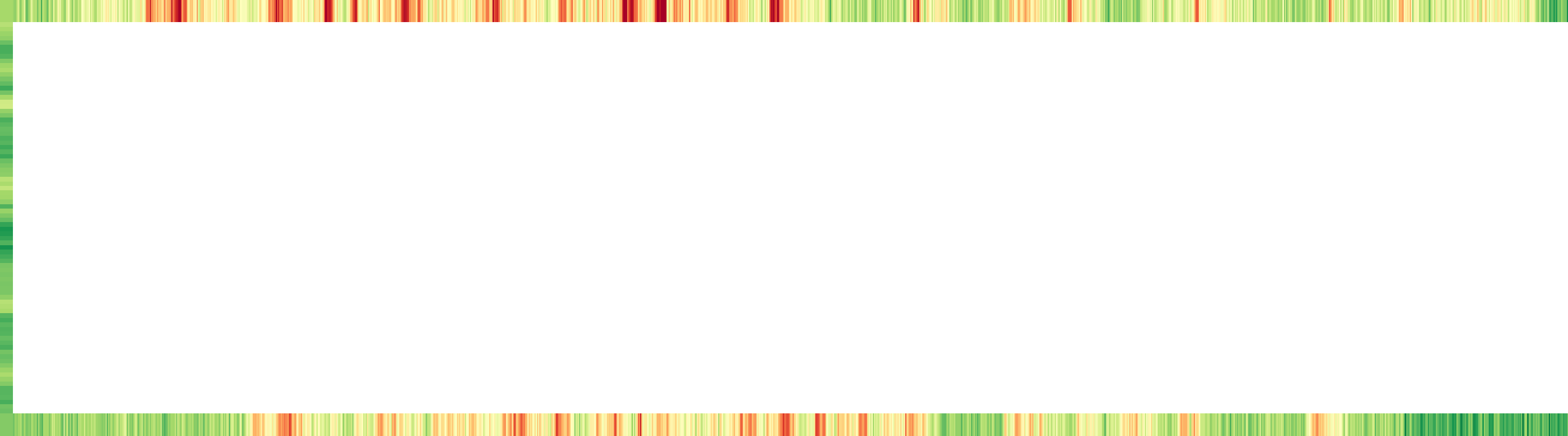}};
        \end{tikzpicture}    
        \caption{Nov 29, 2022 boundary conditions for $\text{NFV}_{11}^{11}$}
    \end{subfigure}
    \caption{\textbf{Boundary conditions used by $\text{NFV}$ during training and evaluation on the I-24 dataset.} The figures show the input provided to the model: the initial condition at $t=0$ on the left, and boundary conditions at $x=0$ (bottom) and $x = x_{\max}$ (top). The model must then predict the interior (i.e., the region shown in \Cref{fig:i24_data}) autoregressively: it uses its own output at time $t$ to predict the state at time $t + dt$, without receiving any additional data beyond the fixed boundaries. Note that both figures use the same underlying data; for $\text{NFV}_{11}^{11}$, the boundary values are duplicated to provide the required input padding.}
    \label{fig:i24_boundaries}
\end{figure}

For models that require a wider input stencil (e.g., those larger than $\text{NFV}_3^1$), we pad the boundaries by duplicating the available single-cell values. This ensures that all models, no matter their size-Godunov, $\text{NFV}3^1$, or $\text{NFV}{11}^{11}$—receive the same amount of boundary information. \Cref{fig:i24_boundaries} illustrates the boundary setup in both cases, showing which values are provided as an input and which are left to be predicted.

Finally, we emphasize that initial and boundary conditions are not included when computing metrics, whether in the training loss or at evaluation.

\subsection{Reading the Heatmaps} \label{sec:how_to_read}

\begin{figure}
	\centering
	\begin{tikzpicture}[scale=0.9, clip=false]

    \node[anchor=south west, inner sep=0] (colorbar) at (0,0) {\includegraphics[height=0.6\linewidth,angle=-90]{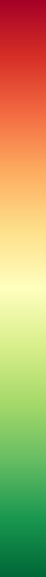}};

		\draw[black, thick] (colorbar.south west) rectangle (colorbar.north east);

		\path (colorbar.south west) -- (colorbar.south east) node[midway, anchor=center, xshift=0.0cm, yshift=-0.3cm] {Density};

		\node[below, xshift=+0.03cm] at (colorbar.south west) {0};
		\node[below, xshift=+0.03cm] at (colorbar.south east) {1};

	\end{tikzpicture}
	\caption{Colorbar showing density scale for all I-24 data heatmaps.}
    \label{fig:i24_colorbars}
\end{figure}
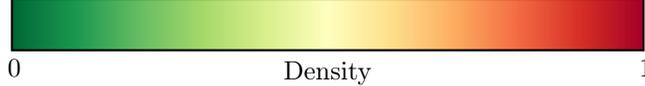

This section provides a brief explanation and intuition for interpreting the heatmaps displaying I-24 MOTION data. The horizontal axis represents time, increasing from left to right, while the vertical axis represents space along the road, increasing from bottom to top. The color encodes traffic density, normalized between 0 and 1, according to the colormap shown in \Cref{fig:i24_colorbars}, where green indicates low density traffic (free flow) and red indicates high density traffic (congestion). Unless otherwise specified, only the model predictions are shown, while initial and boundary conditions are omitted for clarity. Stop-and-go waves appear as high-density (red) bands that propagate upstream, i.e., move backward through traffic.

\subsection{Predictions}

\begin{figure}
	\centering

    \begin{subfigure}[t]{.49\textwidth}
        \begin{tikzpicture}
            \node[anchor=south west, inner sep=0, draw=black, line width=1.5pt] (img) at (0,0) {\includegraphics[width=0.99\textwidth]{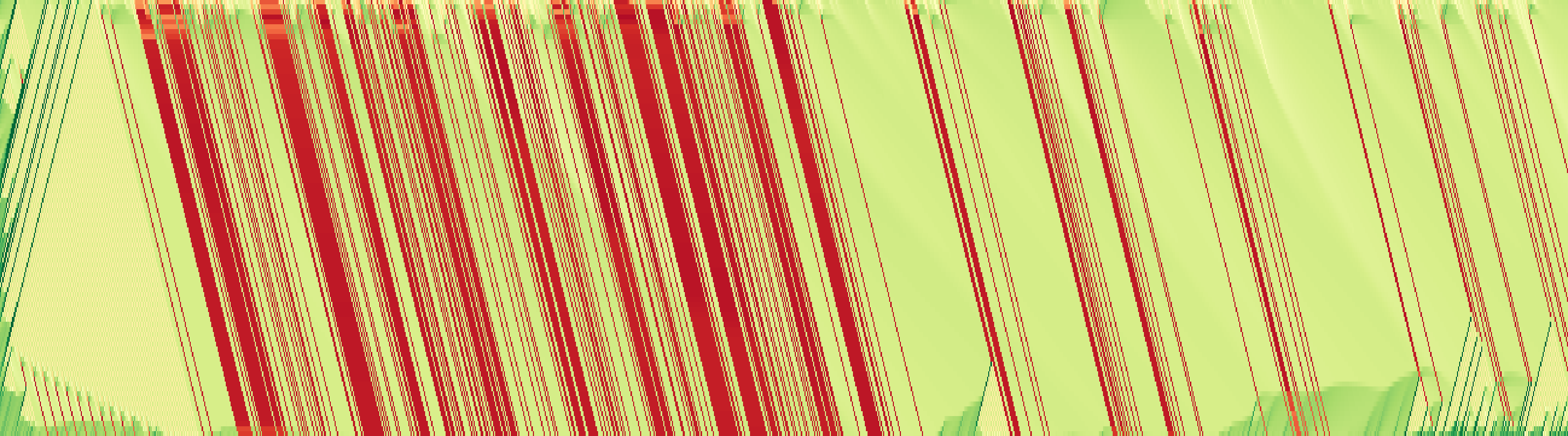}};
        \end{tikzpicture}
        \caption{FV fit: Greenshields}
    \end{subfigure}
    \hfill
    \begin{subfigure}[t]{.49\textwidth}
        \begin{tikzpicture}
            \node[anchor=south west, inner sep=0, draw=black, line width=1.5pt] (img) at (0,0) {\includegraphics[width=0.99\textwidth]{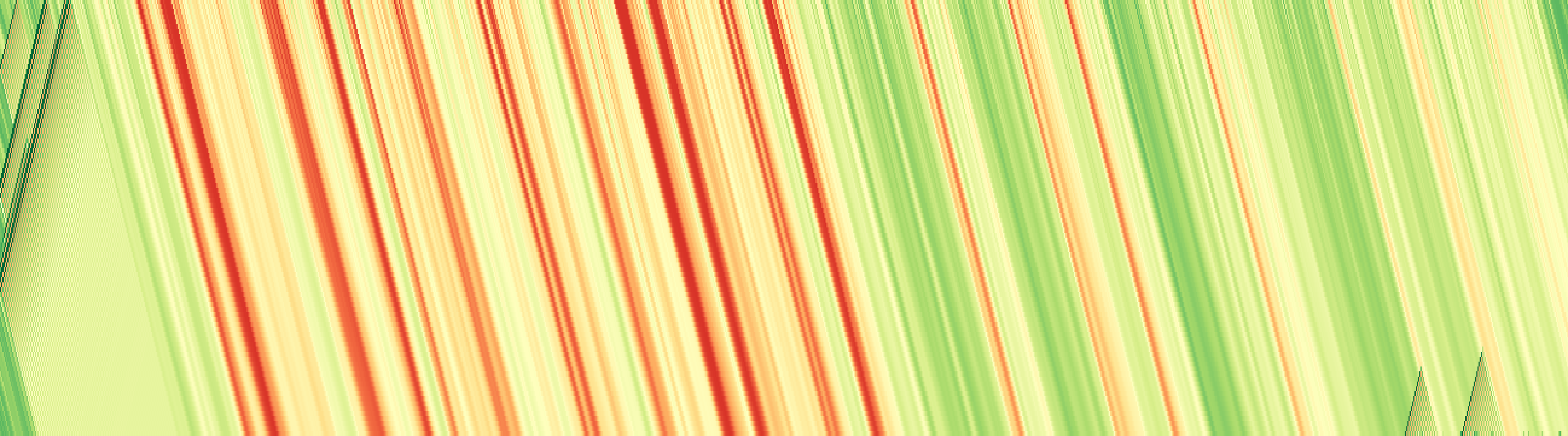}};
        \end{tikzpicture}    
        \caption{FV fit: Triangular}
    \end{subfigure}

    \vspace{0.2cm}
    \begin{subfigure}[t]{.49\textwidth}
        \begin{tikzpicture}
            \node[anchor=south west, inner sep=0, draw=black, line width=1.5pt] (img) at (0,0) {\includegraphics[width=0.99\textwidth]{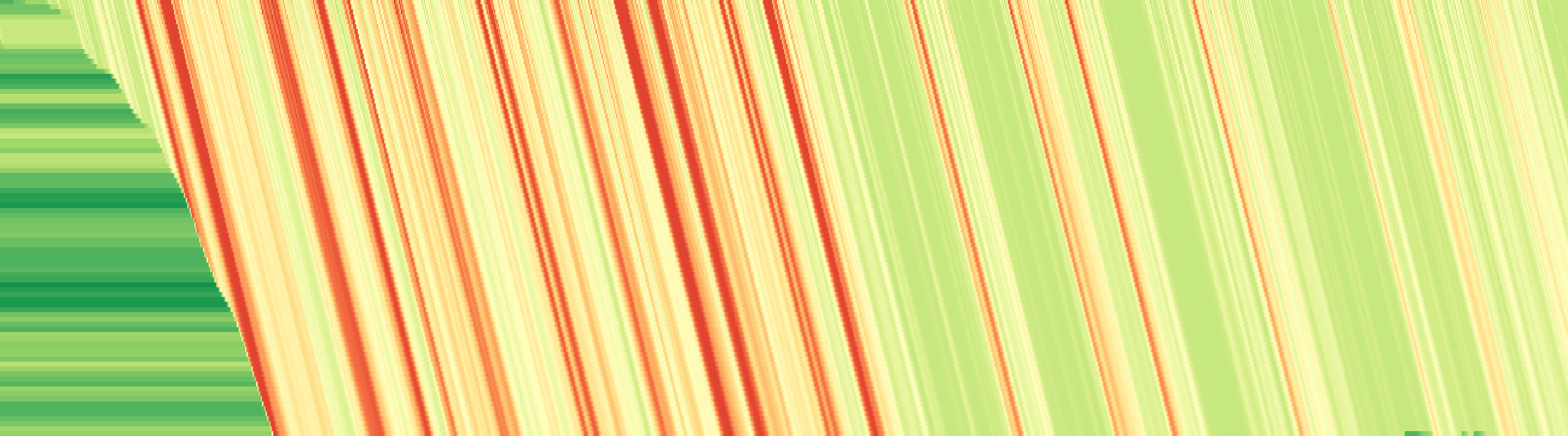}};
        \end{tikzpicture}
        \caption{FV fit: Trapezoidal}
    \end{subfigure}
    \hfill
    \begin{subfigure}[t]{.49\textwidth}
        \begin{tikzpicture}
            \node[anchor=south west, inner sep=0, draw=black, line width=1.5pt] (img) at (0,0) {\includegraphics[width=0.99\textwidth]{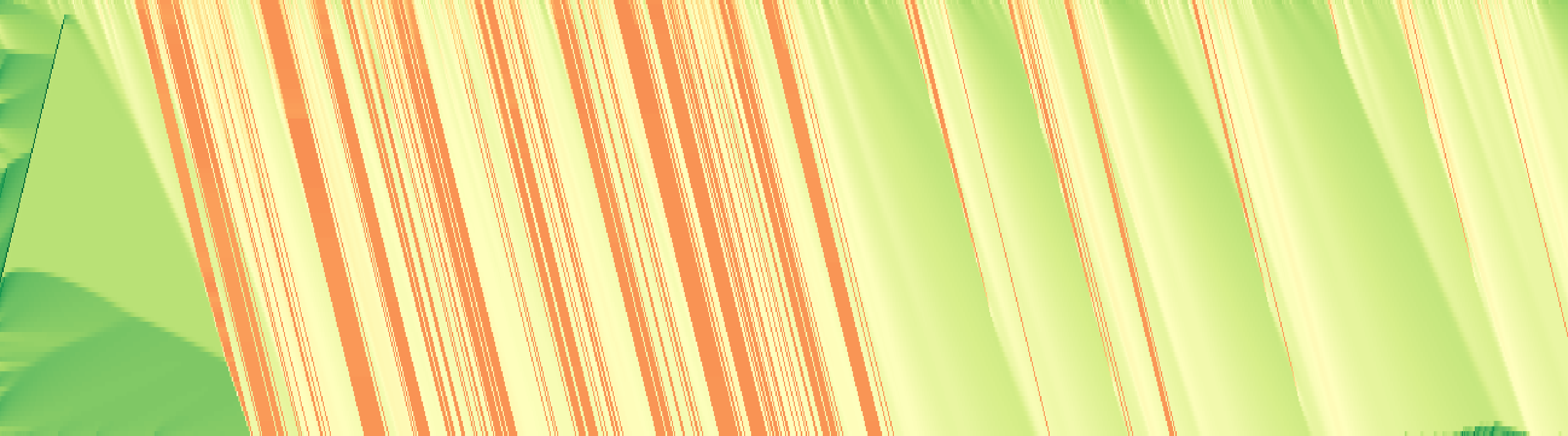}};
        \end{tikzpicture}    
        \caption{FV fit: Greenberg}
    \end{subfigure}
    
    \vspace{0.2cm}
    \begin{subfigure}[t]{.49\textwidth}
        \begin{tikzpicture}
            \node[anchor=south west, inner sep=0, draw=black, line width=1.5pt] (img) at (0,0) {\includegraphics[width=0.99\textwidth]{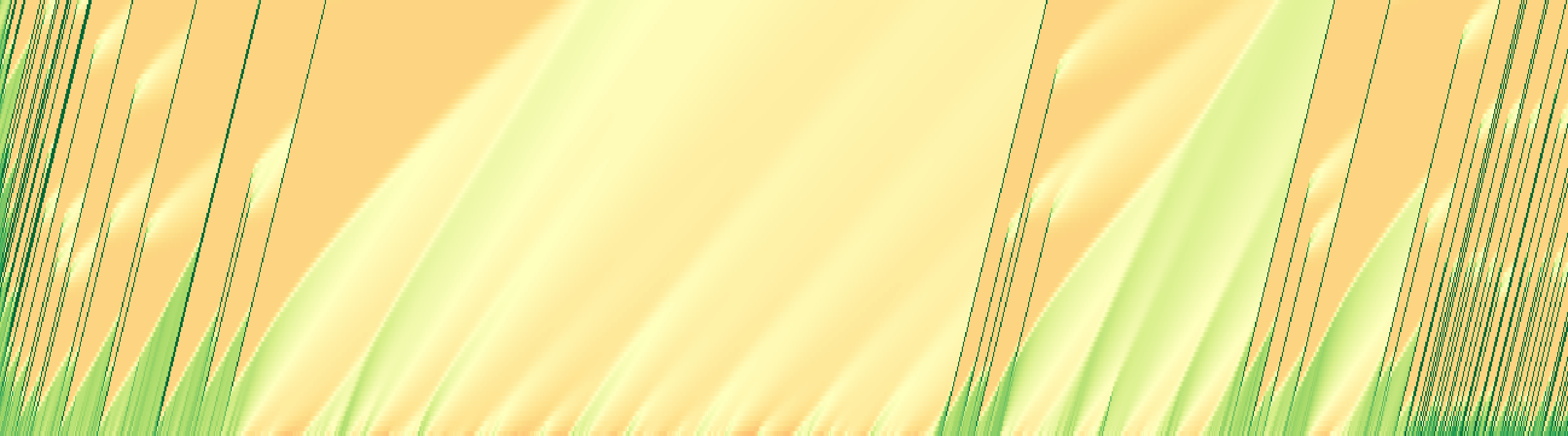}};
        \end{tikzpicture}
        \caption{FV fit: Underwood}
    \end{subfigure}
    \hfill
    \begin{subfigure}[t]{.49\textwidth}
        \begin{tikzpicture}
            \node[anchor=south west, inner sep=0, draw=black, line width=1.5pt] (img) at (0,0) {\includegraphics[width=0.99\textwidth]{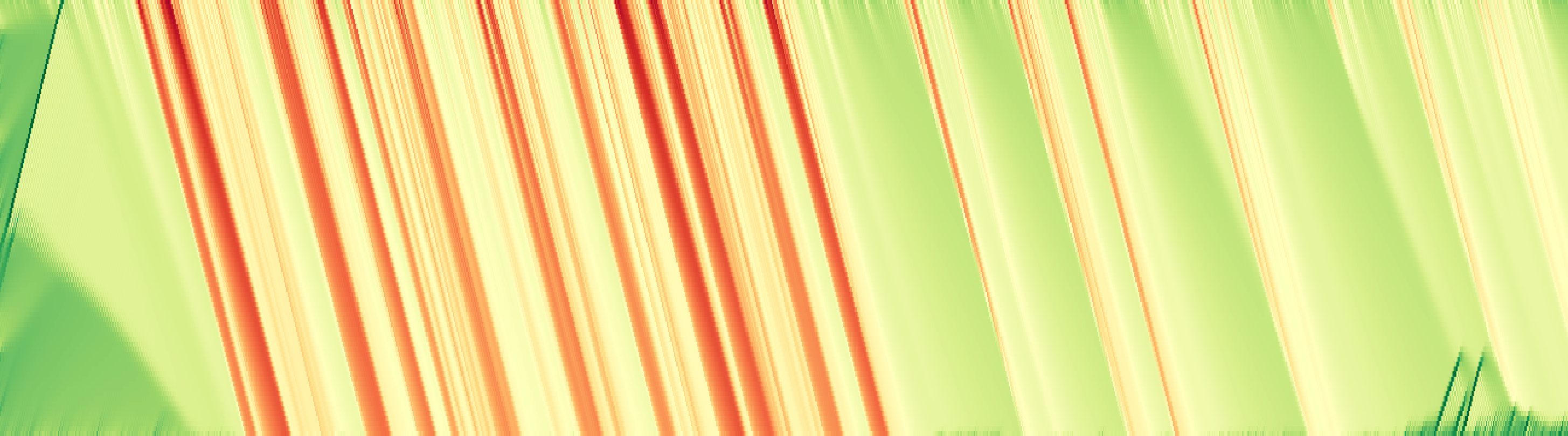}};
        \end{tikzpicture}    
        \caption{FV fit: $\text{NFV}_3^1$}
    \end{subfigure}

    \vspace{0.2cm}
    \begin{subfigure}[t]{.49\textwidth}
        \begin{tikzpicture}
            \node[anchor=south west, inner sep=0, draw=black, line width=1.5pt] (img) at (0,0) {\includegraphics[width=0.99\textwidth]{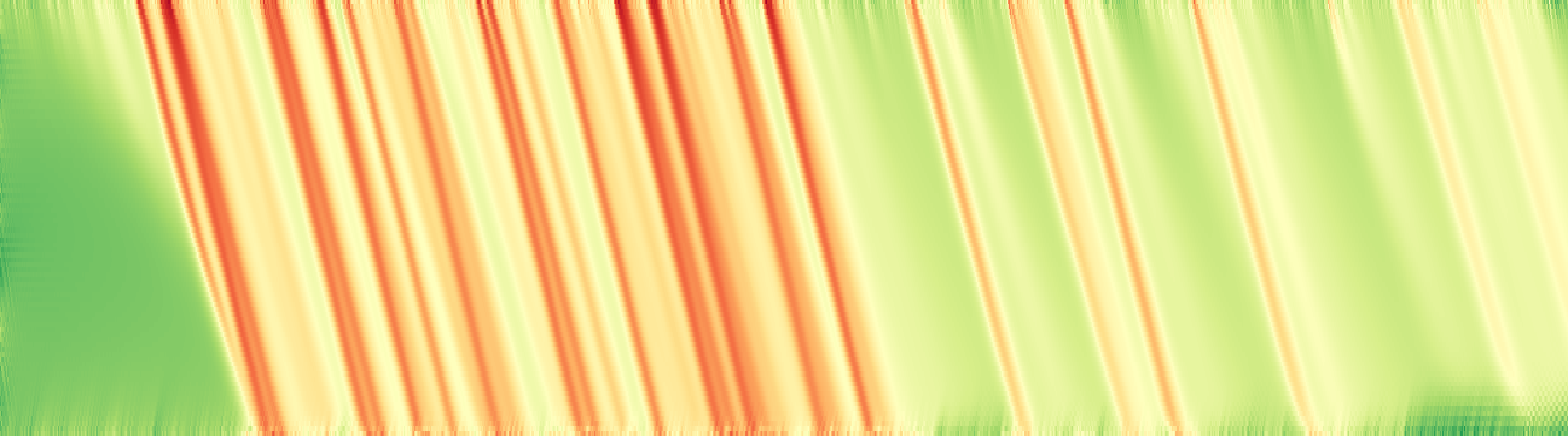}};
        \end{tikzpicture}
        \caption{FV fit: $\text{NFV}_5^5$}
    \end{subfigure}
    \hfill
    \begin{subfigure}[t]{.49\textwidth}
        \begin{tikzpicture}
            \node[anchor=south west, inner sep=0, draw=black, line width=1.5pt] (img) at (0,0) {\includegraphics[width=0.99\textwidth]{pred_pred_nx10_nt11_TX.png}};
        \end{tikzpicture}    
        \caption{FV fit: $\text{NFV}_{11}^{11}$}
    \end{subfigure}

    \vspace{0.2cm}
    \begin{subfigure}[t]{.49\textwidth}
        \begin{tikzpicture}
            \node[anchor=south west, inner sep=0, draw=black, line width=1.5pt] (img) at (0,0) {\includegraphics[width=0.99\textwidth]{ground_truth_TX.png}};
        \end{tikzpicture}
        \caption{Ground truth}
    \end{subfigure}
    \caption{\textbf{Predictions of FV methods and trained NFV.} Corresponding metrics are reported in \Cref{tab:i24_part1_metrics_table}. Among the FV methods, only the Triangular, Trapezoidal, and Greenberg flows provide a reasonable fit to the I-24 MOTION data. In contrast, NFV models show increasing predictive accuracy with model complexity. For example, $\text{NFV}_{11}^{11}$ captures significantly more stop-and-go waves (in red) than $\text{NFV}_{5}^{5}$ or $\text{NFV}_{3}^{1}$, as well as fast low-density waves (in green), enabling it to correctly predict the early dissipation of the final two waves. However, it exhibits oscillations toward the end of the prediction window, likely due to limited generalization caused by the scarcity of low-density (dark green) patterns in the training data; nevertheless, the primary objective when modeling experimental data is to accurately capture the evolution of congestion waves, whereas free-flow traffic is of lesser interest. All models were trained on only the first 25\% of the ground truth sequence, and the predictions are generated fully autoregressively. See \Cref{sec:how_to_read} for how to read the heatmaps.}
    \label{fig:i24_part1_all_predictions}
\end{figure}

\begin{figure}
    \centering
    
    \caption*{\textbf{Prediction on training day}}
    
    \begin{subfigure}[t]{.325\textwidth}
      \begin{tikzpicture}
          \node[anchor=south west, inner sep=0, draw=black, line width=1.5pt] (img) at (0,0) {\includegraphics[width=0.99\textwidth]{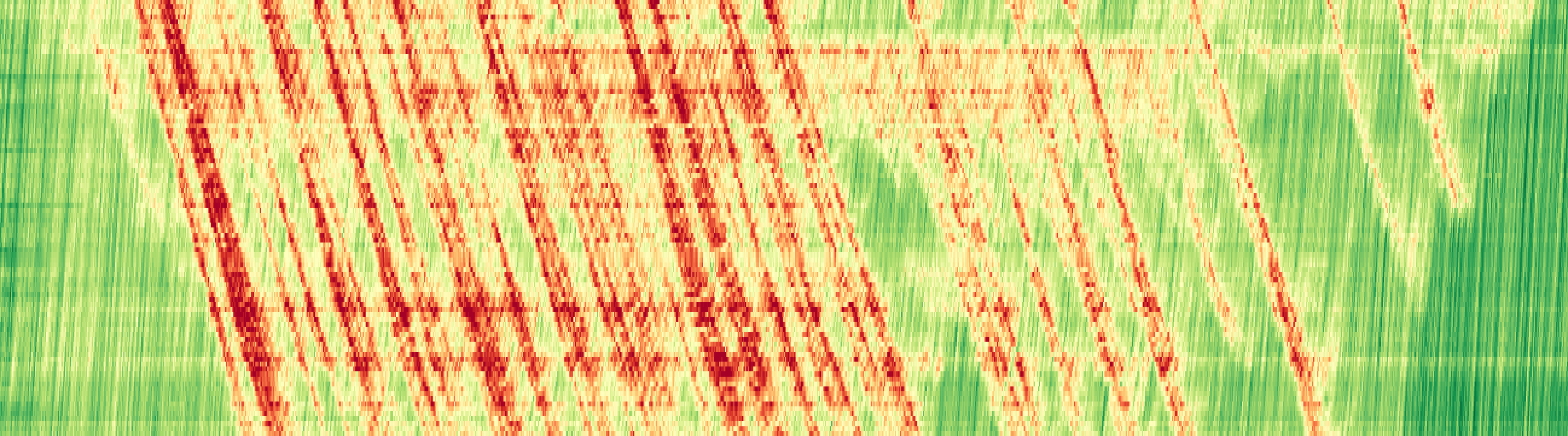}};
      \end{tikzpicture}
      \caption{Ground truth (Nov 29, 2022)}
    \end{subfigure}
    \hfill
    \begin{subfigure}[t]{.325\textwidth}
        \begin{tikzpicture}
            \node[anchor=south west, inner sep=0, draw=black, line width=1.5pt] (img) at (0,0) {\includegraphics[width=0.99\textwidth]{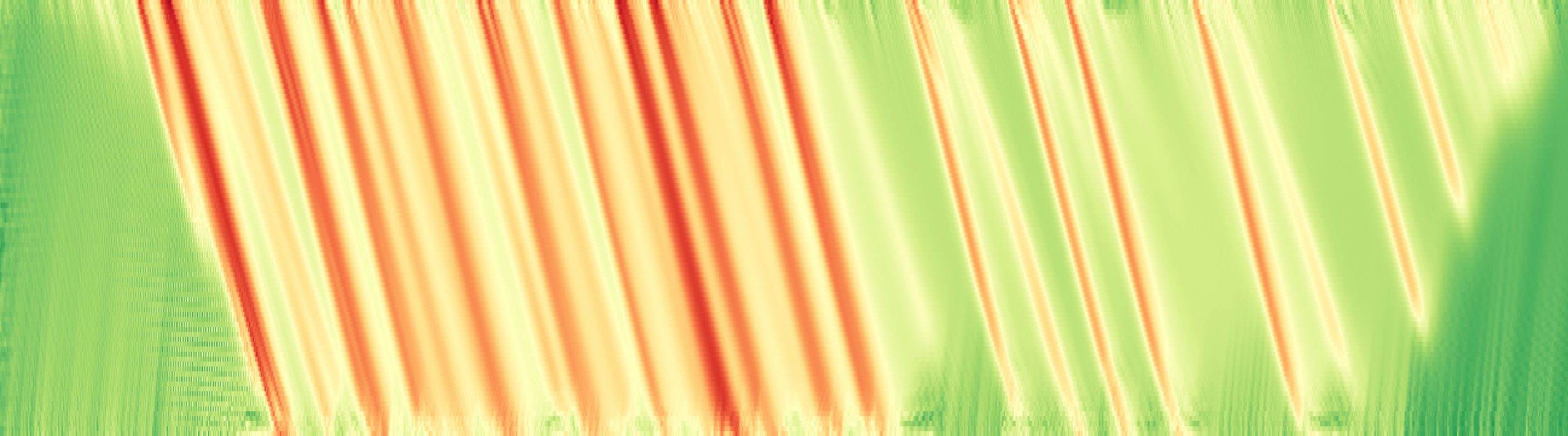}};
        \end{tikzpicture}
        \caption{$\text{NFV}_{11}^{11}$ prediction}
    \end{subfigure}
    \hfill
    \begin{subfigure}[t]{.325\textwidth}
        \begin{tikzpicture}
            \node[anchor=south west, inner sep=0, draw=black, line width=1.5pt] (img) at (0,0) {\includegraphics[width=0.99\textwidth]{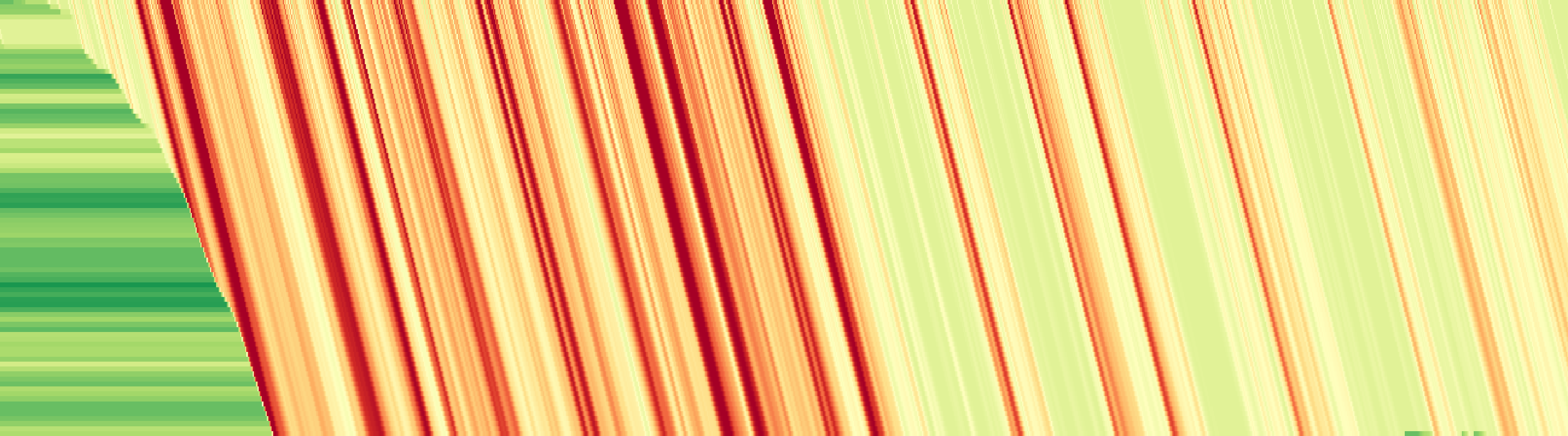}};
        \end{tikzpicture}
        \caption{Godunov prediction}
    \end{subfigure}
    
    \vspace{0.2cm}
    
    \caption*{\textbf{Generalization on unseen days}}
    
    \begin{subfigure}[t]{.325\textwidth}
      \begin{tikzpicture}
          \node[anchor=south west, inner sep=0, draw=black, line width=1.5pt] (img) at (0,0) {\includegraphics[width=0.99\textwidth]{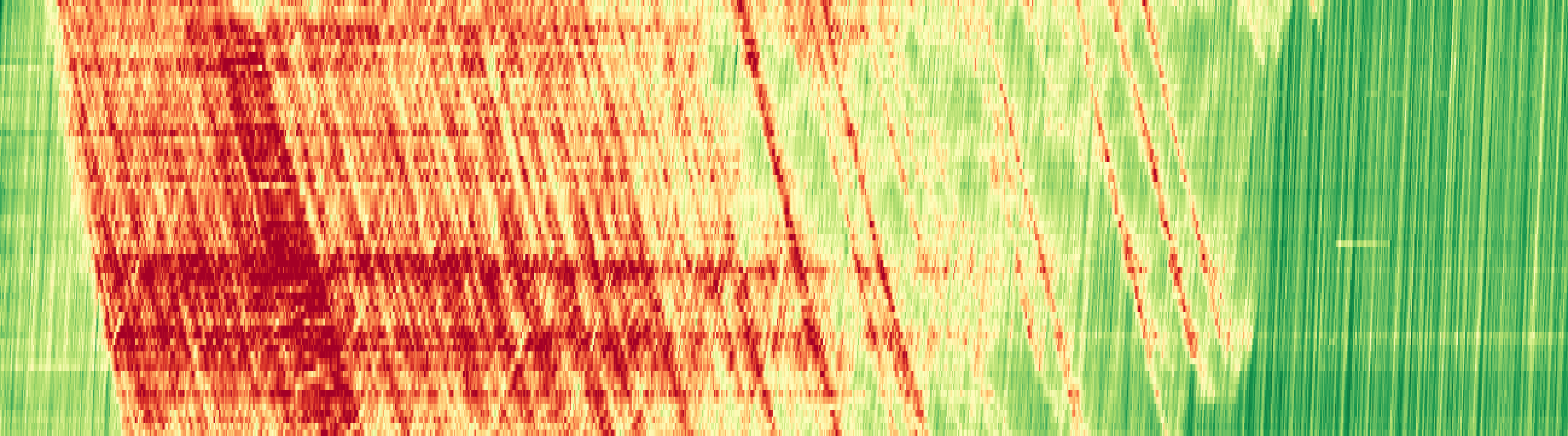}};
      \end{tikzpicture}
      \caption{Ground truth (Nov 21, 2022)}
    \end{subfigure}
    \hfill
    \begin{subfigure}[t]{.325\textwidth}
        \begin{tikzpicture}
            \node[anchor=south west, inner sep=0, draw=black, line width=1.5pt] (img) at (0,0) {\includegraphics[width=0.99\textwidth]{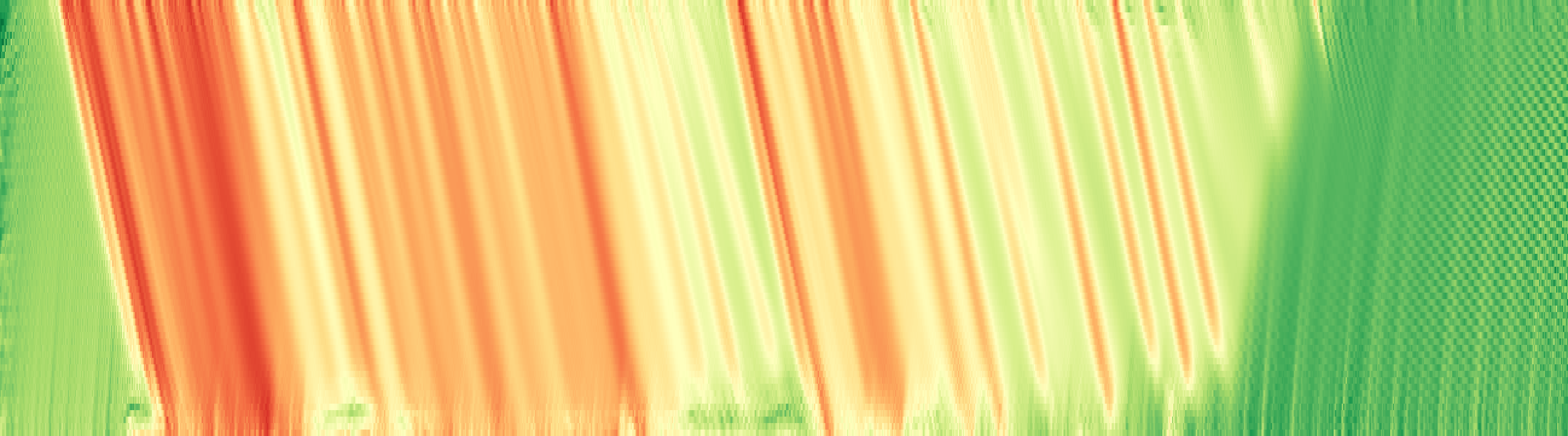}};
        \end{tikzpicture}
        \caption{$\text{NFV}_{11}^{11}$ prediction}
    \end{subfigure}
    \hfill
    \begin{subfigure}[t]{.325\textwidth}
        \begin{tikzpicture}
            \node[anchor=south west, inner sep=0, draw=black, line width=1.5pt] (img) at (0,0) {\includegraphics[width=0.99\textwidth]{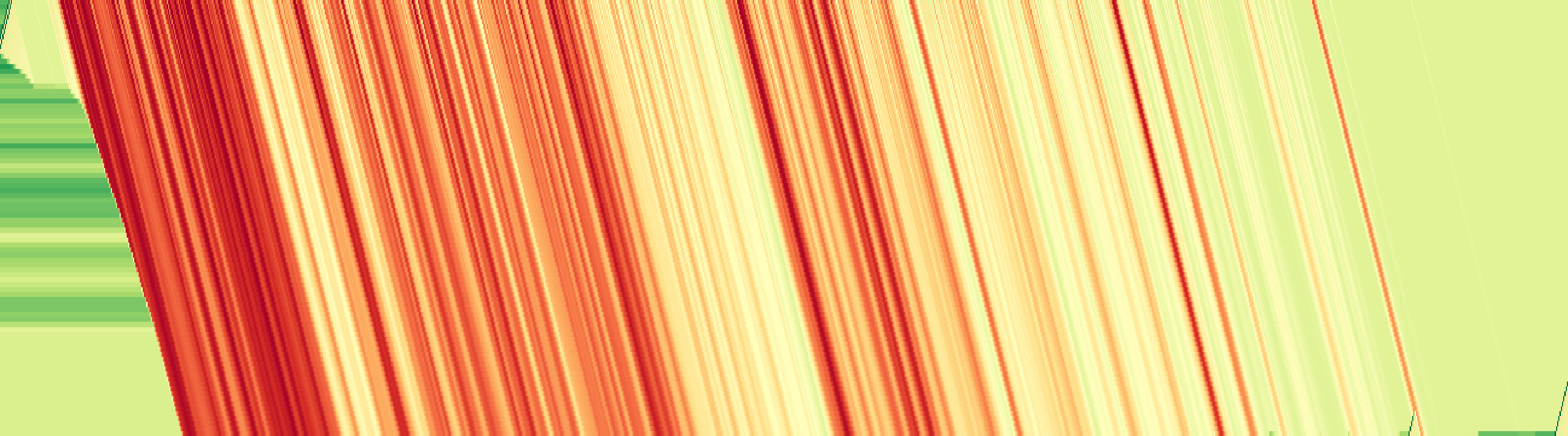}};
        \end{tikzpicture}
        \caption{Godunov prediction}
    \end{subfigure}
    
    \vspace{0.2cm}
    
    \begin{subfigure}[t]{.325\textwidth}
      \begin{tikzpicture}
          \node[anchor=south west, inner sep=0, draw=black, line width=1.5pt] (img) at (0,0) {\includegraphics[width=0.99\textwidth]{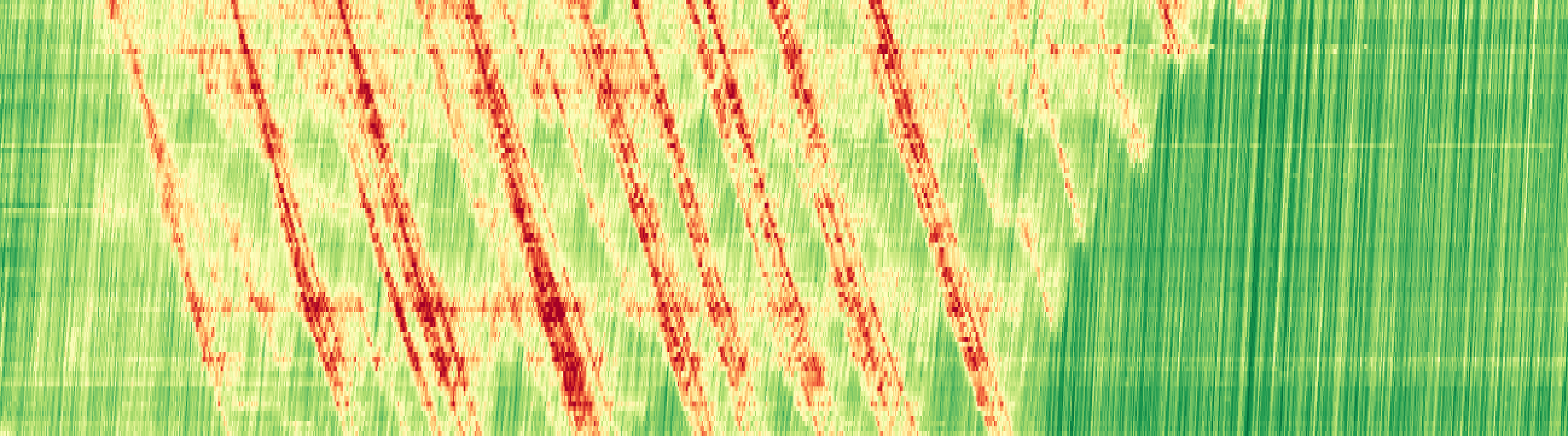}};
      \end{tikzpicture}
      \caption{Ground truth (Nov 22, 2022)}
    \end{subfigure}
    \hfill
    \begin{subfigure}[t]{.325\textwidth}
        \begin{tikzpicture}
            \node[anchor=south west, inner sep=0, draw=black, line width=1.5pt] (img) at (0,0) {\includegraphics[width=0.99\textwidth]{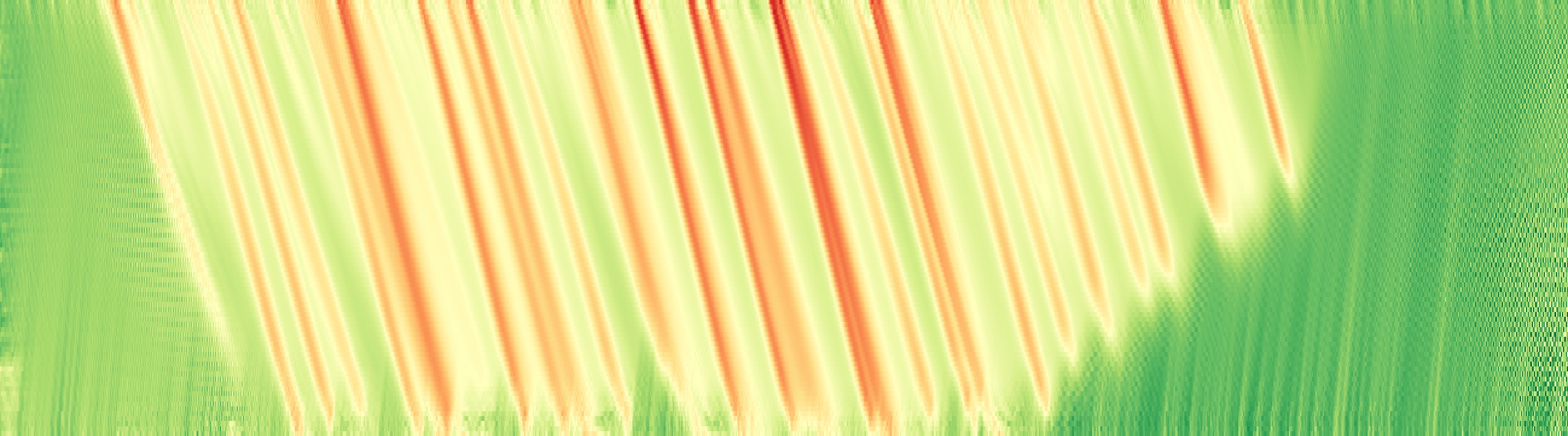}};
        \end{tikzpicture}
        \caption{$\text{NFV}_{11}^{11}$ prediction}
    \end{subfigure}
    \hfill
    \begin{subfigure}[t]{.325\textwidth}
        \begin{tikzpicture}
            \node[anchor=south west, inner sep=0, draw=black, line width=1.5pt] (img) at (0,0) {\includegraphics[width=0.99\textwidth]{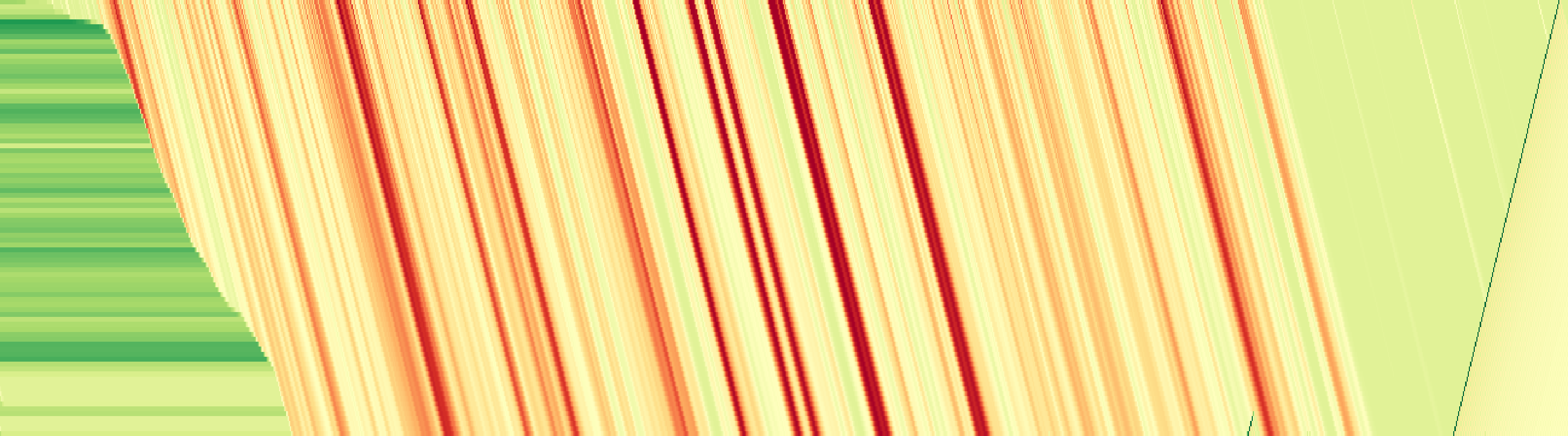}};
        \end{tikzpicture}
        \caption{Godunov prediction}
    \end{subfigure}
    
    \vspace{0.2cm}
    
    \begin{subfigure}[t]{.325\textwidth}
      \begin{tikzpicture}
          \node[anchor=south west, inner sep=0, draw=black, line width=1.5pt] (img) at (0,0) {\includegraphics[width=0.99\textwidth]{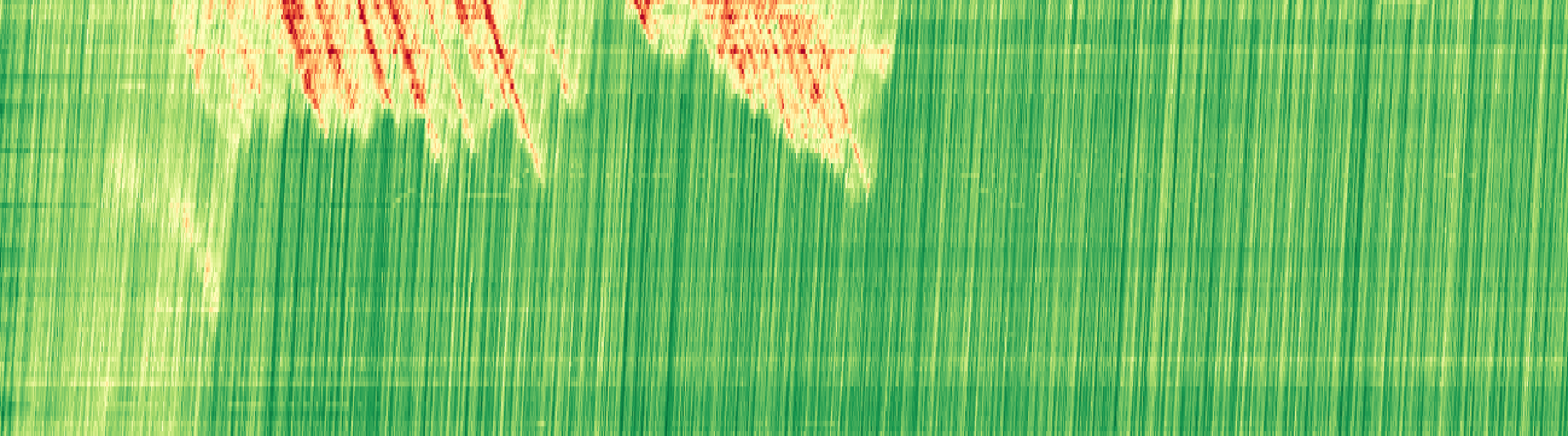}};
      \end{tikzpicture}
      \caption{Ground truth (Nov 23, 2022)}
    \end{subfigure}
    \hfill
    \begin{subfigure}[t]{.325\textwidth}
        \begin{tikzpicture}
            \node[anchor=south west, inner sep=0, draw=black, line width=1.5pt] (img) at (0,0) {\includegraphics[width=0.99\textwidth]{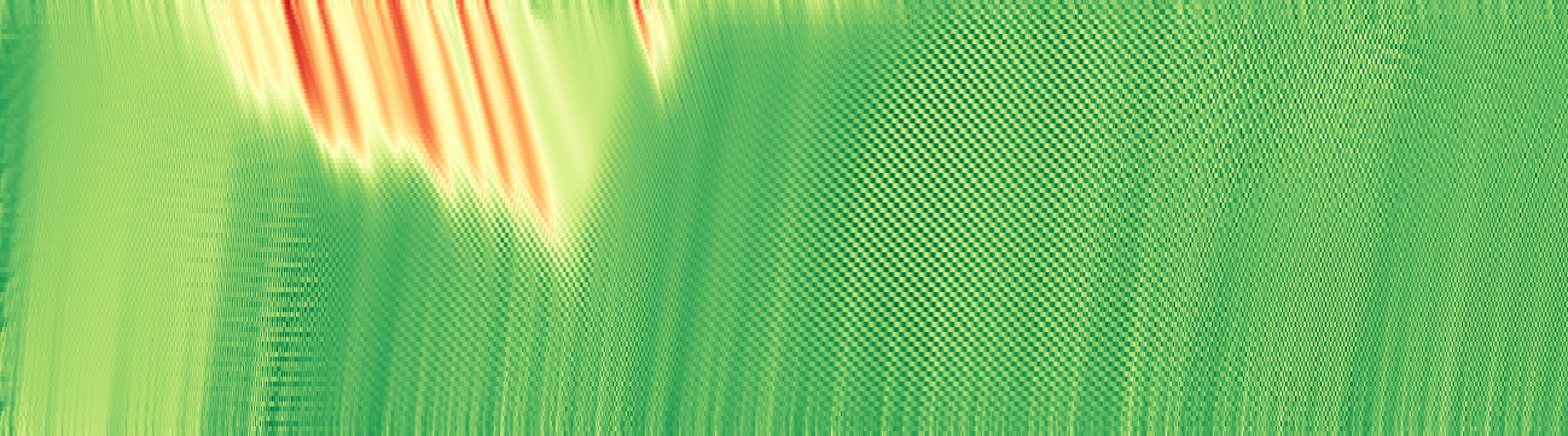}};
        \end{tikzpicture}
        \caption{$\text{NFV}_{11}^{11}$ prediction}
    \end{subfigure}
    \hfill
    \begin{subfigure}[t]{.325\textwidth}
        \begin{tikzpicture}
            \node[anchor=south west, inner sep=0, draw=black, line width=1.5pt] (img) at (0,0) {\includegraphics[width=0.99\textwidth]{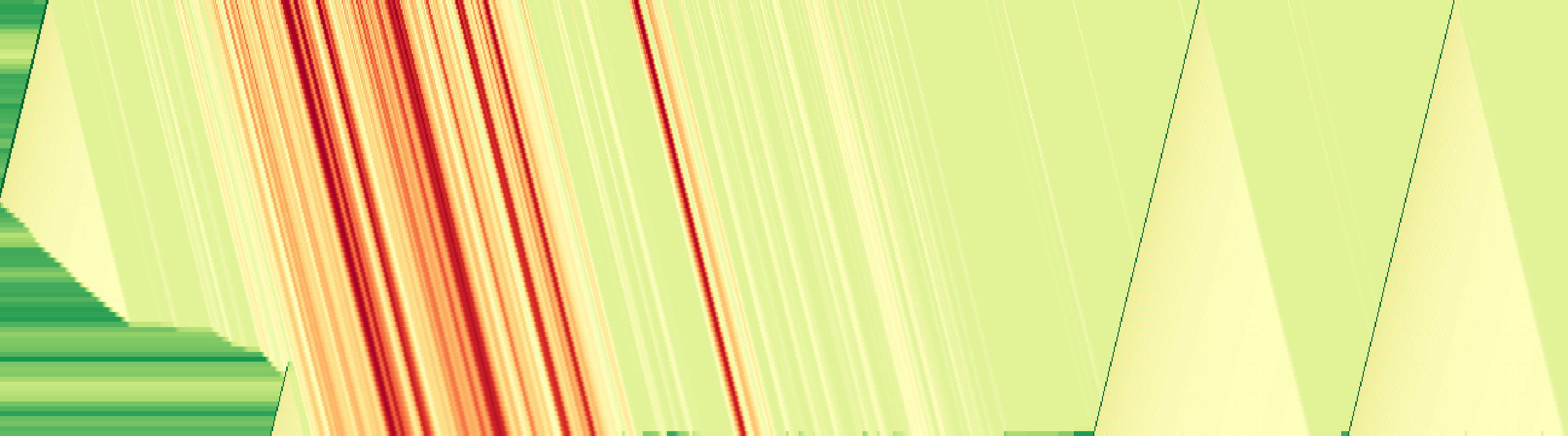}};
        \end{tikzpicture}
        \caption{Godunov prediction}
    \end{subfigure}
    
    \vspace{0.2cm}
    
    \begin{subfigure}[t]{.325\textwidth}
      \begin{tikzpicture}
          \node[anchor=south west, inner sep=0, draw=black, line width=1.5pt] (img) at (0,0) {\includegraphics[width=0.99\textwidth]{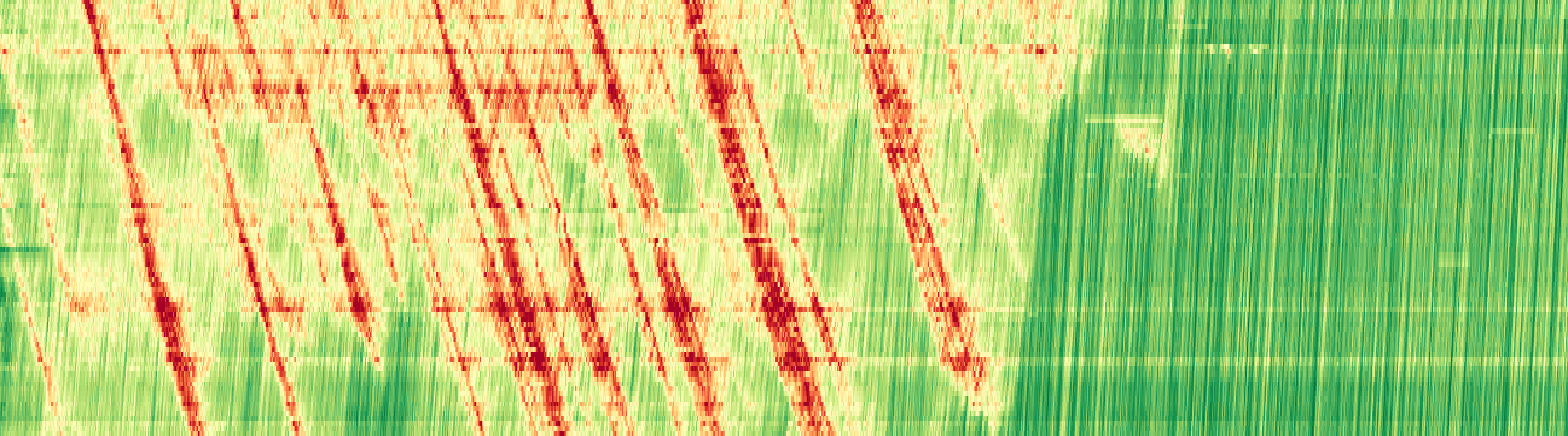}};
      \end{tikzpicture}
      \caption{Ground truth (Nov 28, 2022)}
    \end{subfigure}
    \hfill
    \begin{subfigure}[t]{.325\textwidth}
        \begin{tikzpicture}
            \node[anchor=south west, inner sep=0, draw=black, line width=1.5pt] (img) at (0,0) {\includegraphics[width=0.99\textwidth]{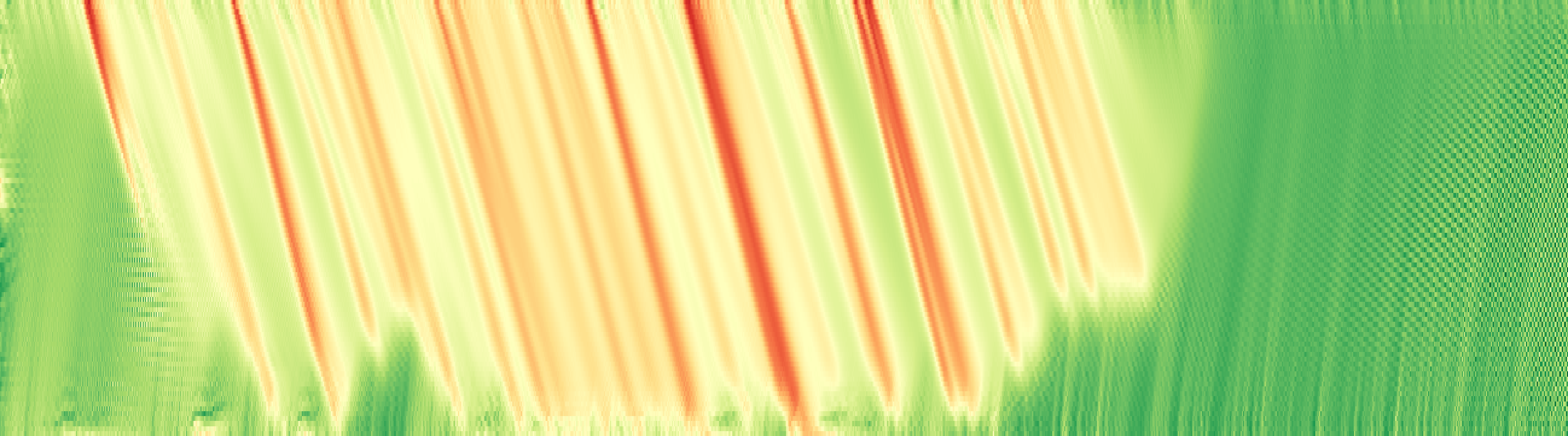}};
        \end{tikzpicture}
        \caption{$\text{NFV}_{11}^{11}$ prediction}
    \end{subfigure}
    \hfill
    \begin{subfigure}[t]{.325\textwidth}
        \begin{tikzpicture}
            \node[anchor=south west, inner sep=0, draw=black, line width=1.5pt] (img) at (0,0) {\includegraphics[width=0.99\textwidth]{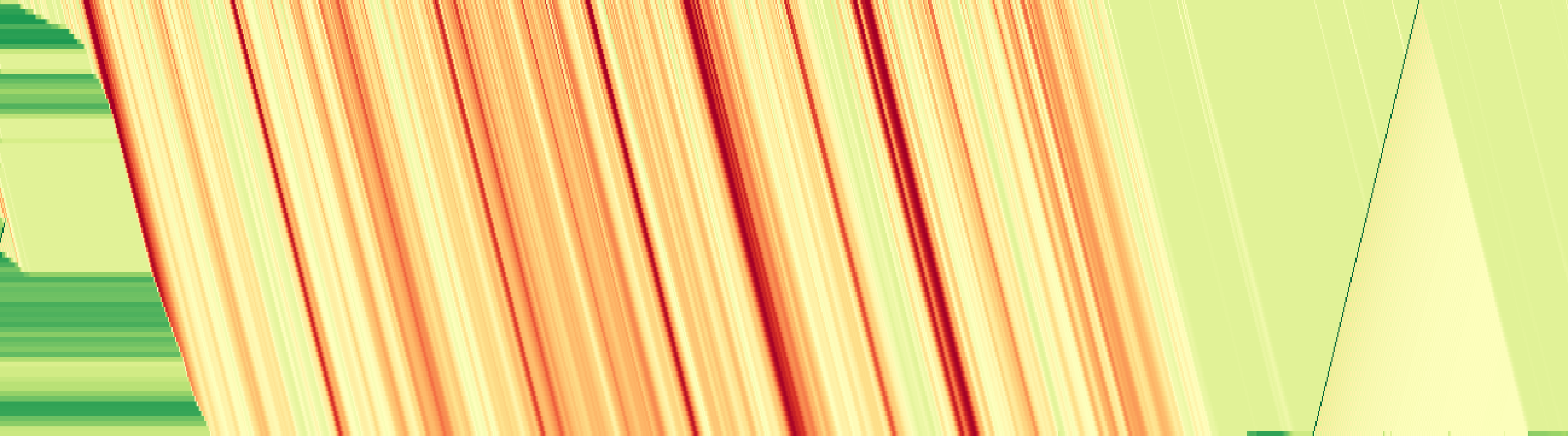}};
        \end{tikzpicture}
        \caption{Godunov prediction}
    \end{subfigure}
    
    \vspace{0.2cm}
    
    \begin{subfigure}[t]{.325\textwidth}
      \begin{tikzpicture}
          \node[anchor=south west, inner sep=0, draw=black, line width=1.5pt] (img) at (0,0) {\includegraphics[width=0.99\textwidth]{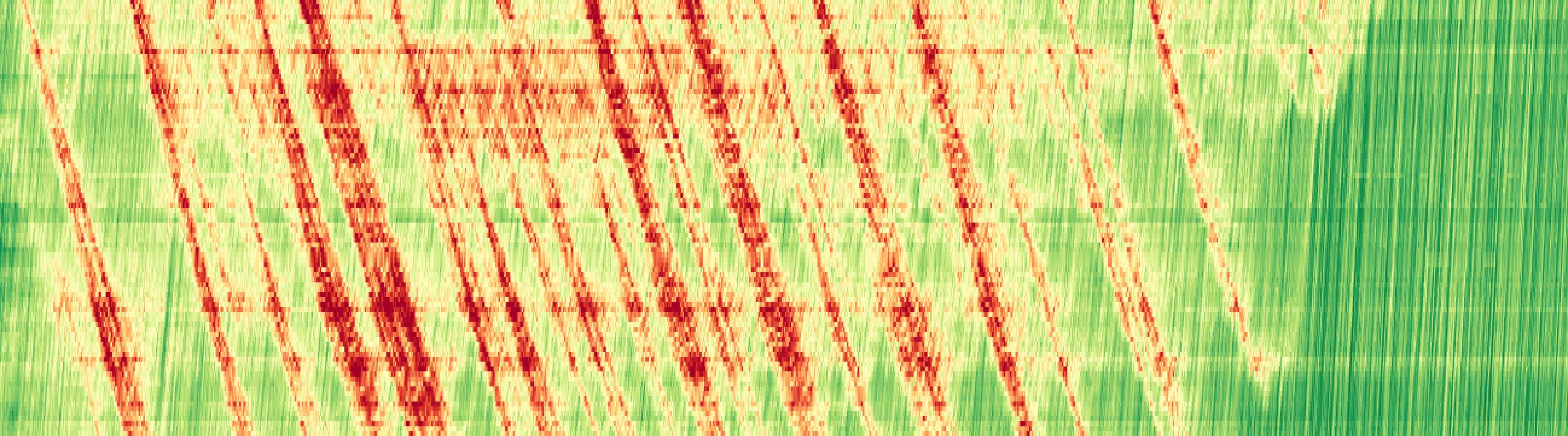}};
      \end{tikzpicture}
      \caption{Ground truth (Dec 01, 2022)}
    \end{subfigure}
    \hfill
    \begin{subfigure}[t]{.325\textwidth}
        \begin{tikzpicture}
            \node[anchor=south west, inner sep=0, draw=black, line width=1.5pt] (img) at (0,0) {\includegraphics[width=0.99\textwidth]{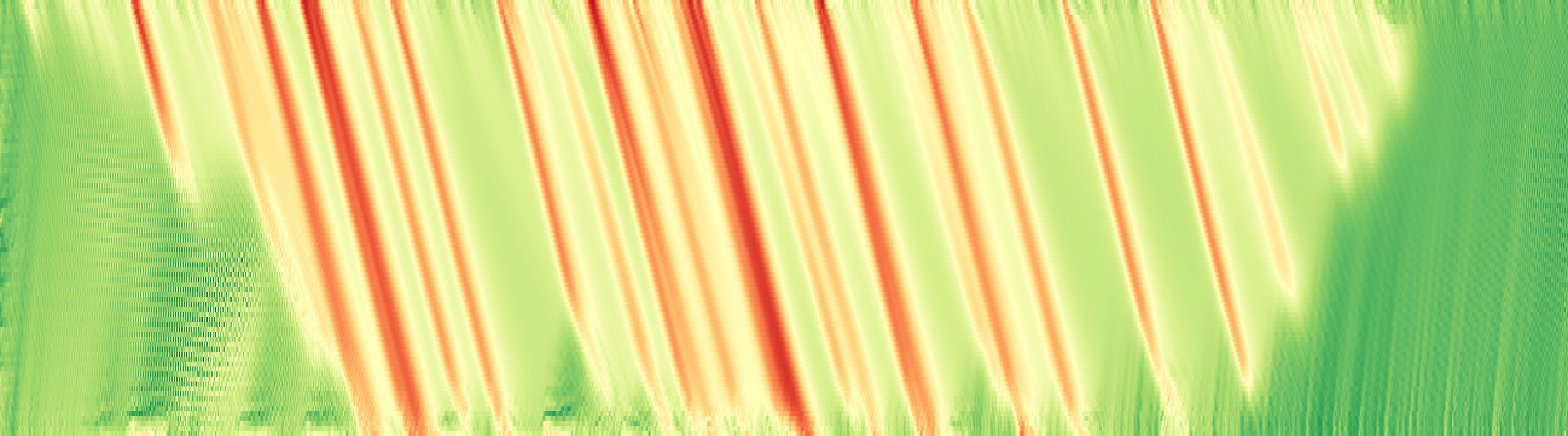}};
        \end{tikzpicture}
        \caption{$\text{NFV}_{11}^{11}$ prediction}
    \end{subfigure}
    \hfill
    \begin{subfigure}[t]{.325\textwidth}
        \begin{tikzpicture}
            \node[anchor=south west, inner sep=0, draw=black, line width=1.5pt] (img) at (0,0) {\includegraphics[width=0.99\textwidth]{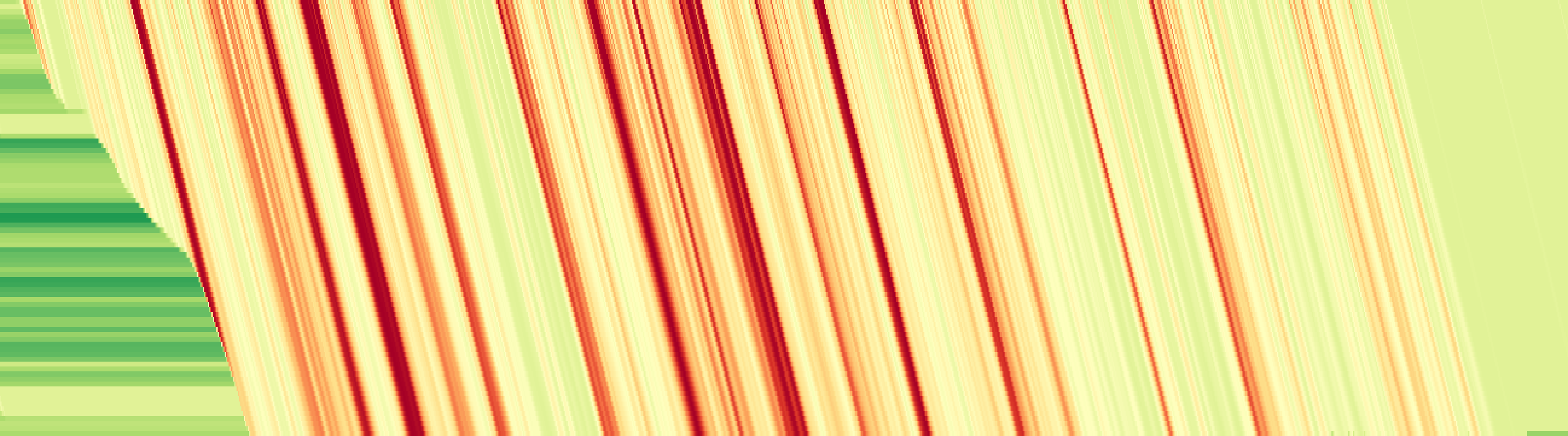}};
        \end{tikzpicture}
        \caption{Godunov prediction}
    \end{subfigure}
    
    \vspace{0.2cm}
    
    \begin{subfigure}[t]{.325\textwidth}
      \begin{tikzpicture}
          \node[anchor=south west, inner sep=0, draw=black, line width=1.5pt] (img) at (0,0) {\includegraphics[width=0.99\textwidth]{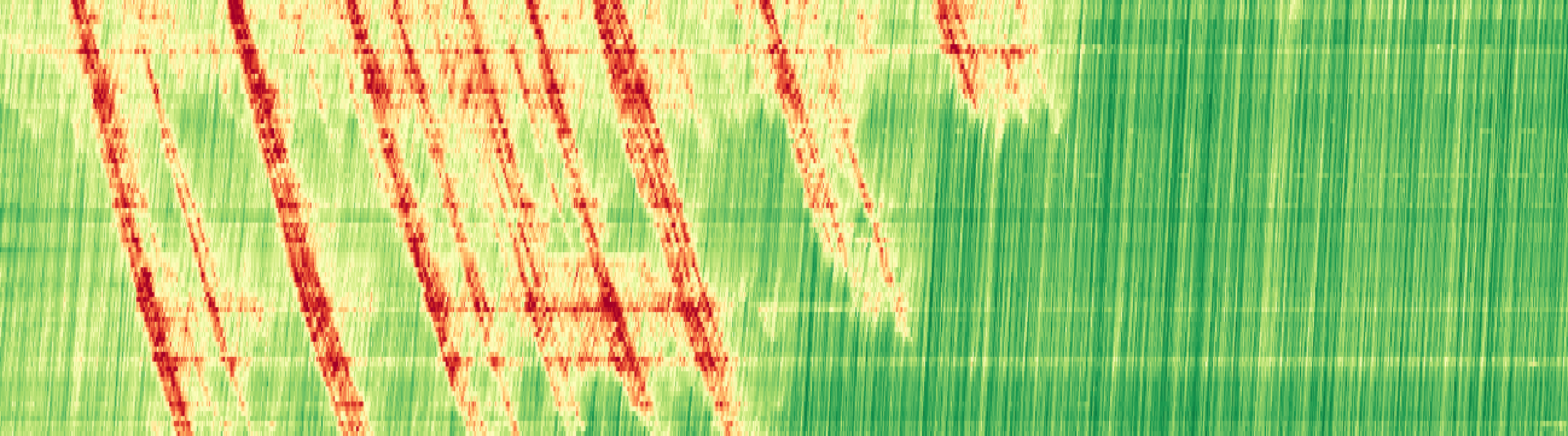}};
      \end{tikzpicture}
      \caption{Ground truth (Dec 02, 2022)}
    \end{subfigure}
    \hfill
    \begin{subfigure}[t]{.325\textwidth}
        \begin{tikzpicture}
            \node[anchor=south west, inner sep=0, draw=black, line width=1.5pt] (img) at (0,0) {\includegraphics[width=0.99\textwidth]{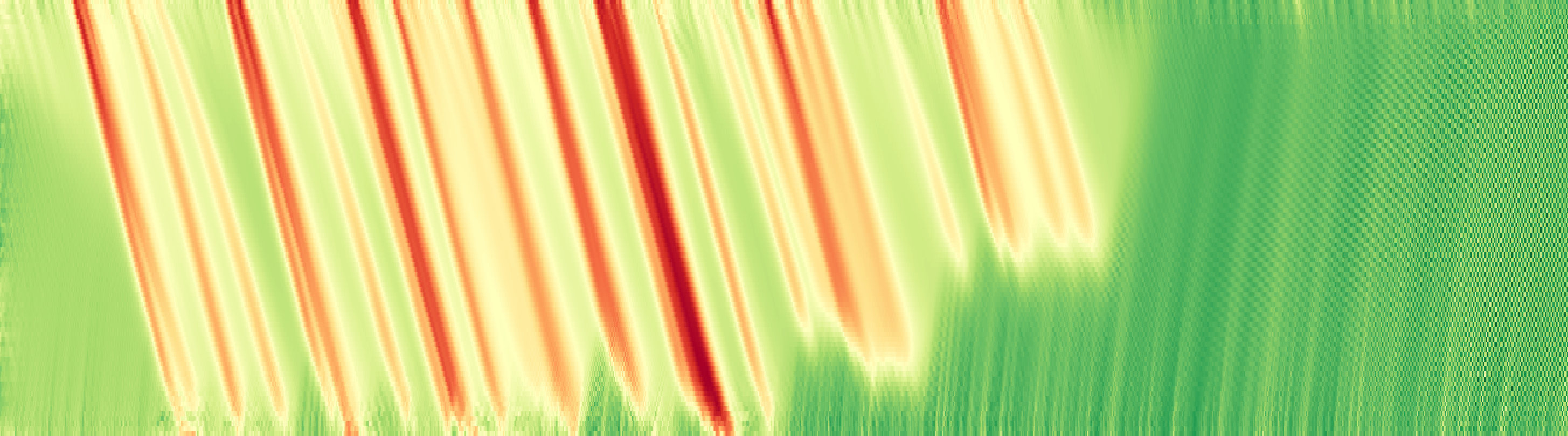}};
        \end{tikzpicture}
        \caption{$\text{NFV}_{11}^{11}$ prediction}
    \end{subfigure}
    \hfill
    \begin{subfigure}[t]{.325\textwidth}
        \begin{tikzpicture}
            \node[anchor=south west, inner sep=0, draw=black, line width=1.5pt] (img) at (0,0) {\includegraphics[width=0.99\textwidth]{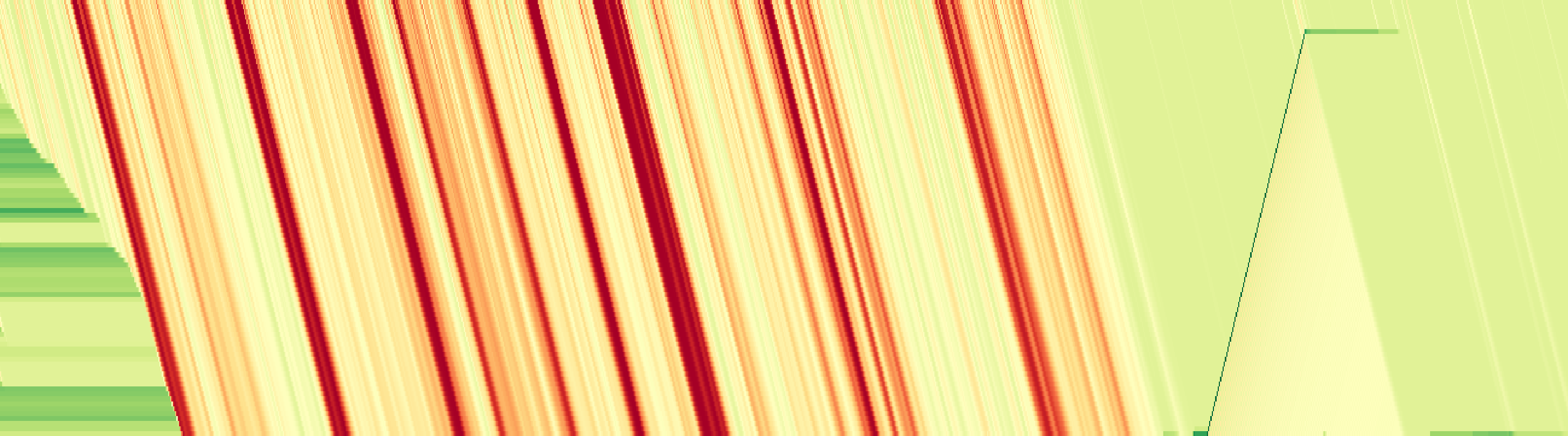}};
        \end{tikzpicture}
        \caption{Godunov prediction}
    \end{subfigure}
    
    \vspace{0.2cm}
    
    \begin{subfigure}[t]{.325\textwidth}
      \begin{tikzpicture}
          \node[anchor=south west, inner sep=0, draw=black, line width=1.5pt] (img) at (0,0) {\includegraphics[width=0.99\textwidth]{11-21-2022_gt_TX.png}};
      \end{tikzpicture}
      \caption{Ground truth (Nov 21, 2022)}
    \end{subfigure}
    \hfill
    \begin{subfigure}[t]{.325\textwidth}
        \begin{tikzpicture}
            \node[anchor=south west, inner sep=0, draw=black, line width=1.5pt] (img) at (0,0) {\includegraphics[width=0.99\textwidth]{11-21-2022_pred_nfv11x11_TX.png}};
        \end{tikzpicture}
        \caption{$\text{NFV}_{11}^{11}$ prediction}
    \end{subfigure}
    \hfill
    \begin{subfigure}[t]{.325\textwidth}
        \begin{tikzpicture}
            \node[anchor=south west, inner sep=0, draw=black, line width=1.5pt] (img) at (0,0) {\includegraphics[width=0.99\textwidth]{11-21-2022_pred_god_TX.png}};
        \end{tikzpicture}
        \caption{Godunov prediction}
    \end{subfigure}
    
        \caption{\textbf{Predictions of best FV fit and trained $\text{NFV}_{11}^{11}$.} Godunov is derived by fitting a flow function on the prediction and comparing it against the ground truth; we keep the fitted Trapezoidal flow as it performed best (see \Cref{fig:i24_part1_all_predictions} and \Cref{tab:i24_part1_metrics_table}). Both the Godunov fit and the $\text{NFV}_{11}^{11}$ training are realized using the same data, namely the first 1 hour of Nov 29, 2022 data (i.e., the first 25\% of subfigure (a)). This means that the remainder of the data on that day (row 1), as well as the prediction on all subsequent days (rows 2-8) are generalization on data that was never seen before by either models. See \Cref{sec:how_to_read} for how to read the heatmaps.}
        \label{fig:i24_part1_generalization_all_predictions}
    \end{figure}

Predictions on the training day from \Cref{sec:i24_part1} are displayed in \Cref{fig:i24_part1_all_predictions}. Predictions on evaluation days are displayed in \Cref{fig:i24_part1_generalization_all_predictions}.

\section{Experiment details} \label{app:exp_details}

\paragraph{Model Architecture.}
The model is applied locally on each cell to estimate the corresponding numerical flux. It is implemented as a one-dimensional CNN. The first layer uses a kernel of size $a-1$, followed by five 1D convolutional layers with 15 channels and kernel size 1. Using a CNN enables efficient vectorized computation over all stencils, which is equivalent to sliding a fully connected network along the input but significantly faster. Each time step is represented as a separate input channel, for a total of $b$ input channels. The output is a single channel providing the estimated flux.

\paragraph{Training on Synthetic Data (LWR).}
For the LWR model, training is performed autoregressively: the model predicts future time steps by feeding its own outputs as inputs. To mitigate error accumulation, the prediction horizon is gradually increased from 10 to 250 steps during training. Initially, most training uses a 10-step horizon, which already yields satisfying performance. Fine-tuning with longer horizons improves stability. The model is trained with the Adam optimizer and an exponentially decaying learning rate, from $10^{-4}$ to $10^{-8}$. Training is done on an RTX 6000 GPU and takes approximately 30 minutes.

\paragraph{Dataset.} $400$ Riemann problems are randomly sampled for training, with discretization parameters \(\Delta t = 5\cdot 10^{-3}\) and \(\Delta x = 10^{-2}\), with $100$ space cells and up to $250$ time steps. Training details can be found in \Cref{app:exp_details}. A finer grid is used to compute the $100$ test cases for evaluation using the Lax-Hopf algorithm: \(\Delta t = 10^{-4}\) and \(\Delta x = 10^{-3}\) with \(200\) cells and \(1,000\) time steps.

\paragraph{Training on Experimental Data.}
All models and fitted finite volume schemes are trained on the first hour of data from November 29, 2022 (arbitrarily selected), and evaluated on the full morning period (nearly four hours) and the remaining days of data. To ensure fairness and reflect practical deployment constraints, each model receives only a single boundary cell on each side, as described in \Cref{app:i24_boundary_conditions}, even though larger models could benefit from additional context. Each NFV model consists of 6 hidden layers of width 15, totaling $1105 + 16 \cdot ((a - 1) \cdot b + 1)$ parameters for $\text{NFV}_a^b$. Training takes 15 to 30 minutes on an RTX 4080 GPU. As with synthetic data, the prediction horizon increases progressively during training, from 10 to 100 steps, and the learning rate decays from $10^{-3}$ to $10^{-4}$ over 3000-5000 epochs depending on model size, until convergence.

\end{document}